\documentclass[11pt]{elsarticle}

\makeatletter
\def\ps@pprintTitle{%
	\let\@oddhead\@empty
	\let\@evenhead\@empty
	\def\@oddfoot{}%
	\let\@evenfoot\@oddfoot}
\usepackage{bm}
\usepackage{bbm}
\usepackage{amsmath}
\usepackage{amsthm}
\usepackage{empheq}
\usepackage{float}
\usepackage{indentfirst}
\usepackage{cases}
\usepackage{algorithm}  
\usepackage{algorithmicx} 
\usepackage{algpseudocode} 
\usepackage{graphicx}
\usepackage{booktabs}
\usepackage{diagbox}
\usepackage[all]{xy}
\usepackage{indentfirst}
\usepackage{epstopdf}
\usepackage{epsfig}
\usepackage{overpic}
\usepackage{array}
\usepackage{color}
\usepackage{multirow}
\usepackage[mathscr]{euscript}
\usepackage{latexsym,bm}
\usepackage{tikz}
\usetikzlibrary{shapes,arrows}
\usepackage[left=2.5cm,right=2.5cm,top=2cm,bottom=2cm]{geometry}
\usepackage{enumerate}
\usepackage{autobreak}
\usepackage{subfigure}
\allowdisplaybreaks

\usepackage{ulem}

\usepackage{framed,multirow}

\usepackage{amssymb}
\usepackage{latexsym}
\usepackage{mathtools}
\usepackage{multirow}

\usepackage{url}
\usepackage{xcolor}
\definecolor{newcolor}{rgb}{.8,.349,.1}

\renewcommand{\vec}[1]{\boldsymbol{#1}} 

\def\bx{\bm{x}}

\def\by{\bm{y}}

\def\bmm{\bm{m}}
\def\bu{\bm{u}}

\def\D{\textup{d}}
\def\mone{\mathbbm{1}}

\makeatletter

\newcommand{\Rmnum}[1]{\expandafter@slowromancap\romannumeral #1@}
\makeatother

\tikzset{global scale/.style={
		scale=#1,
		every node/.append style={scale=#1}
	}
}
\tikzstyle{startstop} = [rectangle,rounDSP corners, minimum width=3cm,minimum height=1cm,text centered, draw=black,fill=red!30]
\tikzstyle{io} = [trapezium, trapezium left angle = 70,trapezium right angle=110,minimum width=3cm,minimum height=1cm,text centered,draw=black,fill=blue!30]
\tikzstyle{process} = [rectangle,minimum width=3cm,minimum height=1cm,text centered,text width =3cm,draw=black,fill=orange!30]
\tikzstyle{decision} = [diamond,minimum width=1.7cm,minimum height=0.5cm,shape aspect=2, text centered,draw=black,fill=green!30]
\tikzstyle{arrow} = [thick,->,>=stealth]

\newtheorem{definition}{Definition}[section]
\newtheorem{theorem}[definition]{Theorem}
\newtheorem{lemma}[definition]{Lemma}

\begin{document}

\begin{frontmatter}

\title{\LARGE Density estimation via mixture discrepancy and moments}

\author[PKU]{Zhengyang Lei}
\ead{leizy@stu.pku.edu.cn}

\author[BNU]{Lirong Qu}
\ead{lirong.qu@mail.bnu.edu.cn}

\author[PKU]{Sihong Shao}
\ead{sihong@math.pku.edu.cn}

\author[BNU]{Yunfeng Xiong}
\ead{yfxiong@bnu.edu.cn}


\address[PKU]{CAPT, LMAM and School of Mathematical Sciences,  Peking University, Beijing 100871, China}
\address[BNU]{School of Mathematical Sciences, Beijing Normal University, Beijing 100875, China}


\begin{abstract}
With the aim of generalizing histogram statistics to higher dimensional cases, density estimation via discrepancy based sequential partition (DSP) has been proposed to learn an adaptive piecewise constant approximation defined on a binary sequential partition of the underlying domain, where the star discrepancy is adopted to measure the uniformity of particle distribution. However, the calculation of the star discrepancy is NP-hard and it does not satisfy the reflection invariance and rotation invariance either. To this end, we use the mixture discrepancy and the comparison of moments as a replacement of the star discrepancy, leading to the density estimation via mixture discrepancy based sequential partition (DSP-mix) and density estimation via moment-based sequential partition (MSP), respectively. Both DSP-mix and MSP are computationally tractable and exhibit the reflection and rotation invariance. Numerical experiments in reconstructing Beta mixtures, Gaussian mixtures and  heavy-tailed Cauchy mixtures up to 30 dimension are conducted, demonstrating that MSP can maintain the same accuracy compared with DSP, while gaining an increase in speed by a factor of two to twenty for large sample size, and DSP-mix can achieve satisfactory accuracy and boost the efficiency in low-dimensional tests ($d \le 6$), but might lose accuracy in high-dimensional problems due to a reduction in partition level.
\end{abstract}


\end{frontmatter}

\section{Introduction}

As a fundamental problem in statistics, density estimation aims at constructing an estimate of their common density function $q(\by)$ for given $N$  observations $\by_1, \by_2, \dots, \by_N$. It has various applications in uncertainty quantification \cite{KirstenBilginPetangodaStanleyMarbell2025}, Bayesian inference \cite{Castillo2017} and scientific computing \cite{Yan2015,DolgovAnayaIzquierdoFoxScheichl2020,RenZhaoKhooYing2023,XiongShao2020Overcoming}. A simple and classical approach is the histogram statistics as a uniform piecewise constant density estimator. Unfortunately,  it is rather difficult to apply the histogram to high-dimensional problems as both bin size and required sample size grow exponentially in dimension, which is well known as the curse of dimensionality (CoD) \cite{bk:Silverman2018,Yan2015}.

 This paper focuses on nonparametric approaches for multivariate density estimation, especially for the case when the dimension $d$ is
moderately large, say, $6–30$ \cite{LuoJiangWong2013,RuzgasLukauskasCepkauskas2021}. The benefit of nonparametric methods
is their ability to achieve estimation optimality for any input distribution as more data are observed. A widely used method is the kernel density estimation (KDE),  treating each data point as the center of a kernel function like the Gaussian functions and smoothed splines \cite{bk:Silverman2018}. But  the accuracy of
the kernel estimator becomes very sensitive to the choice of the window size and the shape
of the kernel \cite{WuYuWangRao2024}, and  the magic lies in how to balance the bias and variance of the estimator, like using the smoothing properties of linear diffusion processes \cite{BotevGrotowskiKroese2010} or choosing adaptive bandwidths \cite{Zame2024kernel,WuYuWangRao2024}. 
Another approach is based on tensor decompositions, with huge advantages in that the computational cost of the construction, the storage requirements and the operations required for conditional distribution method sampling from the distributional approximation all scale linearly with dimension \cite{DolgovAnayaIzquierdoFoxScheichl2020,RenZhaoKhooYing2023,WangLinLiaoLiuXie2024}. However, it still relies heavily on the low-rank assumption of the underlying density, whereas the detection and characterization of local features of multivariate density estimation are usually prohibitive \cite{LuoJiangWong2013}. Machine learning approaches have recently been drawing a growing attention, e.g., the density estimation through normalizing flows
\cite{TakakTurner2013} and deep generative neural networks \cite{LiuXuJiangWong2021}. In practice, they require extensive tuning to
perform well \cite{LiuLiWong2023}.
 
Here we mainly discuss a data-driven density estimation method, termed tree-based density estimation. This features a class of estimators which employs simple and flexible binary partitions to adapt to the underlying density function, along with the decision tree using stopping times in a data-driven way \cite{Castillo2017}. The decision tree based method has a great potential to overcome the burden imposed by the high dimensionality, especially when the dimension is only moderately large  and the density function exhibits certain spatial features that can be leveraged of \cite{LiuLiWong2023}.  Essentially, it is equivalent to cluster the samples into small nonintersecting sets, each supported by a tree-structured density \cite{LiuXuGuGuptaLaffertyWasserman2011}. An adaptive nonparametric density estimation can be constructed by either the Bayesian Sequential Partition (BSP) \cite{LuoJiangWong2013,LiuLiWong2023} or the Discrepancy based Sequential Partition (DSP) \cite{Li2016}. Compared with BSP, the computational cost of DSP is much cheaper due to its greedy construction, and is fully capable to handle higher dimension \cite{XiongShao2020Overcoming}.

To be more specific, DSP uses the star discrepancy, a concept from the quasi-Monte Carlo method, to measure the uniformity of the observations \cite{dick2013high, hickernell1998generalized}. For a point set $\left(\by_1,\dots,\by_n\right)\subset \left[0,1\right]^d$, its star discrepancy reads
\begin{equation}\label{star dis}
	D^*\left(\by_1, \ldots, \by_n\right)=\sup _{\boldsymbol{u} \in[0,1]^d}\left|\frac{1}{n} \sum_{i=1}^n \mone_{[0, \boldsymbol{u})}\left(\by_i\right)-\operatorname{vol}([0, \boldsymbol{u}))\right|,
\end{equation}
and characterizes the uniformity of the points by using the infinity norm of the difference between the cumulative distribution function of the discrete point measure $\frac{1}{n} \sum_{i=1}^{n} \delta_{\by_i}$ and that of the uniform measure. When the star discrepancy of the observations within a sub-domain is small, it can be reasonable to think that their distribution approaches the uniform distribution and thus the corresponding probability density can be regarded as approximately constant in the sub-domain.

The major bottleneck of DSP lies in the complexity of calculating the star discrepancy of pointsets in the $d$-dimensional unit cube, which has been proved to be an NP-hard problem \cite{GnewuchSrivastavWinzen2009}. As a consequence, the external approximate solver \cite{Fang1997thres,gnewuch2012new,ShaoWu2025} is mandatory, but is rather time-consuming when the sample size $N$ is large. Besides, the star discrepancy lacks both reflection and rotation invariance \cite{hickernell1998generalized,fang2018theory,zhou2013mixture} since the origin $\vec{0}$ is distinctive in the definition~\eqref{star dis} of the star discrepancy.
In view of such drawbacks, this paper contributes two new uniformity measurement strategies for density estimation, both of which seek a relaxation of DSP. One is to adopt the mixture discrepancy, another kind of discrepancy \cite{hickernell1998generalized,fang2018theory,zhou2013mixture}. 
	For a point set $\left(\by_1,\dots,\by_n\right)\subset \left[0,1\right]^d$, its mixture discrepancy has an analytical expression,
	\begin{equation}\label{dsc mix}
		\begin{aligned}
			 \left(D^{\text{mix}}(\mathbf{y}_1, \ldots, \mathbf{y}_n)\right)^2
			=&  \left(\frac{19}{12}\right)^d-\frac{2}{n} \sum_{i=1}^n \prod_{j=1}^d\left(\frac{5}{3}-\frac{1}{4}\left|y_{i j}-\frac{1}{2}\right|-\frac{1}{4}\left|y_{i j}-\frac{1}{2}\right|^2\right) \\
			&+\frac{1}{n^2} \sum_{i=1}^n \sum_{k=1}^n \prod_{j=1}^d\left(\frac{15}{8}-\frac{1}{4}\left|y_{i j}-\frac{1}{2}\right|-\frac{1}{4}\left|y_{k j}-\frac{1}{2}\right|\right. \\
			&\left.-\frac{3}{4}\left|y_{i j}-y_{k j}\right|+\frac{1}{2}\left|y_{i j}-y_{k j}\right|^2\right),
		\end{aligned}
	\end{equation}
	where $\by_i = (y_{i1},\dots,y_{id})$. It is thus easy to calculate, and possesses good theoretical properties. The interested readers are referred to \cite{hickernell1998generalized,fang2018theory,zhou2013mixture} for more details about the mixture discrepancy including the motivation and the calculation. We use
	the same algorithm framework as DSP except for replacing the star discrepancy with 
	the mixture discrepancy. The resulting method is dubbed the DSP-mix method. 
	
	 The other is the density estimation via moment-based sequential partition (MSP), which still adopts the partitioning framework in DSP, but the way to measure the uniformity of the observations is replaced by comparing the gap in moments between the particle distribution and the uniform density.	 
Whether the density function can be uniquely determined by the moments of all orders is the classical moment problem in probability theory \cite{billingsley2012probability, Gut2013prob, Serfling1980Approx}. For the case where the probability density is defined on a compact set, this problem is called the Hausdorff moment problem, and the answer is positive \cite{Kleiber2013moment}. When the moments of truncated orders of the observations on the compact set are close to those of the uniform distribution, we can consider that the observations are approximately uniformly distributed. 
Numerical experiments in reconstructing Beta mixtures, Gaussian mixtures and  heavy-tailed Cauchy mixtures up to 30 dimension are conducted, demonstrating that MSP can maintain the same accuracy compared with DSP, while gaining an increase in speed by a factor of two to twenty for large sample size, and DSP-mix can achieve satisfactory accuracy  in low-dimensional tests ($d \le 6$) and run significantly faster than DSP,  but might lose accuracy in high-dimensional problems due to a reduction in partition level.

The rest of this paper is organized as follows. The DSP-mix and MSP methods are introduced in Sections~\ref{sec:dsp-mix} and \ref{sec msp}, respectively. Section~\ref{sec:num} conducts several numerical experiments and compares DSP-mix and MSP with DSP in performance. The paper is concluded in Section~\ref{conclus}.

\section{Discrepancy sequential partition via star discrepancy or mixed discrepancy}
\label{sec:dsp-mix}

\begin{figure}[!h]
\centering
\includegraphics[width=1.\textwidth]{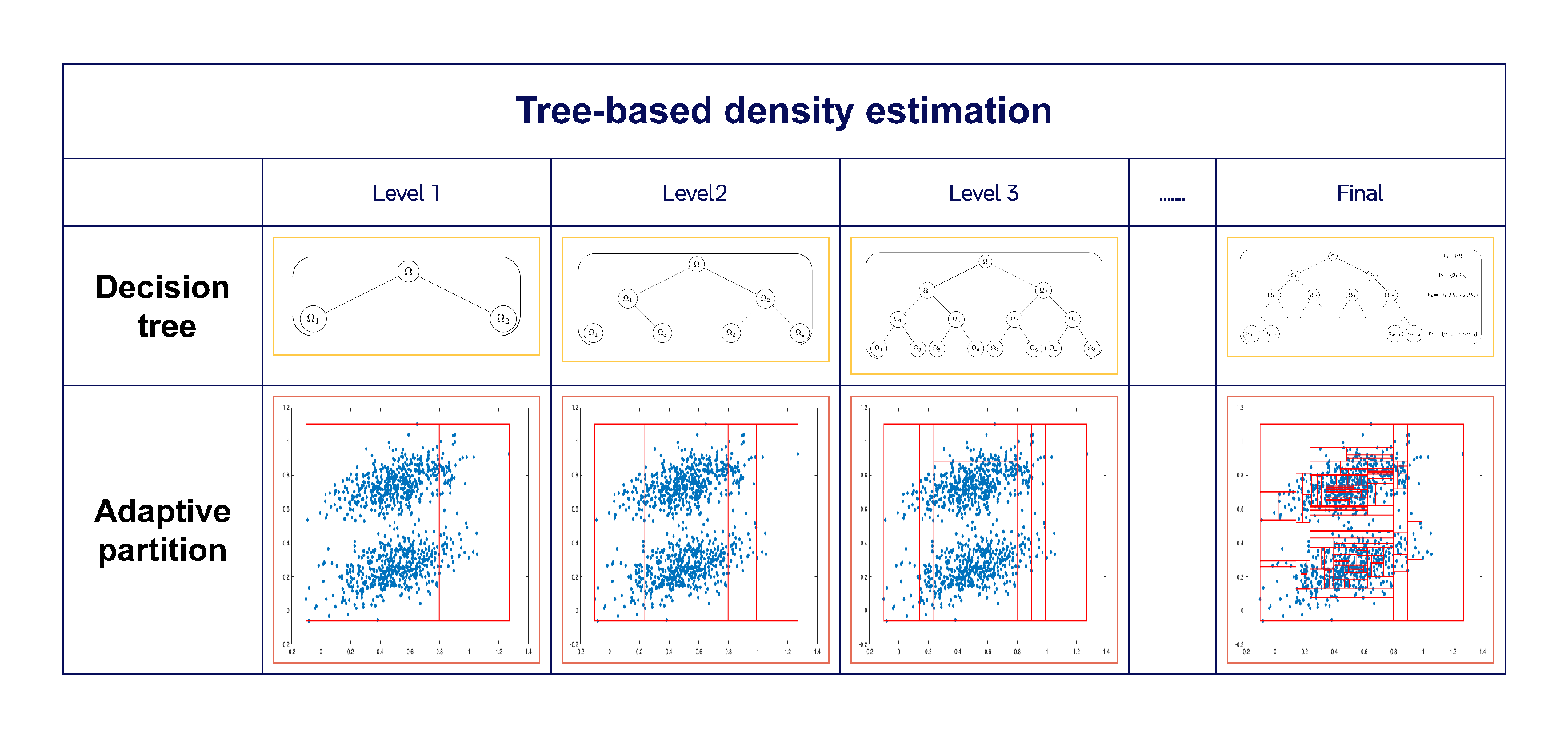}
	\caption{An illustration of tree-based density estimation.}
	\label{TDE}
\end{figure}

Without loss of generality, assume that the observations $\by_i \in \Omega,\,i = 1,2,\dots,N$, and the computational domain $\Omega$ is decomposed into $L$ non-intersecting sub-domains,
\begin{equation}\label{decom omega}
	\Omega=\bigcup_{l=1}^{L} \Omega_l.
\end{equation}
DSP uses an adaptive piecewise constant density estimator to reconstruct the $d$-D probability density function from the observations \cite{Li2016},
\begin{equation}
	\hat{p}(\by) = \sum_{l=1}^{L} c_l \mone_{\Omega_l}(\by), \quad c_l = \frac{1}{N |\Omega_l|} \sum_{i=1}^{N} \mone_{\Omega_l}(\by_i).
\end{equation}
As presented in Figure \ref{TDE}, DSP adopts a sequential binary partitioning strategy to obtain the decomposition~\eqref{decom omega}.

\subsection{Setting of tree-based partitioning}

\begin{algorithm}[!h]
	\caption{The framework of tree-based density estimation.} 
	\hspace*{0.02in} {\bf Input:}
	\label{DSP alg}
	The domain $\Omega$, particle set $\by_i,\,i = 1,2,\dots,N$, parameters $\theta, m$.\\
	\hspace*{0.02in} {\bf Output:} An adaptive piecewise constant approximation of probability density function $q(\by)$.
	\begin{algorithmic}[1]
		\State $L=1, \Omega_1=\Omega, \mathcal{P}_1=\left\{\Omega_1\right\}$;
		\While{true}
		\State flag = 0;
		\ForAll{$\Omega_l=\prod\limits_{j = 1}^{d}\left[a_j^{(l)}, b_j^{(l)}\right] \in \mathcal{P}_L$}
		\State Denote $S_l = \{(y^{(l)}_{i1},\dots,y^{(l)}_{id})\}_{i=1}^n$ as the subset of particles that fall within $\Omega_l$ and $n = |S_l|$;
		\State Measure the uniformity of $S_l$;\label{D line 1}
		\If{$S_l$ cannot be approximated as a uniform distribution, e.g., $D^\ast(\widetilde{S}_l) > \frac{\theta \sqrt{N}}{n_l}$}.
		\label{D line 2}
		\State $L \leftarrow L+1$;
		\State Choose a splitting node $\widetilde{s}_{i_0, j_0}^{(l)}$ according to 
		
		\begin{equation*}
		\widetilde{s}_{i_0, j_0}^{(l)}=\underset{s_{i_0, j_0}^{(l)}}{\arg \max }\left|\frac{n_{i_0,j_0}}{n} - \frac{i_0}{m}\right|
		\end{equation*}
		
		\State Divide $\Omega_l$ into $\Omega_l^{(1)} \cup \Omega_l^{(2)}$ as
		\begin{equation*}
		\Omega_l^{(1)}=\prod_{j=1}^{j_0-1}\left[a_j^{(l)}, b_j^{(l)}\right] \times\left[a_{j_0}^{(l)}, s_{i_0, j_0}^{(l)}\right] \times \prod_{j=j_0+1}^d\left[a_j^{(l)}, 		b_j^{(l)}\right], \quad  \Omega_l^{(2)}=\Omega_l \backslash \Omega_l^{(1)}
		\end{equation*}
		\State $\Omega_l \leftarrow \Omega_l^{(1)}, \Omega_{L+1} \leftarrow \Omega_l^{(2)}$;
		\State flag = 1;
		\State Break;
		\EndIf \label{D line 3}
		\EndFor
		\If{flag = 1}
		\State $\mathcal{P}_{L+1}\leftarrow\left\{\Omega_1, \ldots, \Omega_{L+1}\right\}$; \label{D line 4}
		\Else
		\State Break;
		\EndIf
		\EndWhile
		\State Calculate a piecewise constant density estimator $\hat{p}(\by)$ corresponding to the partition $\mathcal{P}_{L}$ by 
		\begin{equation*}
		\hat{p}(\by) = \sum_{l=1}^{L} c_l \mone_{\Omega_l}(\by), \quad c_l = \frac{1}{N |\Omega_l|} \sum_{i=1}^{N} \mone_{\Omega_l}(\by_i)
		\end{equation*}
	\end{algorithmic}
\end{algorithm}
When the star discrepancy given in Eq.~\eqref{star dis} of the observations within the sub-domain $\Omega_l$ is less than a predetermined threshold, it can be approximately considered that the observations within this sub-domain are uniformly distributed, and the corresponding density function can be approximated as a constant. Algorithm~\ref{DSP alg} presents the detailed process. It can be readily seen there that if there exists a sub-domain $\Omega_l$ that fails to pass the uniformity test (see Lines~\ref{D line 2}-\ref{D line 3}), then $\Omega_l$ is split along a certain coordinate and two new sub-domains, $\Omega_l^{(1)}$ and $\Omega_l^{(2)}$,  are generated. Subsequently, these two newly generated sub-domains are combined with the remaining sub-domains to update the sub-domain set $\mathcal{P}_{L+1}$ (see Line~\ref{D line 4}).



Suppose $\Omega_l=\prod\limits_{j = 1}^{d}\left[a_j^{(l)}, b_j^{(l)}\right]$ and
$S_l = \{(y^{(l)}_{i1},\dots,y^{(l)}_{id})\}_{i=1}^n$ to be the subset of particles that fall within $\Omega_l$.
Two key issues are left to be specified in Algorithm~\ref{DSP alg}: Where to split and whether to split.

\begin{itemize}
	\item Where to split: 
	 We adopt $m-1$ dividing points along each coordinate, and choose the uniformly distributed points for simplicity:  
$s_{i_0, j_0}^{(l)} = a_{j_0}^{(l)} + \frac{i_0}{m} \left(b_{j_0}^{(l)} - a_{j_0}^{(l)}\right)$
with $i_0 = 1, 2, \dots, m - 1$ and $j_0 = 1, 2, \dots, d$. 
Accordingly, DSP splits $\Omega_l$ using the dividing point $\widetilde{s}_{i_0, j_0}^{(l)}$ that divides $\Omega_l$ into $\Omega_l^{(1)}$ and $\Omega_l^{(2)}$. Here $\widetilde{s}_{i_0, j_0}^{(l)}$ is selected according to 
	\begin{equation}\label{split c}
		\widetilde{s}_{i_0, j_0}^{(l)}=\underset{s_{i_0, j_0}^{(l)}}{\arg \max }\left|\frac{n_{i_0,j_0}}{n} - \frac{i_0}{m}\right|,
	\end{equation}
	where $n_{i_0,j_0}$ count the number of particles in $S_l$ that fall within $\Omega_l^{(1)}$,
	and 
	\begin{equation}\label{split 12}
		\begin{aligned}
			& \Omega_l^{(1)}=\prod_{j=1}^{j_0-1}\left[a_j^{(l)}, b_j^{(l)}\right] \times\left[a_{j_0}^{(l)}, s_{i_0, j_0}^{(l)}\right] \times \prod_{j=j_0+1}^d\left[a_j^{(l)}, b_j^{(l)}\right], \quad \Omega_l^{(2)}=\Omega_l \backslash \Omega_l^{(1)}.
		\end{aligned}
	\end{equation}
 The intuition behind Eq.~\eqref{split c} is to find the most non-uniform sub-domain $\Omega_l^{(1)}$ for a given set of dividing points  $\{s_{i_0, j_0}^{(l)}\big| i_0 = 1, 2, \dots, m - 1, j_0 = 1, 2, \dots, d\}$; 

	\item Whether to split: The linear scaling $\Omega_l \rightarrow \left[0,1\right]^d$ is performed first before using the discrepancy as the uniformity measurement. We scale $S_l$ to $\widetilde{S}_l = \{(\frac{y^{(l)}_{i1}-a_1^{(l)}}{b_1^{(l)}},\dots,\frac{y^{(l)}_{id}-a_d^{(l)}}{b_d^{(l)}})\}_{i=1}^n$. The partitioning terminates when the discrepancy of particles that fall within $\Omega_l$ satisfying 
	\begin{equation}\label{star stop}
		D^{\ast}(\widetilde{S}_l) \leq \frac{\theta \sqrt{N}}{n_l}.
	\end{equation}
	The parameter $\theta$ governs the depth of partition. The smaller $\theta$ is, the finer the partitioning it induces. 

\end{itemize}

\subsection{Theoretical analysis of DSP}
We can use  the $\ast$-total variation distance to measure the difference between the density estimator $\hat{p}(\by) $ and the empirical measure $\mu  = \frac{1}{N} \sum_{i=1}^N \delta_{\by_i}$. For  two measures $ \mu $ and $ \nu $, the $\ast$-total variation distance is defined as
\[
D^\ast[\mu, \nu] = \sup \left\{ |\mu(A) - \nu(A)| :  \textup{for any measurable set}~A \subset \Omega~\textup{anchored at the left corner of $\Omega$}\right\},
 \]
where $\mu(A) = \int_{\Omega} 1_A(\by) \D \mu(\by)$. A rough error estimator is given as follows.
\begin{theorem}
For the partition $\Omega = \cup_{l=1}^L \Omega_l$, suppose $D^*(\widetilde{S}_l)\leq \theta\sqrt{N}/n_l$ in each subregion $\Omega_l$,  then it has
\begin{equation}\label{TV_bound}
D^\ast[\mu, \hat{p}] \leq \frac{L \theta}{\sqrt{N}}.
\end{equation}
Thus for any function $f$ with bounded variation $V_{HK}(f; \Omega)$ in the sense of Hardy and Krause, the transport cost has an upper bound
\begin{equation}\label{HK_bound}
\Big |\int_{\Omega} f \hat{p} \D \by - \int_{\Omega} f \D \mu \Big | \le \frac{L \theta}{\sqrt{N}} V_{HK}(f; \Omega).
\end{equation}
\end{theorem}

\begin{proof}
Denote by $\#\{\by_{i}\in A\}$ the count of particles in $A$, and $\by_i^{(l)}$ the particles in the subregion $\Omega_l$. For arbitrary $A\subset\Omega$, it has
\begin{equation*}
\mu(A)=\frac{1}{N}\#\{\by_{i}\in A\}, \quad \hat{p}(A)=\sum\limits_{l=1}^{L}\frac{n_l}{N}\frac{\textup{vol}(A\cap \Omega_l)}{\textup{vol}(\Omega_l)}.
\end{equation*}

Since $A\cap \Omega_l$ is still anchored at the left corner of $\Omega_l$, it yields that
\begin{equation*}
 \begin{split}
 |\mu(A)-\nu(A)| 
 &=\left|\frac{1}{N}\#\{\by_{i}\in A\}-\sum_{l=1}^{L}\frac{n_l}{N}\frac{\textup{vol}(A\cap \Omega_l)}{\textup{vol} (\Omega_l)}\right| \\
 &= \left|\frac{1}{N} \sum_{l=1}^L \#\{\by_{i}^{(l)}\in A\cap \Omega_l\}-\sum_{l=1}^{L}\frac{n_l}{N}\frac{\textup{vol}(A\cap \Omega_l)}{\textup{vol} (\Omega_l)}\right| \\ 
&\leq\frac{1}{N}\sum_{l=1}^{L} n_l\left|\frac{\#\{\by_{i}^{(l)}\in A \cap \Omega_l \}}{n_l}-\frac{\textup{vol}(A\cap \Omega_l)}{\textup{vol} (\Omega_l)} \right| \\
&\leq\frac{1}{N}\sum_{l=1}^{L} n_l D^*(\widetilde S_l) \leq \frac{1}{N}\sum_{l=1}^{L}n_l\frac{\theta\sqrt{N}}{n_l} =\frac{L \theta}{\sqrt{N}}.
 \end{split}
\end{equation*}
By taking supremum over $A \subset \Omega$, it arrives at $D^*[\mu,\hat{p}]\leq\frac{L\theta}{\sqrt{N}}$. Eq.~\eqref{HK_bound} is a direct consequence of 
 the Koksma-Hlawka inequality.
\end{proof}

Since the partition level $L$ actually depends on $\theta$ and $N$, the above rough bound is useful only when $L$ is not so large.  In particular,  the total variation $V_{HK}(f; \Omega)$  scales linearly with $d$. For  $f(\by) = \sum_{i=1}^d y_i^2$ or $f(\by) = \sum_{j \ne i}\sum_{i=1}^d y_i y_j$, $V_{HK}(f; \Omega)$  scales quadratically with $d$. As a result, Eq.~\eqref{HK_bound} controls the discrepancy in the first and second moments of $\hat{p}$ and $\mu$. 



\subsection{Relaxation by the mixed discrepancy}

Despite that there are some efficient heuristic algorithms for calculating the star discrepancy, like the TA-improved method \cite{gnewuch2012new} and the ODE annealing approach \cite{ShaoWu2025}, the computational cost dramatically grows as sample size $N$ and dimension $d$ increase. Apart from the huge computational complexity, the star discrepancy possesses two drawbacks as follows \cite{fang2018theory}.



\begin{figure}[!h]
\centering
\subfigure[$D^* = \frac{23}{48}$, $D^{\text{mix}}=\frac{2719}{18432}$.\label{ref a}]{
\includegraphics[width=0.4\textwidth,height=0.3\textwidth]{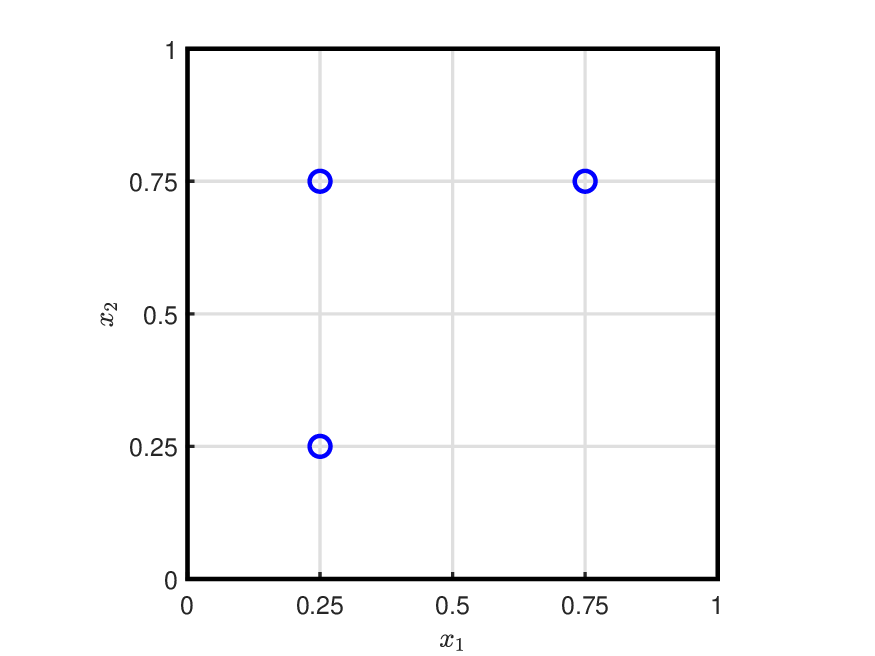}}
\subfigure[$D^* = \frac{9}{16}$, $D^{\text{mix}}=\frac{2719}{18432}$.\label{ref b}]{
\includegraphics[width=0.4\textwidth,height=0.3\textwidth]{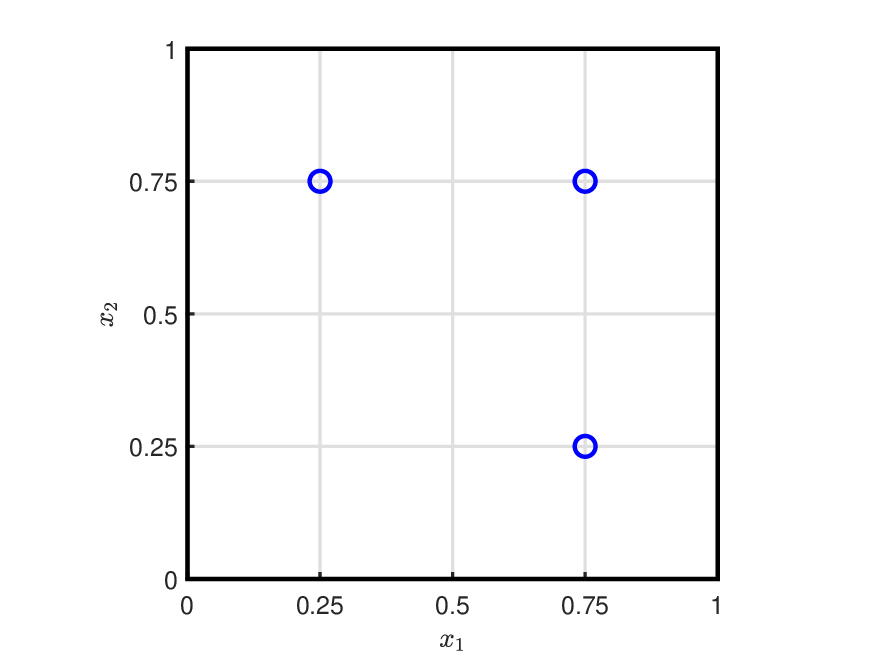}}
	\caption{Reflection invariance test, $d = 2, N = 3$: Figures~\ref{ref a} and \ref{ref b} are symmetric about $x_1 = 0.5$. Intuitively, the reflection transformation should not change the uniformity of points. However, the star discrepancies of Figures~\ref{ref a} and \ref{ref b} are different.}
	\label{reflection}
\end{figure}

\begin{figure}[!h]
\centering
\subfigure[$D^* = \frac{13}{24}, D^{\text{mix}}=\frac{301}{1024}$.\label{rot a}]{
\includegraphics[width=0.4\textwidth,height=0.3\textwidth]{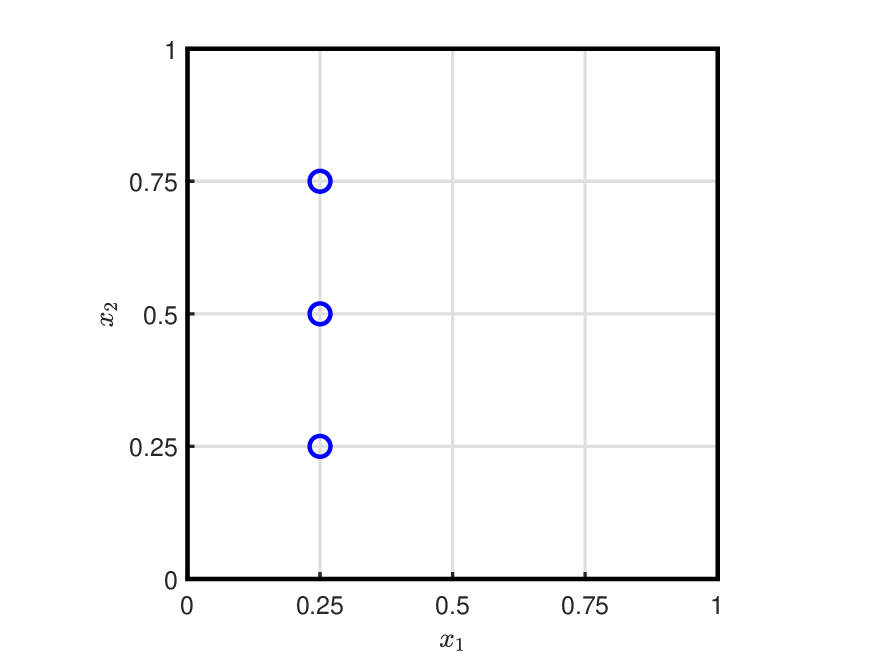}}
\subfigure[$D^* = \frac{7}{16}, D^{\text{mix}}=\frac{301}{1024}$.\label{rot b}]{
\includegraphics[width=0.4\textwidth,height=0.3\textwidth]{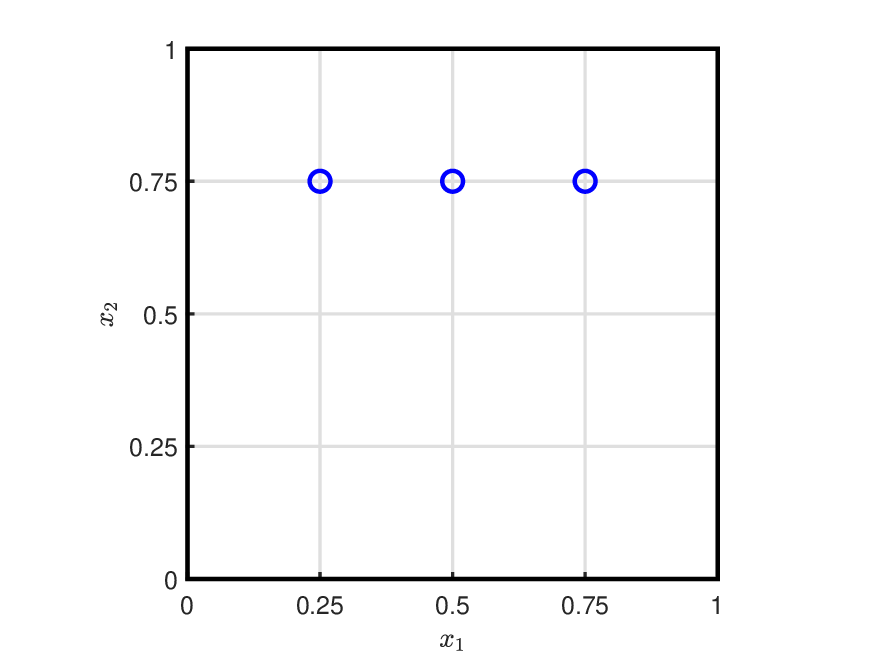}}
	\caption{Rotation invariance test, $d = 2, N = 3$: Figure~\ref{rot b} is obtained by rotating Figure~\ref{rot a} 90 degrees clockwise. Intuitively, rotation should not change the uniformity of the points. However, the star discrepancies of Figure~\ref{rot a} and \ref{rot b} are different. }
	\label{rot}
\end{figure}

\begin{enumerate}
	\item[(1)] The star discrepancy does not satisfy reflection invariance \cite{fang2018theory, zhou2013mixture}: The point sets in Figures~\ref{ref a} and \ref{ref b} are symmetric about the line $x_1 = 0.5$, but the values of their star discrepancies are not equal, which is counterintuitive. The reason for this phenomenon lies in the fact that the origin $\vec{0}$ is distinctive in the definition of the star discrepancy;
	\item[(2)] The star discrepancy does not satisfy rotation invariance \cite{fang2018theory, zhou2013mixture}: Figure~\ref{rot b} is obtained by rotating Figure~\ref{rot a} 90 degrees clockwise, and the value of the star discrepancy changes after the rotation. The reason for this phenomenon is also that the origin $\vec{0}$ is distinctive in the definition of the star discrepancy. 
\end{enumerate}

To fulfill reflection and rotation invariances, we adopt the mixed discrepancy~\eqref{dsc mix} as introduced in \cite{hickernell1998generalized,fang2018theory,zhou2013mixture}. The resulting DSP-mix inherits the algorithm framework of DSP and only differs in the measure of uniformity, where the mixture discrepancy $D^{\text{mix}}(\widetilde{S}_l)$ to adopted to measure the uniformity of $S_l$ in Line \ref{D line 1} of Algorithm \ref{DSP alg}, 
	\begin{equation}\label{star stop_2}
		 D^{\text{mix}}(\widetilde{S}_l) \leq \frac{\theta \sqrt{N}}{n_l}.
	\end{equation}
Compared with the star discrepancy adopted in DSP, 
the mixture discrepancy has the following advantages  \cite{zhou2013mixture}.
\begin{enumerate}
	\item[(1)] Easy to calculate: The mixture discrepancy can be accurately calculated with a computational complexity of $\mathcal{O}(n^2 d)$ and thus any external solver is not required. Hence DSP-mix is expected to handle the density estimation problems more efficiently than DSP. In \cite{Leihdpas2024}, DSP-mix is used to efficiently reconstruct the values of the Coulomb collision term from particles;

	\item[(2)] Reflection invariance: The mixture discrepancy satisfies the reflection invariance \cite{zhou2013mixture}. Suppose that $\left(\widetilde{\by}_1,\dots, \widetilde{\by}_n\right)$ is the points obtained by reflecting $\left(\by_1,\dots,\by_n\right)$ about the plane $y_{j_0} = 1/2$, then
	\begin{equation}
		\widetilde{\by}_i = \left(y_{i1},\dots,y_{i,j_0-1},1-y_{i,j_0},y_{i,j_0+1},\dots,y_{id}\right),\quad i = 1\dots,n,
	\end{equation} 
	and $\left(D^{\text{mix}}(\widetilde{\by}_1,\dots, \widetilde{\by}_n\right))^2	=\left(D^{\text{mix}}(\by_1, \ldots, \by_n\right))^2$. For the two point sets in Figures \ref{ref a} and \ref{ref b}, their mixture discrepancies are both $\frac{2719}{18432}$;
	
	\item[(3)] Rotation invariance: The mixture discrepancy satisfies the rotation invariance \cite{zhou2013mixture}. Suppose that $\left(\widetilde{\by}_1,\dots, \widetilde{\by}_n\right)$ is the points obtained by rotating $\left(\by_1,\dots,\by_n\right)$ 90 degrees clockwise about the $y_{i_0}-y_{j_0}$ plane, then for $\forall i = 1,\dots,n$,
	\begin{equation*}
			\widetilde{y}_{ij} =
		\begin{cases}
			y_{ij}, &  j \neq i_0, j_0, \\
			\cos\left(-\pi/2\right)(y_{i,i_0}-0.5)-\sin\left(-\pi/2\right)(y_{i,j_0}-0.5)+0.5 = y_{i,j_0}, &  j = i_0, \\
			\sin\left(-\pi/2\right)(y_{i,i_0}-0.5)+\cos\left(-\pi/2\right)(y_{i,j_0}-0.5)+0.5=1-y_{i,i_0}, &  j=j_0,
		\end{cases}
	\end{equation*}

	and it can be easily proved that $(D^{\text{mix}}\left(\widetilde{\by}_1,\dots, \widetilde{\by}_n\right))^2 = (D^{\text{mix}}\left(\by_1, \ldots, \by_n\right))^2$.
	For the two point sets in Figures \ref{rot a} and \ref{rot b}, their mixture discrepancies are both $\frac{301}{1024}$. 
\end{enumerate}


\section{MSP: Moment-based Sequential Partition}
\label{sec msp}


MSP still adopts the framework of  Algorithm \ref{DSP alg}. Instead of using the discrepancy as the uniformity measurement, MSP judges whether the particle set $S_l$ in $\Omega_l$ is close to a uniform distribution by comparing whether the moments  of the discrete point measure $\frac{1}{N} \sum_{i=1}^N \delta_{\by_i}$ and the uniform measure are close enough. 

The motivation comes from the following fact: Suppose that the $d$-dimensional random variable $\by = (y_1,y_2,\dots,y_d)$ follows the probability density $q(\by)$, the $d$-dimensional index $\vec{k} = (k_1,k_2,\dots,k_d)$, $\by^{\vec{k}} = y_1^{k_1}y_2^{k_2}\dots y_d^{k_d}$, $\mathbb{N}_0 = \{0,1,2,\dots\}$, then the moment of order $|\vec{k}| = \sum\limits_{i=1}^{d} k_i$ of the probability density $q(\by)$ is defined as 
\begin{equation}
	m_{\vec{k}}(q) = \int_{\mathbb{R}^d} \by^{\vec{k}} q(\by)\, \D \by, \quad \vec{k} \in \mathbb{N}_0^d.
\end{equation}
We call $\{m_{\vec{k}}(q), \vec{k} \in \mathbb{N}_0^d\}$ the moment sequence. Whether the probability density on a compact set can be uniquely determined by the moment sequence is the Hausdorff moment problem in probability theory, and its answer is positive \cite{Kleiber2013moment}. 
\begin{lemma}
	(\cite{Kleiber2013moment}) The probability density function defined on a compact set is uniquely determined by its moment sequence. 
\end{lemma}

As a consequence, for two probability densities defined on a compact set, we can judge whether they are close by comparing their moment sequences.  Detailedly,  for the test function $f(\by)$,
\begin{equation}
\begin{split}
&\int_{\Omega} f(\mathbf{y}) \hat{p}(\mathbf{y}) \mathrm{d} \mathbf{y} - \int_{\Omega} f(\mathbf{y}) q(\mathbf{y}) \mathrm{d} \mathbf{y} \\
=& \int_{\Omega} f(\mathbf{y}) \hat{p}(\mathbf{y}) \mathrm{d} \mathbf{y} - \frac{1}{N} \sum_{i=1}^N  f(\mathbf{y}_i) +  \frac{1}{N} \sum_{i=1}^N f(\mathbf{y}_i)  -  \int_{\Omega} f(\mathbf{y}) q(\mathbf{y}) \mathrm{d} \mathbf{y} \\
\le&  \sum_{l = 1}^L \frac{n_l}{N} \Big | \frac{1}{\textup{vol}(\Omega_l)} \int_{\Omega_l} f(\mathbf{y}) \mathrm{d} \mathbf{y}  - \frac{1}{n_l} \sum_{\mathbf{y}_i^{(l)} \in S_l} f(\mathbf{y}_i^{(l)}) \Big | \\
&+  \Big | \frac{1}{N} \sum_{i=1}^N f(\mathbf{y}_i)  -  \int_{\Omega} f(\mathbf{y}) q(\mathbf{y}) \mathrm{d} \mathbf{y} \Big |.
\end{split}
\end{equation}
Thus we can choose $f(\by) = \by^{k}$ to control the moment sequence in the density estimator,  while the second term depends only on the Monte Carlo sampling error.


In actual numerical implementation, we discover that conducting comparisons up to the order of $|\vec{k}|\leq 2$ is capable of attaining satisfactory numerical effects, that is, only the expectation and the covariance matrix are considered. Denote the expectation of the uniform measure on the sub-domain $\Omega_l$ as $\vec{\mu}^{(l)} = (\mu^{(l)}_1, \dots, \mu^{(l)}_d) \in \mathbb{R}^d$, and the covariance matrix as $\vec{\Sigma}^{(l)} = (\Sigma_{ij}^{(l)})\in \mathbb{R}^{d\times d}$. The expectation of the particle subset $S_l$ is $\hat{\vec{\mu}}^{(l)} = (\hat{\mu}^{(l)}_1, \dots, \hat{\mu}^{(l)}_d) \in \mathbb{R}^d$, and the covariance matrix is $\vec{\hat{\Sigma}}^{(l)} = (\hat{\Sigma}_{ij}^{(l)})\in \mathbb{R}^{d\times d}$. They have the explicit calculation expressions,
\begin{equation}
	\begin{aligned}
		\mu^{(l)}_j &= \frac{a_j^{(l)}+b_j^{(l)}}{2}, \quad j = 1,\dots,d,\\
		\Sigma_{jj}^{(l)} &= \frac{\left(b_j^{(l)}-a_j^{(l)}\right)^2}{12}, \quad j = 1,\dots,d, \quad \textup{and} \quad \Sigma_{ij}^{(l)} &= 0,\quad i\neq j,
	\end{aligned}
\end{equation}
and
\begin{equation}
\hat{\vec{\mu}}^{(l)} = \frac{1}{n} \sum_{i=1}^{n} \by_i^{(l)}, \quad \vec{\hat{\Sigma}}^{(l)} = \frac{1}{n} \sum_{i=1}^{n} \left(\by_i^{(l)}-\hat{\vec{\mu}}^{(l)}\right)\left(\by_i^{(l)}-\hat{\vec{\mu}}^{(l)}\right)^T.
\end{equation}

MSP runs the uniformity test as follows 
\begin{equation}\label{MSP stop}
	\begin{aligned}
		& \left|\mu_j^{(l)}-\hat{\mu}_j^{(l)}\right|<\theta\left(b_j^{(l)}-a_j^{(l)}\right), \quad j = 1,\dots,d,\\
		& \left|\Sigma_{jj}^{(l)}-\hat{\Sigma}_{jj}^{(l)}\right|<\theta\left|\Sigma_{jj}^{(l)}\right|, \quad j = 1,\dots,d, \quad \textup{and} \quad  \left|\hat{\Sigma}_{i j}^{(l)}\right|<\theta,\quad i\neq j, 
	\end{aligned}
\end{equation}
where we choose the same tolerance parameter $\theta$ in Eq.~\eqref{star stop} used by both DSP and DSP-mix for the sake of comparison.
If Eq.~\eqref{MSP stop} is satisfied, MSP terminates the sequential partition.
Otherwise we consider that $S_l$ cannot be approximated as the uniform distribution in Line \ref{D line 2} of Algorithm \ref{DSP alg}. Obviously, MSP has the following advantages over DSP. 
\begin{enumerate}
	\item[(1)] Easy to calculate: The computational complexity of the expectation is $\mathcal{O}(nd)$, and the computational complexity of the covariance matrix is $\mathcal{O}(nd^2)$. Both of them are linear functions of the number of particles $n$;
	\item[(2)] Reflection invariance: Suppose that $\left(\widetilde{\by}_1,\dots, \widetilde{\by}_n\right)$ is the points obtained by reflecting $\left(\by_1,\dots,\by_n\right)$ about the plane $y_{j_0} = \frac{a_{j_0}^{(l)}+b_{j_0}^{(l)}}{2}$, then
	\begin{equation}
		\widetilde{\by}_i = \left(y_{i1},\dots,y_{i,j_0-1},a_{j_0}^{(l)}+b_{j_0}^{(l)}-y_{i,j_0},y_{i,j_0+1},\dots,y_{id}\right),\quad i = 1\dots,n.
	\end{equation} 
	Let $\widetilde{\hat{\vec{\mu}}}^{(l)} = \frac{1}{n} \sum_{i=1}^{n} \widetilde{\by}_i^{(l)}$ and $\widetilde{\vec{\hat{\Sigma}}}^{(l)} = \frac{1}{n} \sum_{i=1}^{n} \left(\widetilde{\by}_i^{(l)}-\widetilde{\hat{\vec{\mu}}}^{(l)}\right)\left(\widetilde{\by}_i^{(l)}-\widetilde{\hat{\vec{\mu}}}^{(l)}\right)^T$, then
	\begin{equation}\label{ref mu}
		\left|\mu_j^{(l)}-\widetilde{\hat{\mu}}_j^{(l)}\right| =
		\begin{cases}
			\left|\mu_j^{(l)}-\hat{\mu}_j^{(l)}\right|, &  j \neq j_0,\\
			\left|\frac{a_j^{(l)}+b_j^{(l)}}{2}-(a_j^{(l)}+b_j^{(l)})+\hat{\mu}_j^{(l)}\right| = \left|\mu_j^{(l)}-\hat{\mu}_j^{(l)}\right|, &  j = j_0.
		\end{cases}
	\end{equation}
	For $j \neq j_0$, it is obvious that $\left|\Sigma_{jj}^{(l)}-\widetilde{\hat{\Sigma}}_{jj}^{(l)}\right| = \left|\Sigma_{jj}^{(l)}-\hat{\Sigma}_{jj}^{(l)}\right|$. For $j = j_0$, we have
	\begin{equation}\label{ref sigma jj}
		\begin{aligned}
			\left|\Sigma_{j_0j_0}^{(l)}-\widetilde{\hat{\Sigma}}_{j_0j_0}^{(l)}\right| =& \left|\Sigma_{j_0j_0}^{(l)}-\frac{1}{n}\sum_{i=1}^{n}\left(a_{j_0}^{(l)}+b_{j_0}^{(l)}-y_{i,j_0}-\frac{1}{n}\sum_{k=1}^{n}(a_{j_0}^{(l)}+b_{j_0}^{(l)}-y_{k,j_0})\right)^2\right| \\
			=& \left|\Sigma_{j_0j_0}^{(l)}-\frac{1}{n}\sum_{k=1}^{n}\left(y_{i,j_0}-\frac{1}{n}\sum_{k=1}^{n}y_{k,j_0}\right)^2\right| = \left|\Sigma_{j_0j_0}^{(l)}-\hat{\Sigma}_{j_0j_0}^{(l)}\right|.
		\end{aligned}
	\end{equation}
	For $i \neq j_0$, $j\neq j_0$ and $i\neq j$, it is patently clear that $\left|\widetilde{\hat{\Sigma}}_{i j}^{(l)}\right|=\left|\hat{\Sigma}_{i j}^{(l)}\right|$. When $i \neq j_0$ and $j=j_0$,
	\begin{equation}\label{ref sigma ij}
		\begin{aligned}
			\left|\widetilde{\hat{\Sigma}}_{i j_0}^{(l)}\right| 
			=& \left|\frac{1}{n}\sum_{k=1}^{n} \left(y_{k,i}-\frac{1}{n}\sum_{h=1}^{n} y_{h,i}\right)\left(a_{j_0}^{(l)}+b_{j_0}^{(l)}-y_{k,j_0}-\frac{1}{n}\sum_{h=1}^{n}(a_{j_0}^{(l)}+b_{j_0}^{(l)}-y_{h,j_0})\right) \right| \\
			=&\left|\frac{1}{n}\sum_{k=1}^{n} -\left(y_{k,i}-\frac{1}{n}\sum_{h=1}^{n} y_{h,i}\right)\left(y_{k,j_0}-\frac{1}{n}\sum_{h=1}^{n}y_{h,j_0}\right) \right| 
			=\left|\hat{\Sigma}_{i j}^{(l)}\right|.
		\end{aligned}
	\end{equation}
	The situation with $i= j_0$ and $j \neq j_0$ is analogous. Combining Eqs.~\eqref{ref mu}, \eqref{ref sigma jj} and \eqref{ref sigma ij} together, the reflection transformation has no impact on the termination criterion \eqref{MSP stop}. 
	For Figures \ref{ref a} and \ref{ref b}, the differences between the expectations and covariance matrices of the two groups of particles and those of the uniform measure are 
	\begin{equation}
		\begin{aligned}
			\left|\vec{\mu}^{(l)}-\hat{\vec{\mu}}^{(l)}\right| &= \left(\frac{1}{12},\frac{1}{12}\right), \quad \left|\vec{\Sigma}^{(l)}-\vec{\hat{\Sigma}}^{(l)}\right| &= 
			\begin{pmatrix}
				\frac{1}{48} & \frac{1}{48} \\
				\frac{1}{48} & \frac{1}{48}
			\end{pmatrix},
		\end{aligned}
	\end{equation}
	where the absolute is performed element-wise;
	
	\item[(3)] Rotation invariance: Suppose that $\left(\widetilde{\by}_1,\dots, \widetilde{\by}_n\right)$ is the points obtained by rotating $\left(\by_1,\dots,\by_n\right)$ 90 degrees clockwise about the $y_{i_0}-y_{j_0}$ plane with the center of $\Omega_l$ as the center of rotation, then for $\forall i = 1,\dots,n$,
	\begin{equation*}
		\begin{aligned}
			\widetilde{y}_{ii_0}=&\cos\left(-\frac{\pi}{2}\right)\left(y_{i,i_0}-\frac{a_{i_0}^{(l)}+b_{i_0}^{(l)}}{2}\right)-\sin\left(-\frac{\pi}{2}\right)\left(y_{i,j_0}-\frac{a_{j_0}^{(l)}+b_{j_0}^{(l)}}{2}\right)+\frac{a_{j_0}^{(l)}+b_{j_0}^{(l)}}{2} \\
			=& y_{i,j_0},   \\
			\widetilde{y}_{ij_0} =&\sin\left(-\frac{\pi}{2}\right)\left(y_{i,i_0}-\frac{a_{i_0}^{(l)}+b_{i_0}^{(l)}}{2}\right)+\cos\left(-\frac{\pi}{2}\right)\left(y_{i,j_0}-\frac{a_{j_0}^{(l)}+b_{j_0}^{(l)}}{2}\right)+\frac{a_{i_0}^{(l)}+b_{i_0}^{(l)}}{2}\\
			=&a_{i_0}^{(l)}+b_{i_0}^{(l)}-y_{i,i_0}, 
		\end{aligned}
	\end{equation*}
	and $\widetilde{y}_{ij}  = y_{ij}$, for $j \neq i_0 \,\text{and}\, j \neq j_0$.
	As is readily apparent,
	\begin{equation}\label{rot mu}
		\left|\mu_j^{(l)}-\widetilde{\hat{\mu}}_j^{(l)}\right| =
		\begin{cases}
			\left|\mu_j^{(l)}-\hat{\mu}_j^{(l)}\right|, &  j \neq i_0 \,\text{and}\, j \neq j_0, \\
			\left|\mu_{j_0}^{(l)}-\hat{\mu}_{j_0}^{(l)}\right|, &  j = i_0, \\
			\left|\mu_{i_0}^{(l)}-\hat{\mu}_{i_0}^{(l)}\right|, &  j=j_0,
		\end{cases}
	\end{equation}
	\begin{equation}\label{rot sigma ii}
		\left|\Sigma_{jj}^{(l)}-\widetilde{\hat{\Sigma}}_{jj}^{(l)}\right| =
		\begin{cases}
			\left|\Sigma_{jj}^{(l)}-\hat{\Sigma}_{jj}^{(l)}\right|, &  j \neq i_0 \,\text{and}\, j \neq j_0, \\
			\left|\Sigma_{j_0j_0}^{(l)}-\hat{\Sigma}_{j_0j_0}^{(l)}\right|, &  j = i_0, \\
			\left|\Sigma_{i_0i_0}^{(l)}-\hat{\Sigma}_{i_0i_0}^{(l)}\right|, &  j=j_0.
		\end{cases}
	\end{equation}
	When $i = i_0$ and $j=j_0$,
	\begin{equation}\label{rot sigma ij}
		\begin{aligned}
			\left|\widetilde{\hat{\Sigma}}_{i_0 j_0}^{(l)}\right| 
			=& \left|\frac{1}{n}\sum_{k=1}^{n} \left(y_{k,j_0}-\frac{1}{n}\sum_{h=1}^{n} y_{h,j_0}\right)\left(a_{i_0}^{(l)}+b_{i_0}^{(l)}-y_{k,i_0}-\frac{1}{n}\sum_{h=1}^{n}(a_{i_0}^{(l)}+b_{i_0}^{(l)}-y_{h,i_0})\right) \right| \\
			=&\left|\frac{1}{n}\sum_{k=1}^{n} -\left(y_{k,j_0}-\frac{1}{n}\sum_{h=1}^{n} y_{h,j_0}\right)\left(y_{k,i_0}-\frac{1}{n}\sum_{h=1}^{n}y_{h,i_0}\right) \right| 
			=\left|\hat{\Sigma}_{i_0 j_0}^{(l)}\right|.
		\end{aligned}
	\end{equation}
	For other cases with $i \neq j$, it can also be similarly deduced that $\left|\widetilde{\hat{\Sigma}}_{i j}^{(l)}\right| = \left|\hat{\Sigma}_{ij}^{(l)}\right|$. Therefore,  according to Eqs.~\eqref{rot mu}-\eqref{rot sigma ij}, we have that the rotation transformation has no impact on the termination criterion \eqref{MSP stop}. 
	For Figure \ref{rot a}, the differences between the expectations and covariances of the two groups of particles and those of the uniform measure are 
	\begin{equation}\label{mom rota}
		\begin{aligned}
			\left|\vec{\mu}^{(l)}-\hat{\vec{\mu}}^{(l)}\right| &= \left(0.25,0\right), \quad \left|\vec{\Sigma}^{(l)}-\vec{\hat{\Sigma}}^{(l)}\right| &= 
			\begin{pmatrix}
				\frac{5}{48} & 0 \\
				0 & \frac{1}{24}
			\end{pmatrix}.
		\end{aligned}
	\end{equation}
	For Figure \ref{rot b}, the differences are
	\begin{equation}\label{mom rotb}
		\begin{aligned}
			\left|\vec{\mu}^{(l)}-\hat{\vec{\mu}}^{(l)}\right| &= \left(0,0.25\right), \quad \left|\vec{\Sigma}^{(l)}-\vec{\hat{\Sigma}}^{(l)}\right| &= 
			\begin{pmatrix}
				\frac{1}{24} & 0 \\
				0 & \frac{5}{48}
			\end{pmatrix}.
		\end{aligned}
	\end{equation}
	Obviously, Eqs.~\eqref{mom rota} and \eqref{mom rotb} merely involve a rearrangement of elements and thus does not affect the termination criterion~\eqref{MSP stop}. That is, the origin $\vec{0}$ in MSP is indistinctive, and there will be no such unreasonable phenomena of the star discrepancy in Figures \ref{reflection} and \ref{rot}.  
\end{enumerate}

\section{Numerical experiments}
\label{sec:num}


\par This section conducts numerical experiments to test the performance of tree-based density estimation. Since a thorough comparative study between DSP and KDE has been performed in \cite{Li2016}, we mainly focus on the performance of DSP-mix and MSP compared with DSP. In Eq.~\eqref{split c}, the parameter $m$ specifying the number of candidate splitting points is set to $64$. To evaluate the performance of DSP, DSP-mix and MSP for high-dimensional datasets, we adopt two standard error metrics as in \cite{Li2016}. 
The first is the Kullback-Leibler (KL) divergence to quantify the similarity between two probability distributions,
\begin{equation}
\textup{KL}( p^\text{ref} || p^\text{num} ) = \int p^\text{ref}( \bm{x}) \log \left(\frac{p^\text{ref}( \bm{x})}{p^\text{num}( \bm{x})}\right) \mathrm{d} \bm{x},
\end{equation}
which can be evaluated by the arithmetic mean of the points $(\bx_1, \bx_2, \dots, \bx_N) \sim p^\text{ref}$, namely, 
 \begin{equation}
\textup{KL}( p^\text{ref} || p^\text{num} ) \approx \frac{1}{N} \sum_{i=1}^N  \left( \log p^\text{ref}( \bm{x}_i) - \log p^\text{num}( \bm{x}_i)\right).
\end{equation}

The second metric is the Hellinger distance,
\begin{equation}
{H}( p^\text{ref}, p^\text{num}) = \frac{1}{2} \int \left(\sqrt{p^\text{ref}(\bm{x})} - \sqrt{p^\text{num}(\bm{x})}\right)^2 \D \bx = 1 - \int \sqrt{\frac{p^\text{num}(\bm{x})}{p^\text{ref}(\bm{x})}} p^\text{ref}(\bm{x}) \D \bx,
\end{equation}
which can be approximated by 
\begin{equation}
\hat{H}( p^\text{ref}, p^\text{num}) \approx \hat{H}^2 = 1 - \frac{1}{N} \sum_{i=1}^N \sqrt{\frac{p^\text{num}(\bm{x}_i)}{p^\text{ref}(\bm{x}_i)}}.
\end{equation}
For each setting of parameters, we run 10 simulations from independent sampling point sets and measure the standard derivation 
\begin{equation}
\textup{SD} = \sqrt{\frac{1}{10}\sum_{i=1}^{10} \left(\textup{KL}^{(i)}\right)^2 -  \left( \frac{1}{10}\sum_{i=1}^{10} \textup{KL}^{(i)}\right)^2},
\end{equation}
or 
\begin{equation}
\textup{SD} = \sqrt{\frac{1}{10}\sum_{i=1}^{10} \left( \hat{H}^{(i)}\right)^2 -  \left( \frac{1}{10}\sum_{i=1}^{10} \hat{H}^{(i)}\right)^2}.
\end{equation}
In practice, we record the maximal values of the standard derivations produced by DSP, DSP-mix and MSP (maxSD for short) to verify their robustness. 

All simulations via C++ implementations run on the platform: AMD Ryzen 9950x (16 cores, 32 threads) with 256GB Memory (5600Mhz). The average computational time in different scenarios is recorded in Tables \ref{beta_K_time_N1e7} and \ref{beta_K_time_high} (Beta mixtures), 
Tables \ref{hdGauss_K_time_N1e7} and \ref{hdGauss_K_time_high}(Gaussian mixtures) as well as  Tables \ref{hdCauchy_K_time_N1e7} and \ref{hdCauchy_K_time_high} (Cauchy mixtures). When the sample size is relatively small ($N \le 10^6$), the computational costs for three methods are comparable. However, for large sample size ($N \ge 10^7$), both DSP-mix and MSP are evidently faster than DSP, and MSP even gains an increase in speed by a factor of two to twenty. Moreover, it is observed that the computational costs of both DSP-mix and MSP scale nearly linearly with respect to the sample size.

\subsection{$d$-D Beta mixtures}

\begin{figure}[!h]
\centering
\subfigure[True density and samples.]{
\includegraphics[width=0.4\textwidth,height=0.3\textwidth]{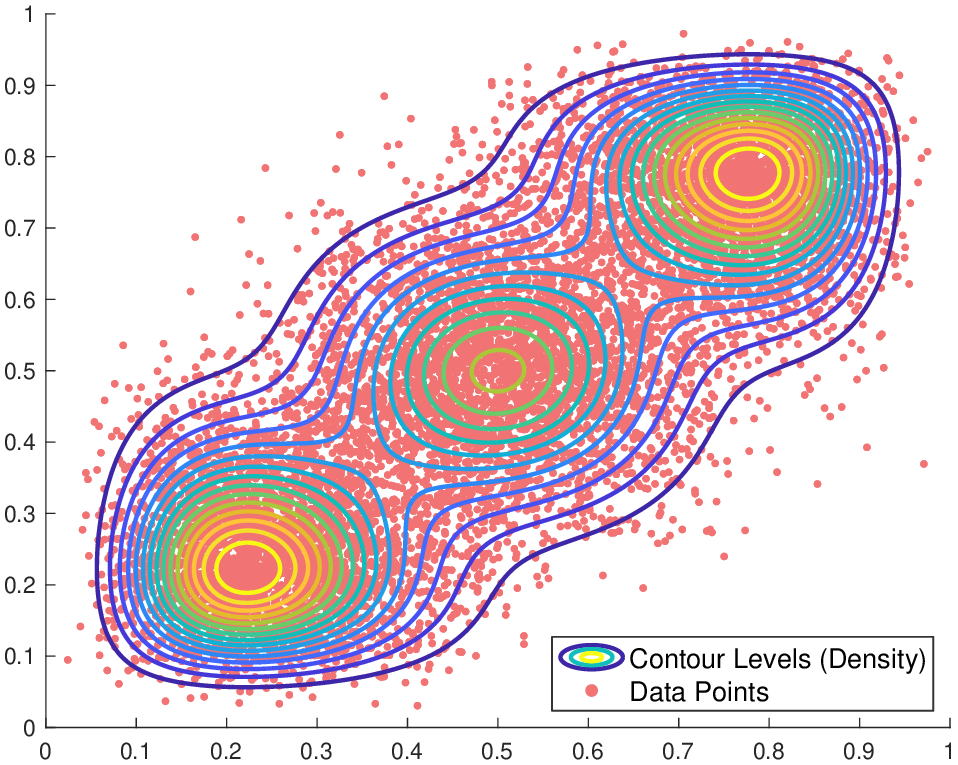}}
\\
\centering
\subfigure[Adaptive partition (left: DSP, middle: DSP-mix, right: MSP).]{
\includegraphics[width=0.32\textwidth,height=0.24\textwidth]{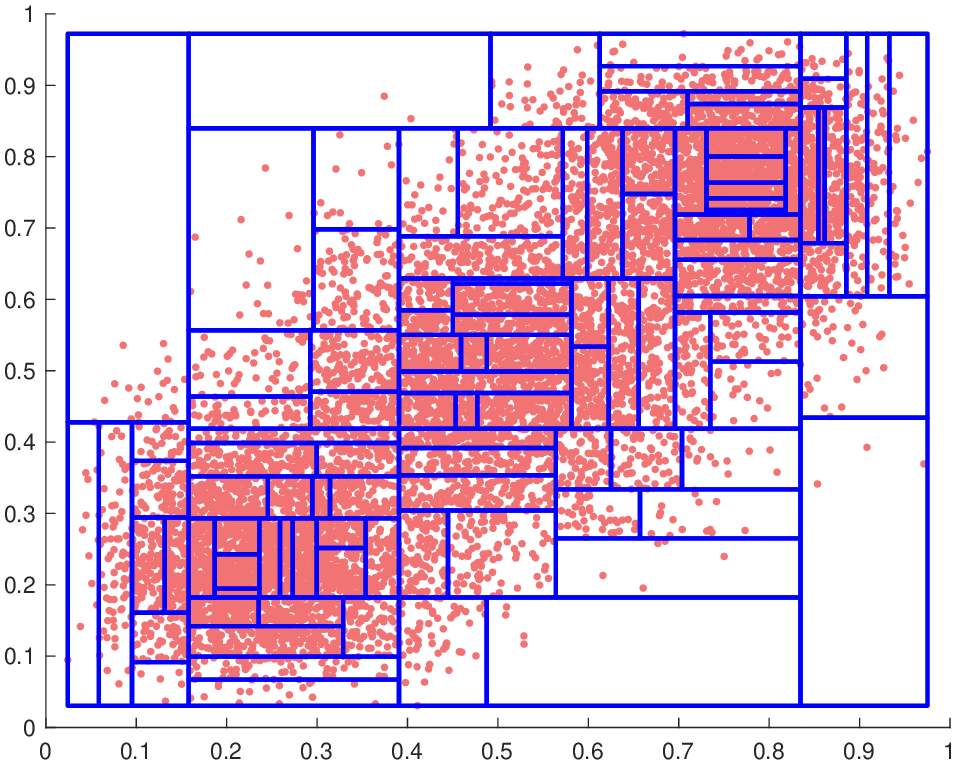}
\includegraphics[width=0.32\textwidth,height=0.24\textwidth]{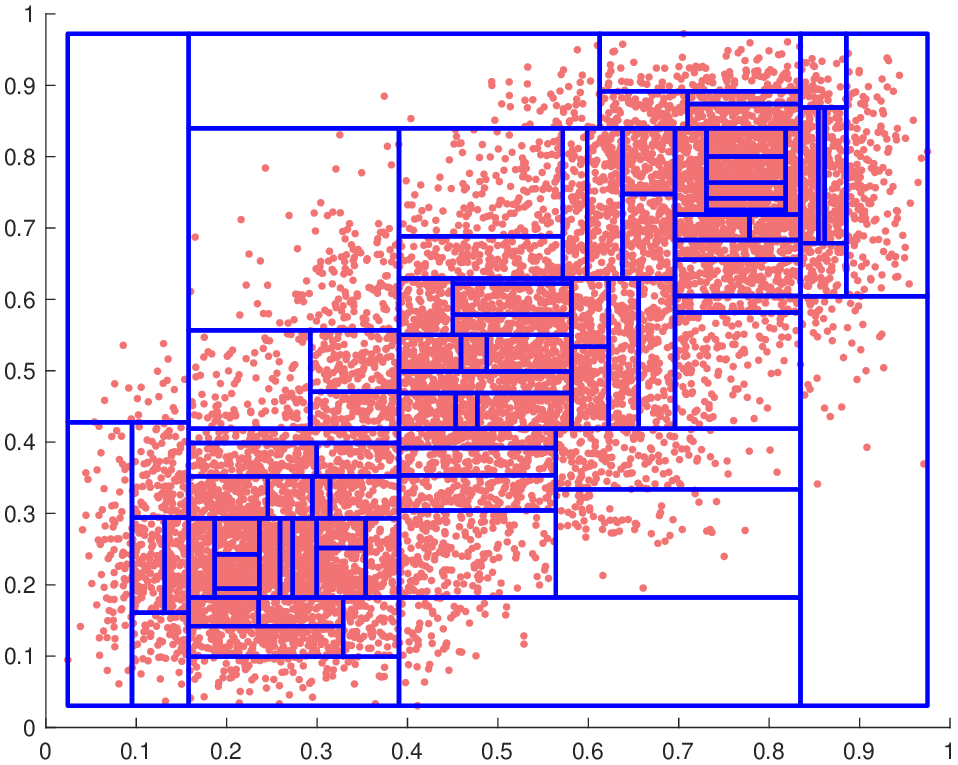}
\includegraphics[width=0.32\textwidth,height=0.24\textwidth]{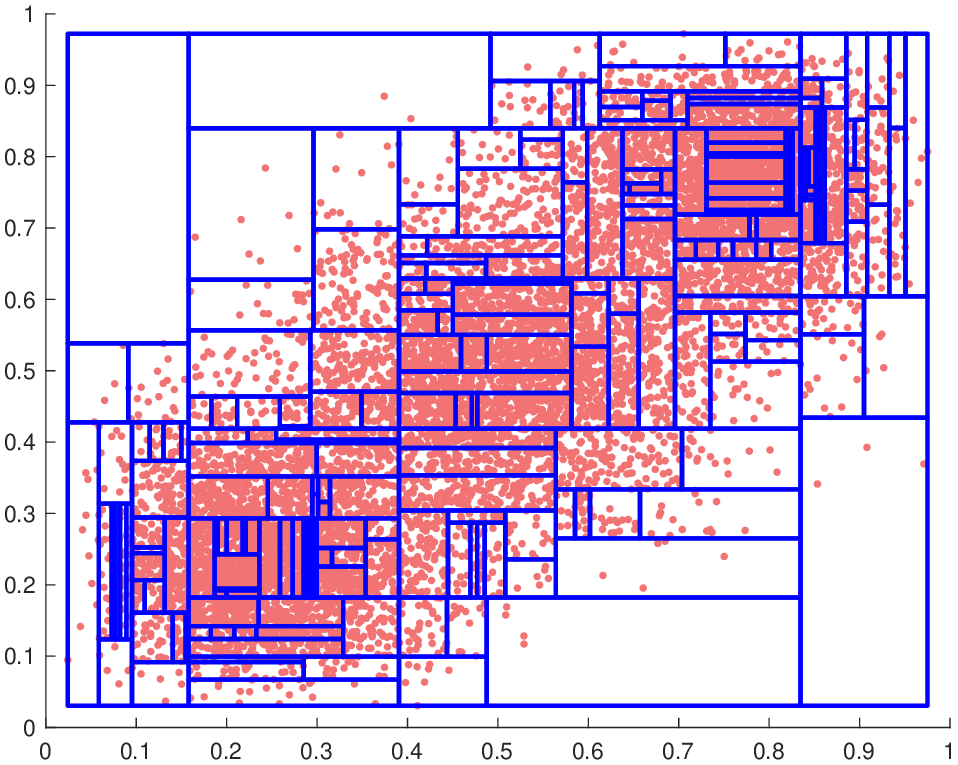}}
\\
\centering
\subfigure[Density estimation (left: DSP, middle: DSP-mix, right: MSP).]{
\includegraphics[width=0.32\textwidth,height=0.24\textwidth]{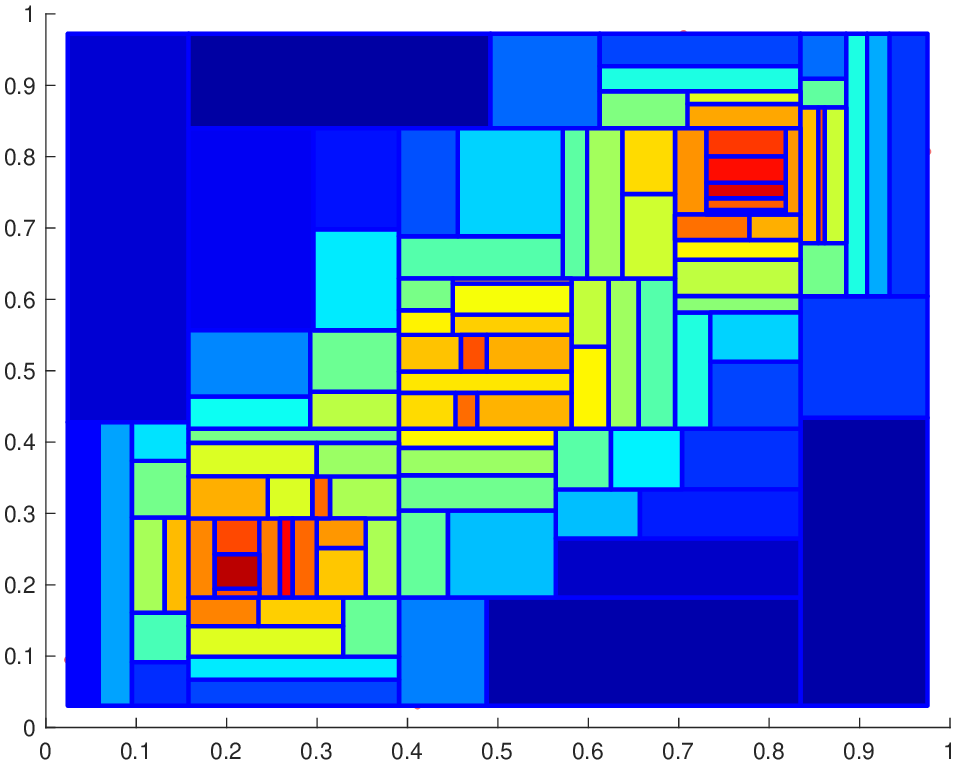}
\includegraphics[width=0.32\textwidth,height=0.24\textwidth]{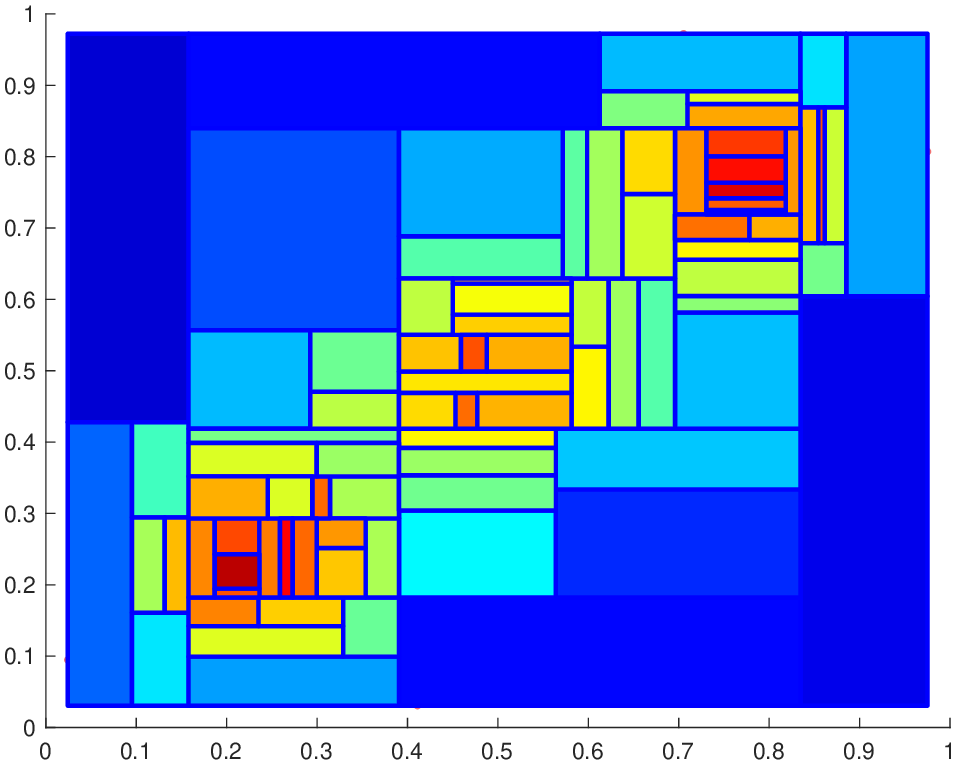}
\includegraphics[width=0.32\textwidth,height=0.24\textwidth]{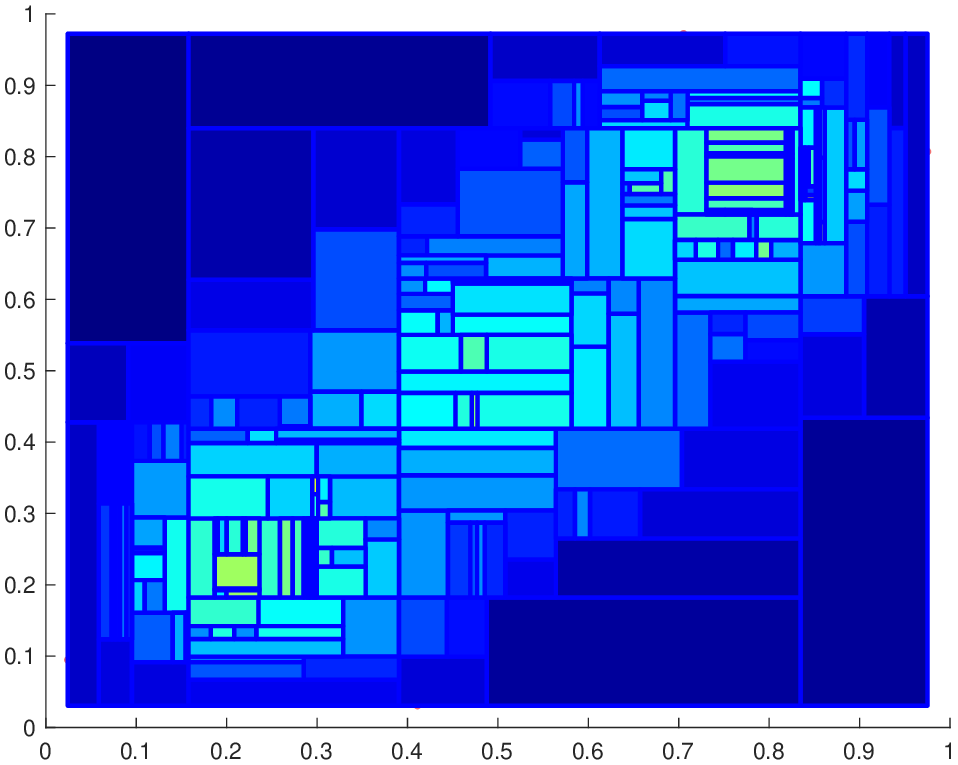}}
	\caption{2-D Beta mixtures: Adaptive partitions and density estimators produced by DSP-mix, MSP and DSP with $N = 1\times 10^4$ and $\theta=0.2$.}
	\label{2dbetamix fig}
\end{figure}

First, we examine the $d$-D mixtures of betas,
\begin{equation}
	\bx \sim  \frac{1}{3}\left(\prod_{j=1}^d  \operatorname{beta}(15,5)(x_j) + \prod_{j=1}^d  \operatorname{beta}(10,10)(x_j) + \prod_{j=1}^d  \operatorname{beta}(5,15)(x_j)\right).
\end{equation}
Figure~\ref{2dbetamix fig} visualizes this example for $d=2$. All three methods capture the main features ( three peaks) of the distribution. 

\begin{figure}[!h]
\centering
\subfigure[KL divergence ($d=2$) (left: DSP, middle: DSP-mix, right: MSP). ]{
{\includegraphics[width=0.32\textwidth,height=0.24\textwidth]{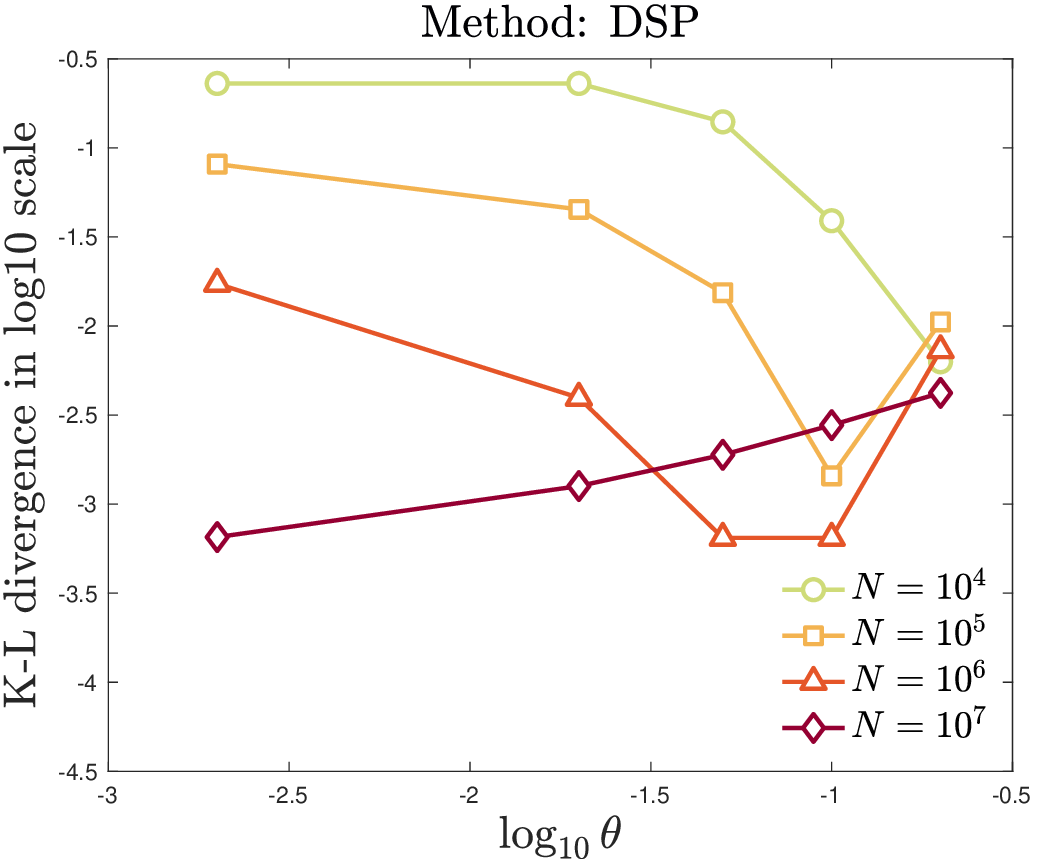}}
{\includegraphics[width=0.32\textwidth,height=0.24\textwidth]{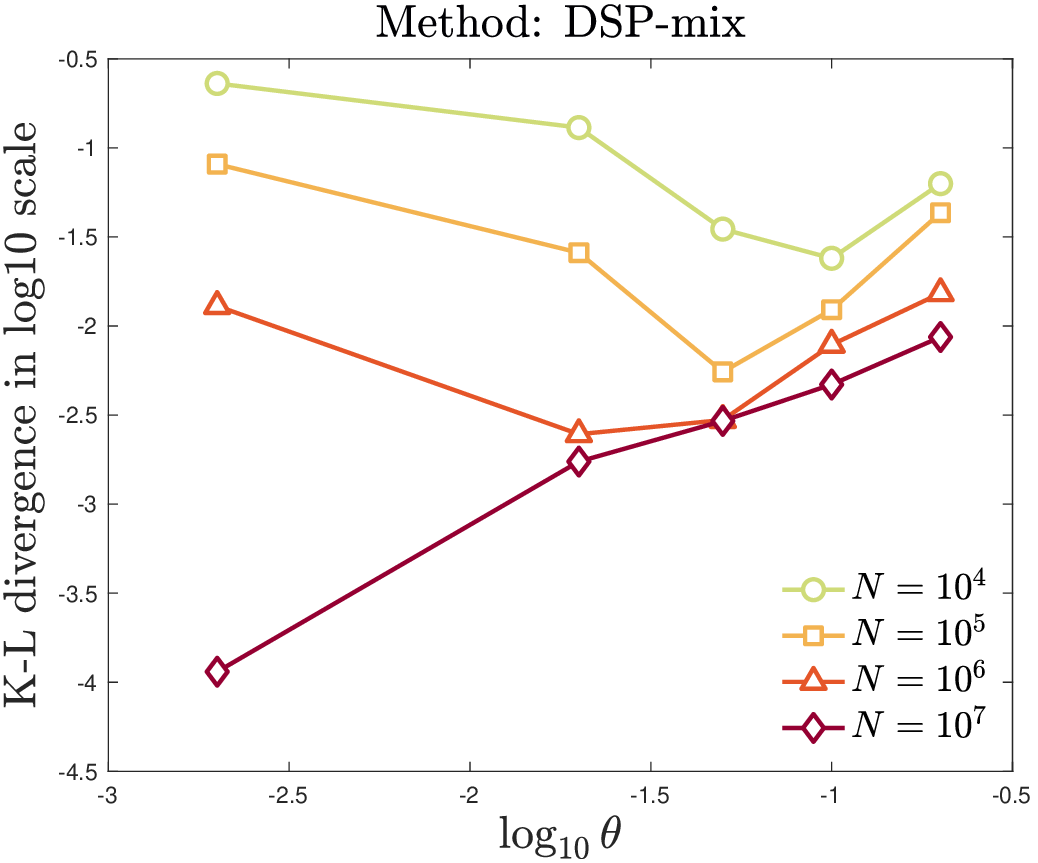}}
{\includegraphics[width=0.32\textwidth,height=0.24\textwidth]{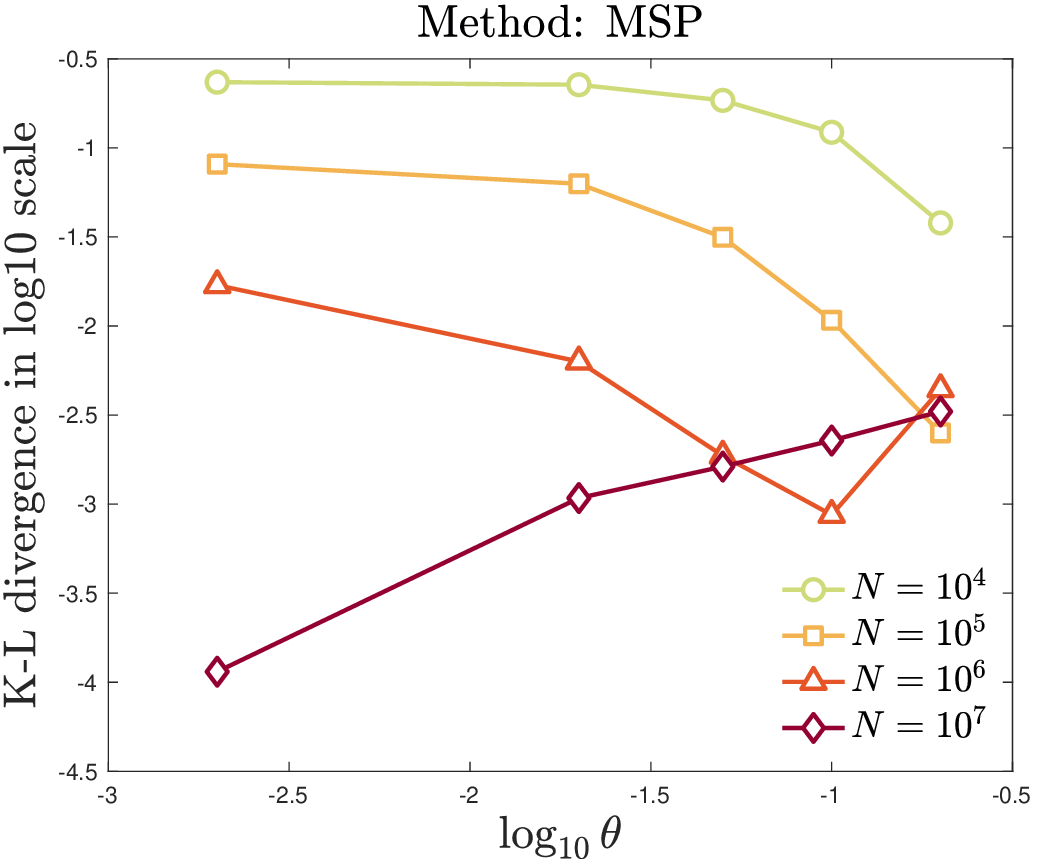}}}
\\
\centering
\subfigure[Hellinger distance ($d=2$) (left: DSP, middle: DSP-mix, right: MSP). ]{
{\includegraphics[width=0.32\textwidth,height=0.24\textwidth]{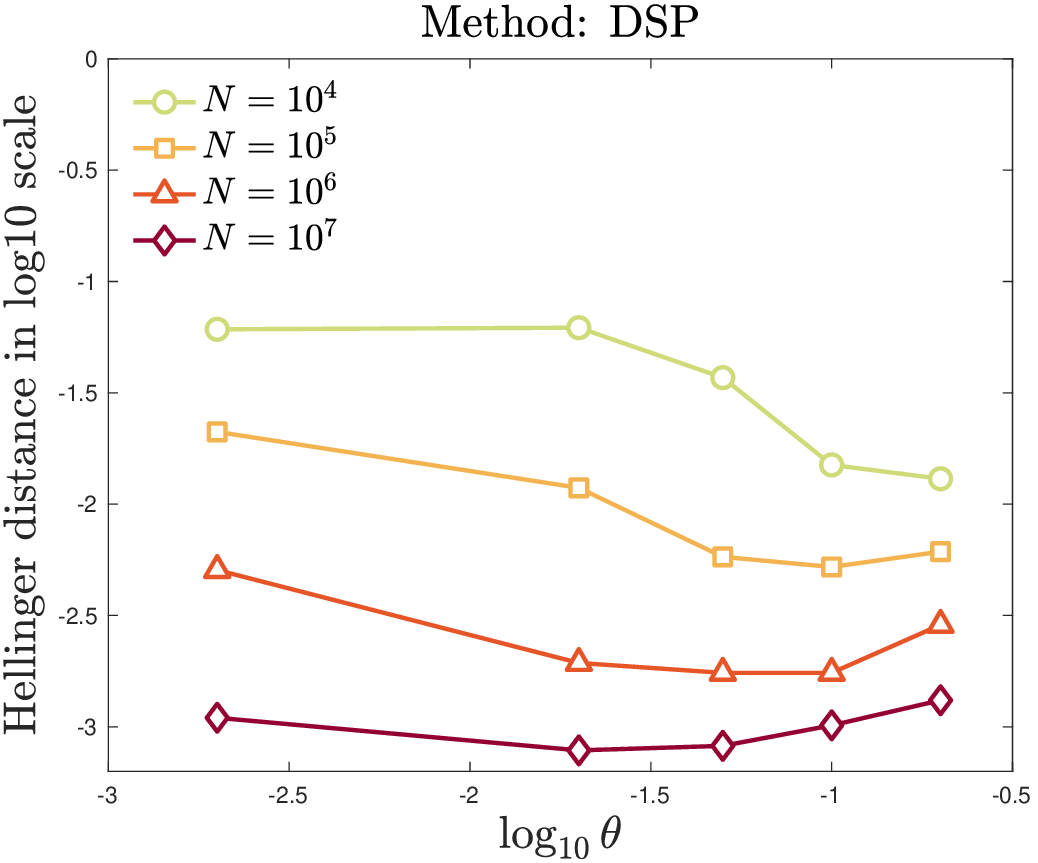}}
{\includegraphics[width=0.32\textwidth,height=0.24\textwidth]{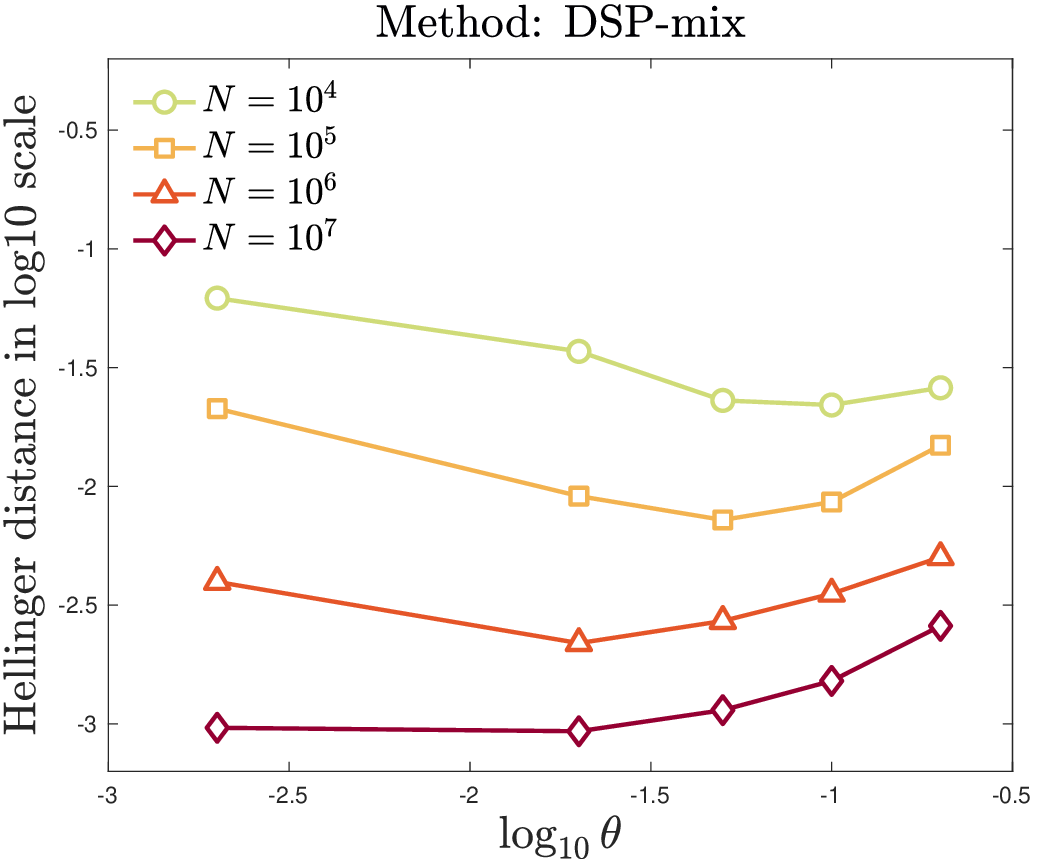}}
{\includegraphics[width=0.32\textwidth,height=0.24\textwidth]{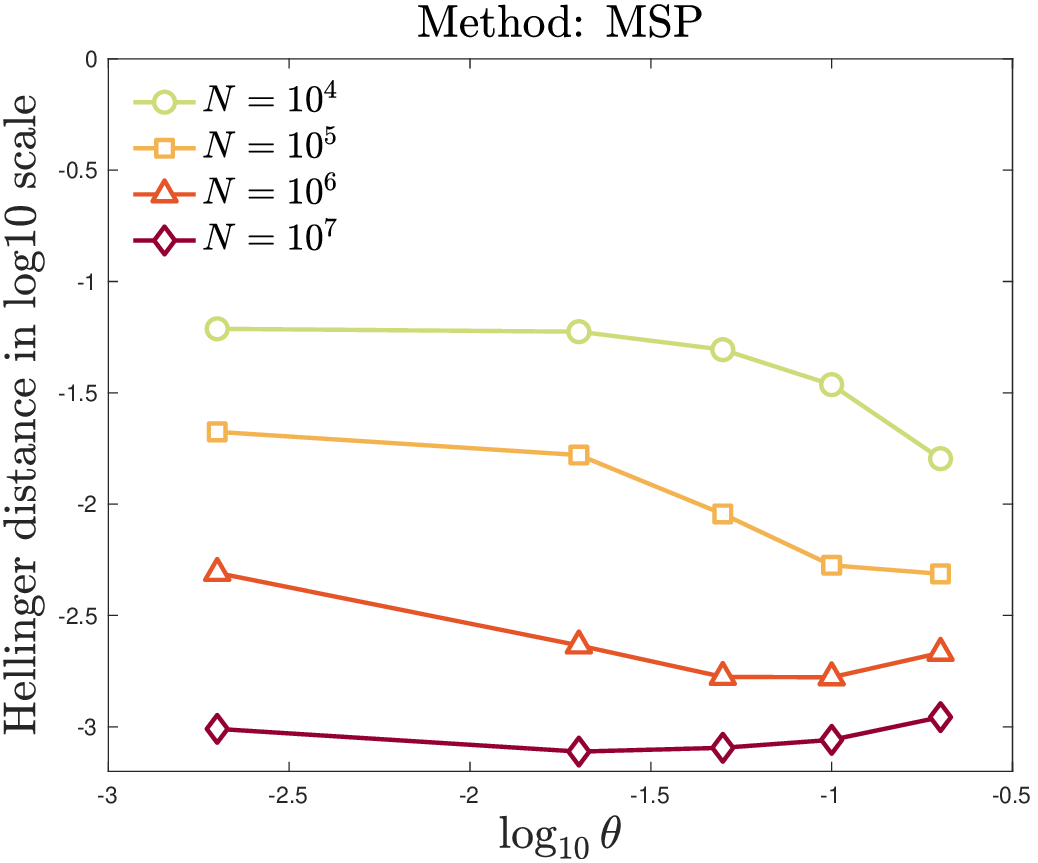}}}
\caption{\small $2$-D Beta mixtures: The KL divergence and Hellinger distance under different $N$ and $\theta$.}
\label{beta_low}
\end{figure}
In Figure~\ref{beta_low}, it is shown that the errors of the three methods are comparable, all decreasing as sample size $N$ increases. For fixed $N$, when the parameter $\theta$ decreases within a certain range, we can obtain finer partitioning and more accurate density estimation results. However, when $\theta$ goes too small, the error will increase instead, which is an indicator of the overfitting problem. Fortunately, the overfitting can be alleviated using a larger sample size.



\begin{figure}[H]
\centering
\subfigure[KL divergence (left: DSP, middle: DSP-mix, right: MSP). ]{
{\includegraphics[width=0.32\textwidth,height=0.24\textwidth]{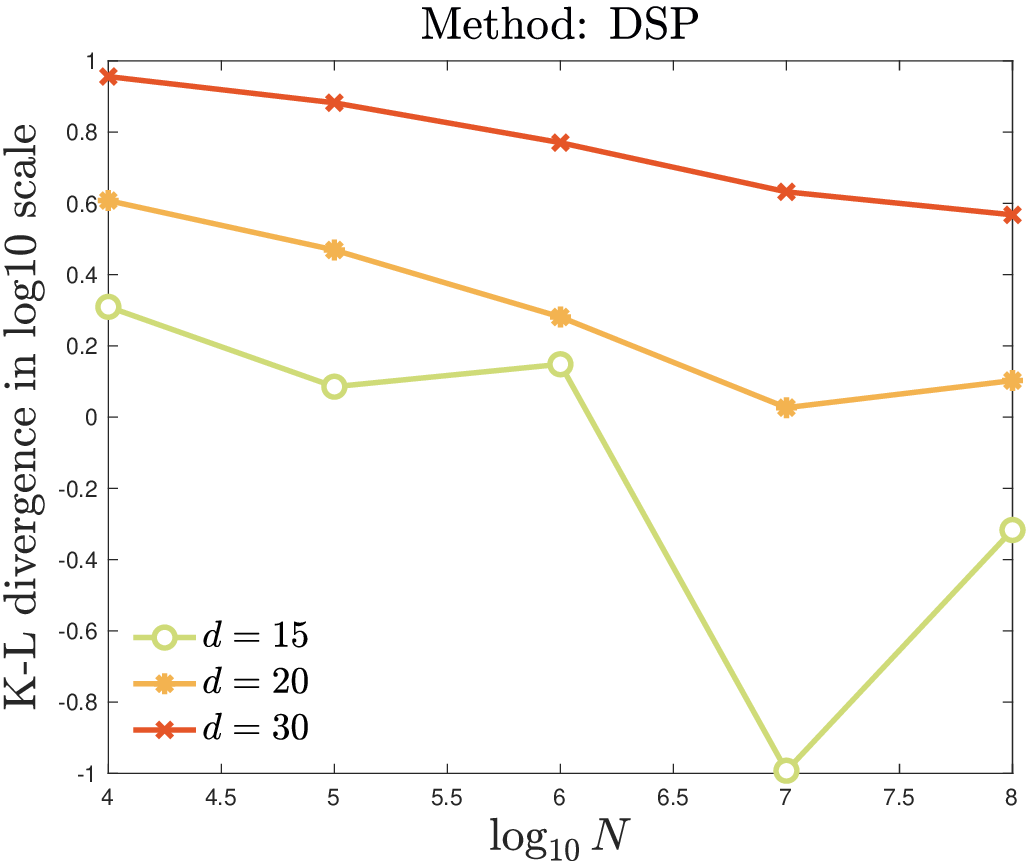}}
{\includegraphics[width=0.32\textwidth,height=0.24\textwidth]{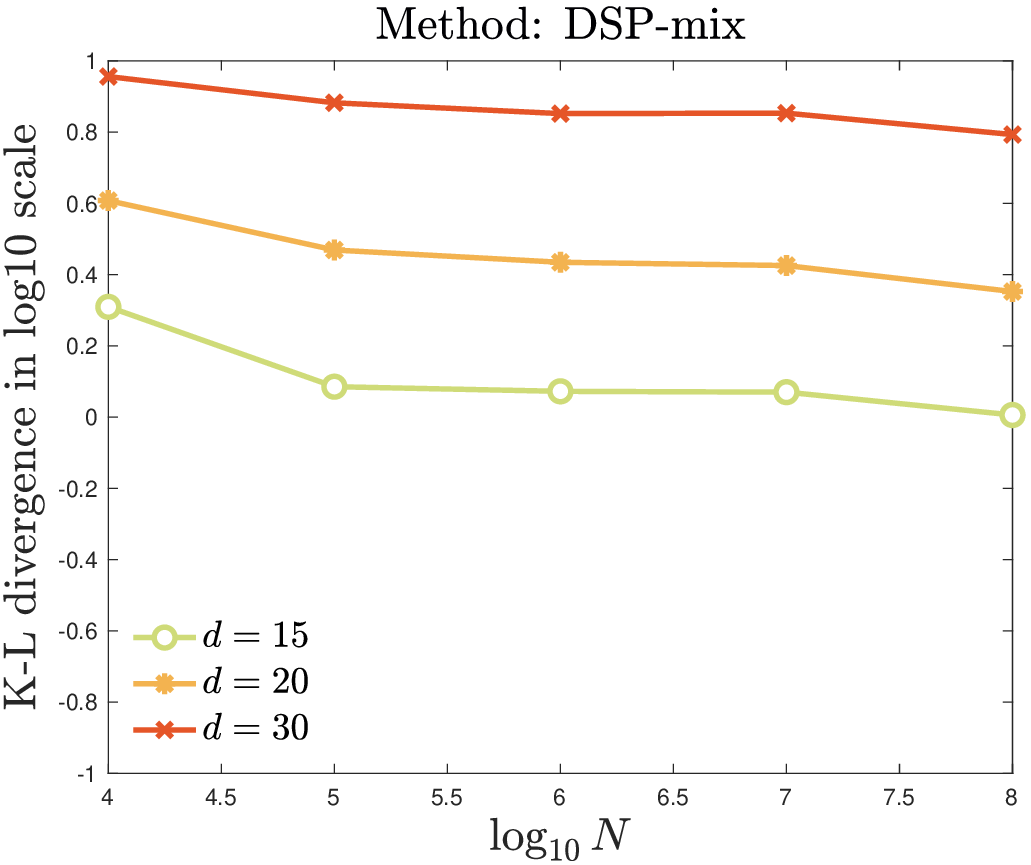}}
{\includegraphics[width=0.32\textwidth,height=0.24\textwidth]{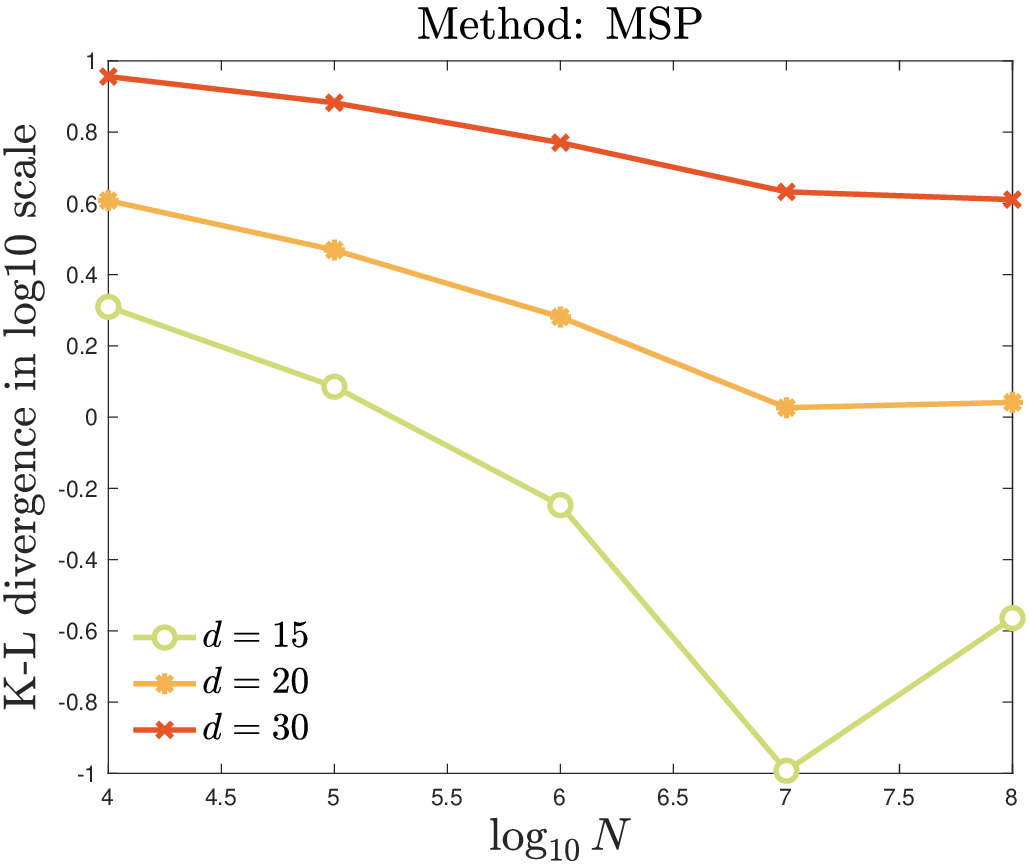}}}
\\
\centering
\subfigure[Hellinger distance (left: DSP, middle: DSP-mix, right: MSP). ]{
{\includegraphics[width=0.32\textwidth,height=0.24\textwidth]{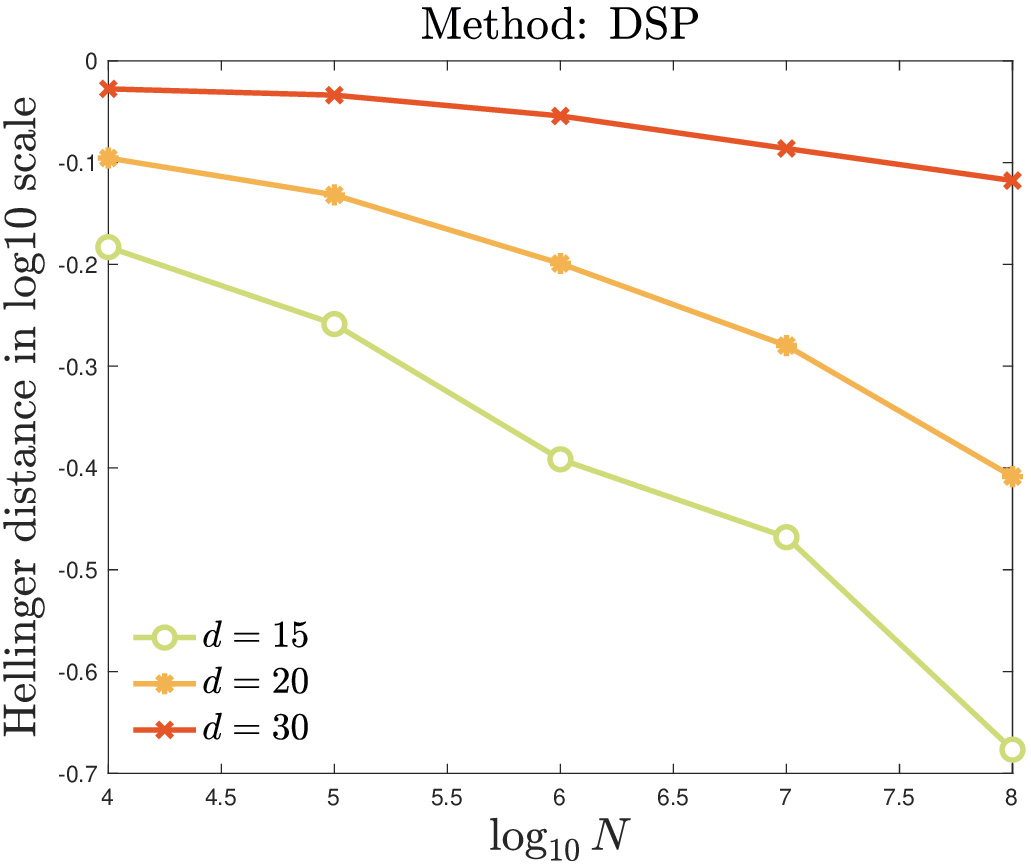}}
{\includegraphics[width=0.32\textwidth,height=0.24\textwidth]{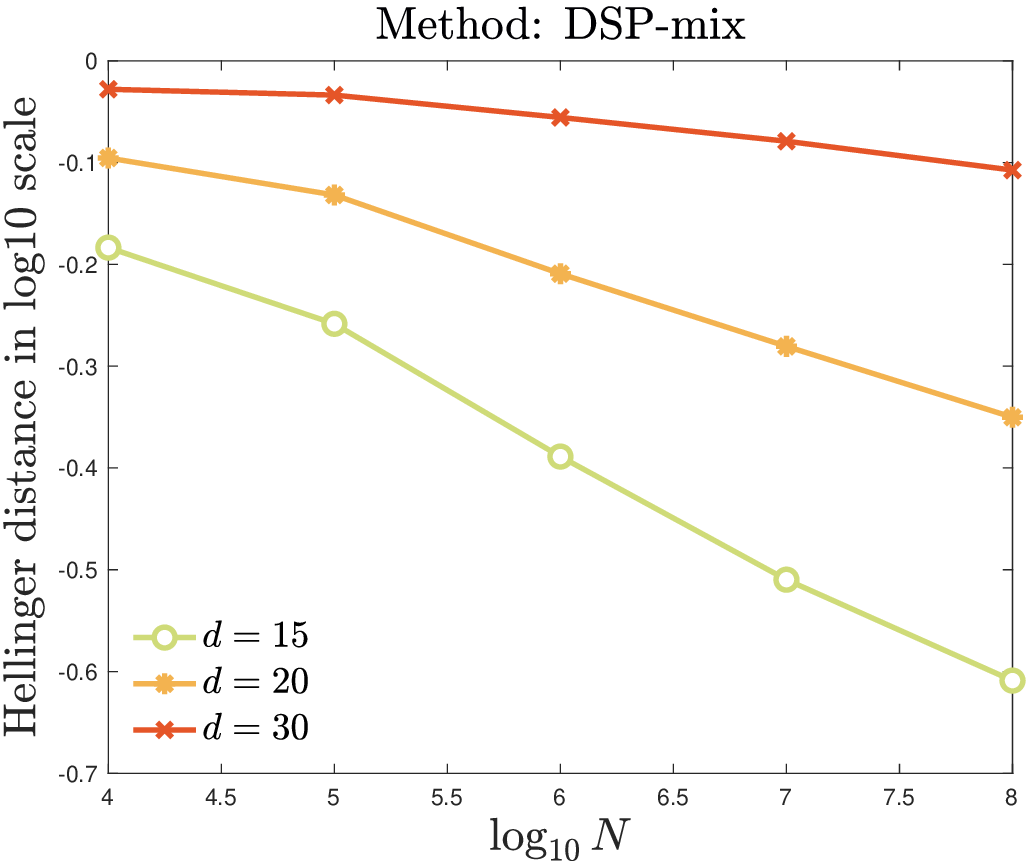}}
{\includegraphics[width=0.32\textwidth,height=0.24\textwidth]{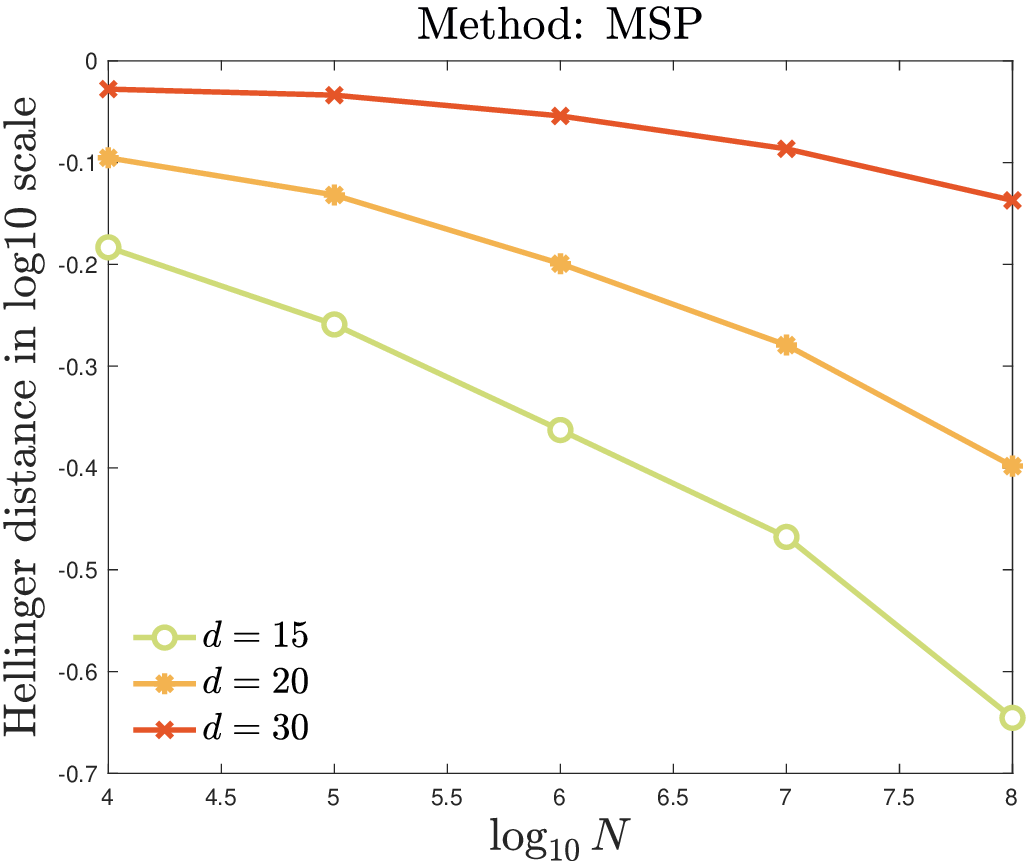}}}
\\
\centering
\subfigure[Partition level (left: DSP, middle: DSP-mix, right: MSP). ]{
{\includegraphics[width=0.32\textwidth,height=0.24\textwidth]{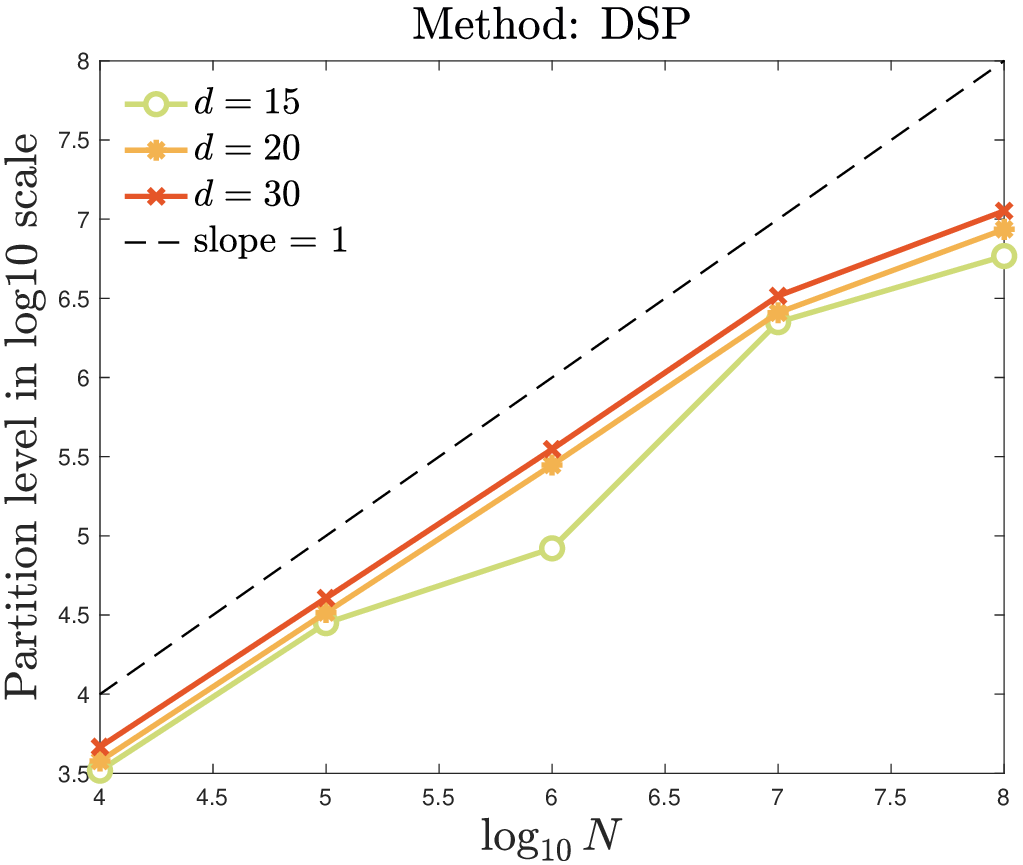}}
{\includegraphics[width=0.32\textwidth,height=0.24\textwidth]{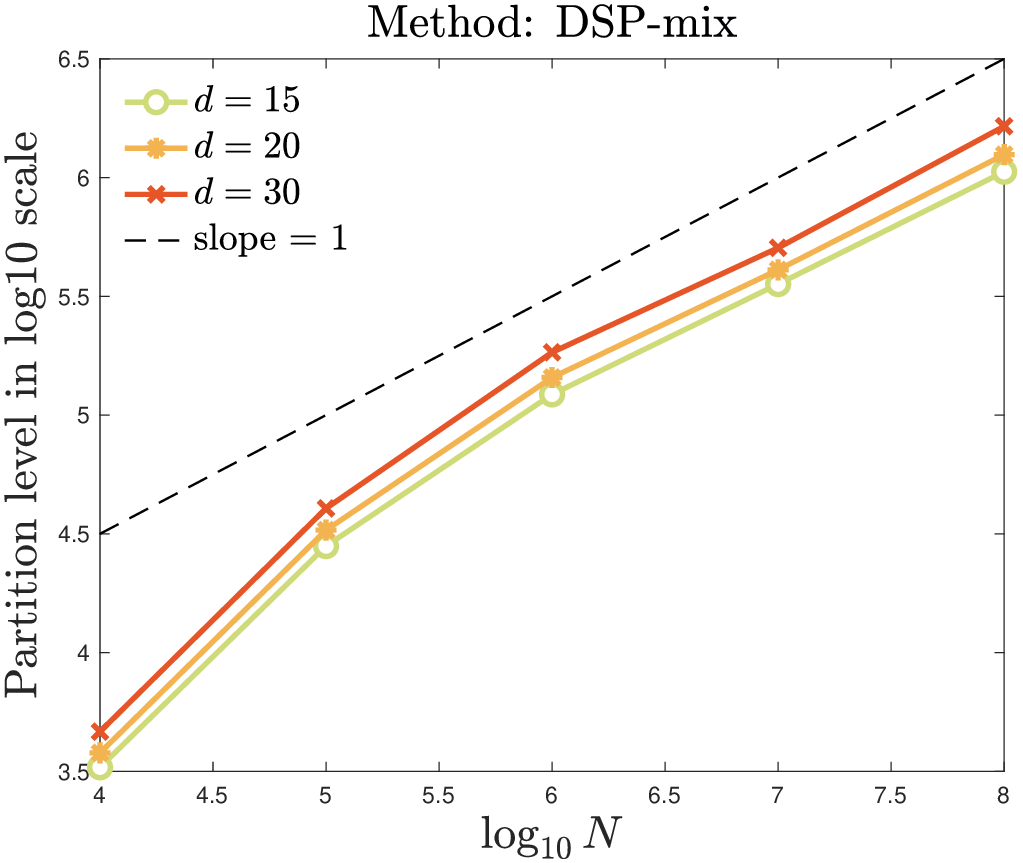}}
{\includegraphics[width=0.32\textwidth,height=0.24\textwidth]{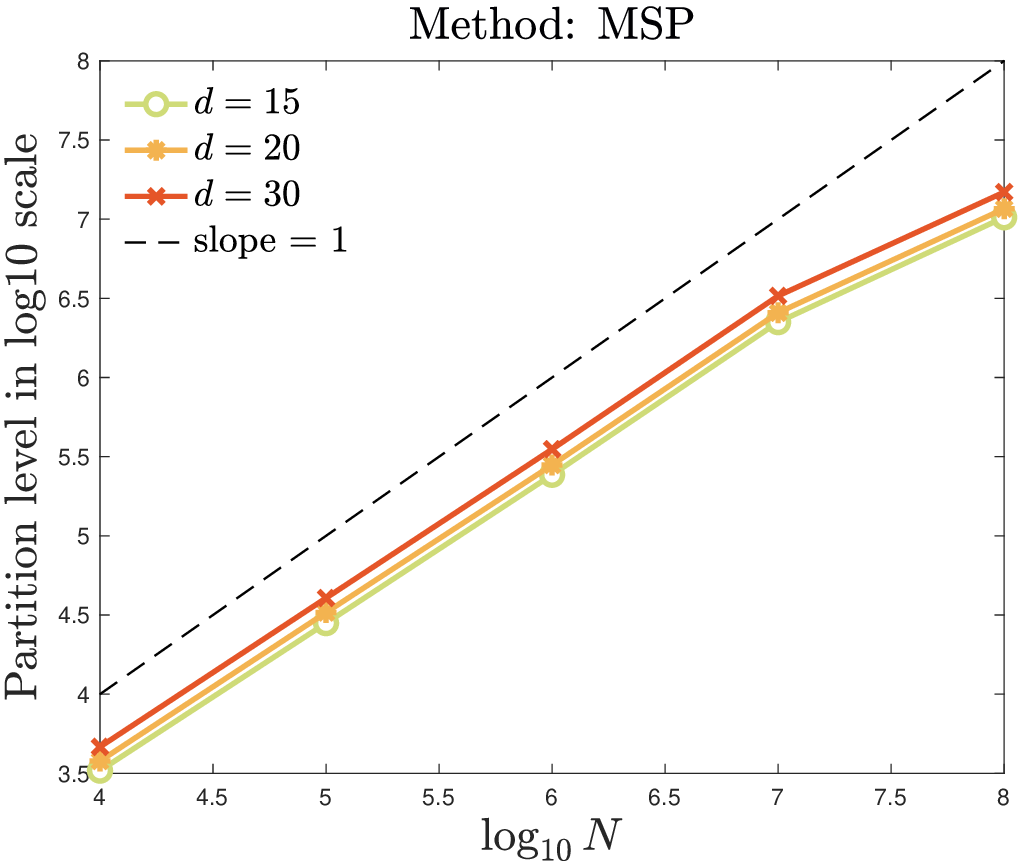}}}
\caption{\small $d$-D Beta mixtures: Convergence of DSP, DSP-mix and MSP and the partition level $L$ with respect to  $N$ for $d=15, 20, 30$. \label{beta_high_dimensional}}
\end{figure}

\begin{figure}[H]
\centering
\subfigure[Partition level ($N = 10^4$) (left: DSP, middle: DSP-mix, right: MSP). ]{
{\includegraphics[width=0.32\textwidth,height=0.24\textwidth]{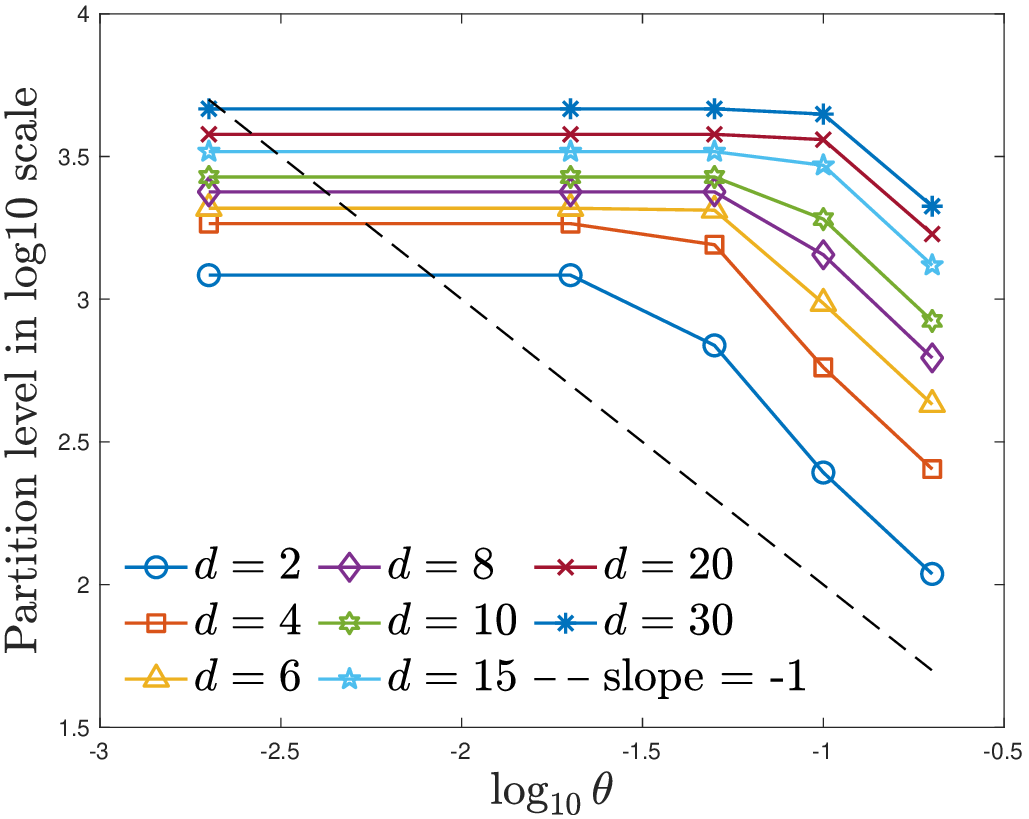}}
{\includegraphics[width=0.32\textwidth,height=0.24\textwidth]{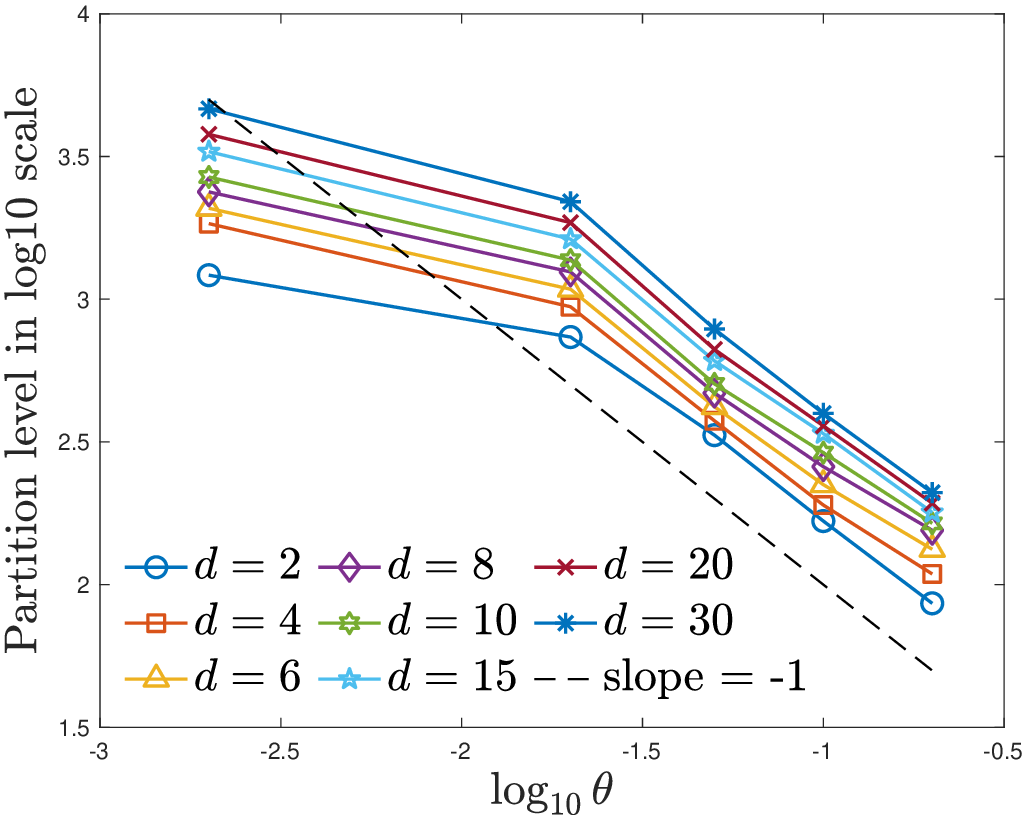}}
{\includegraphics[width=0.32\textwidth,height=0.24\textwidth]{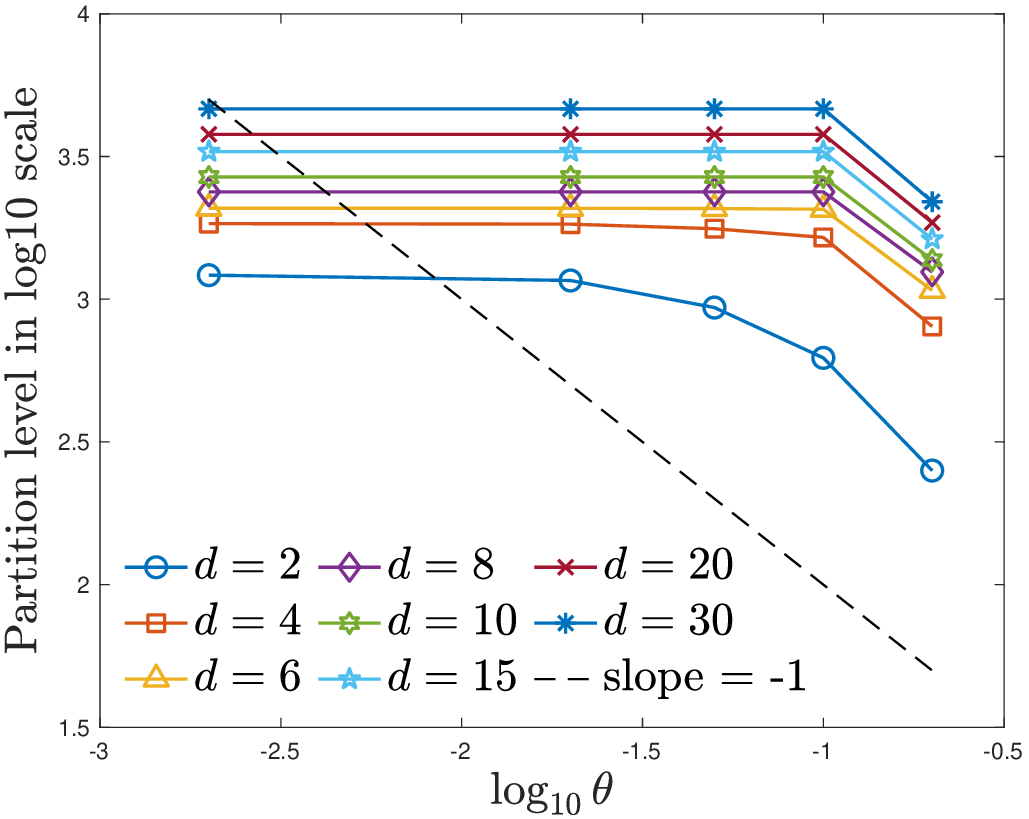}}}
\\
\centering
\subfigure[Partition level ($N = 10^5$) (left: DSP, middle: DSP-mix, right: MSP). ]{
{\includegraphics[width=0.32\textwidth,height=0.24\textwidth]{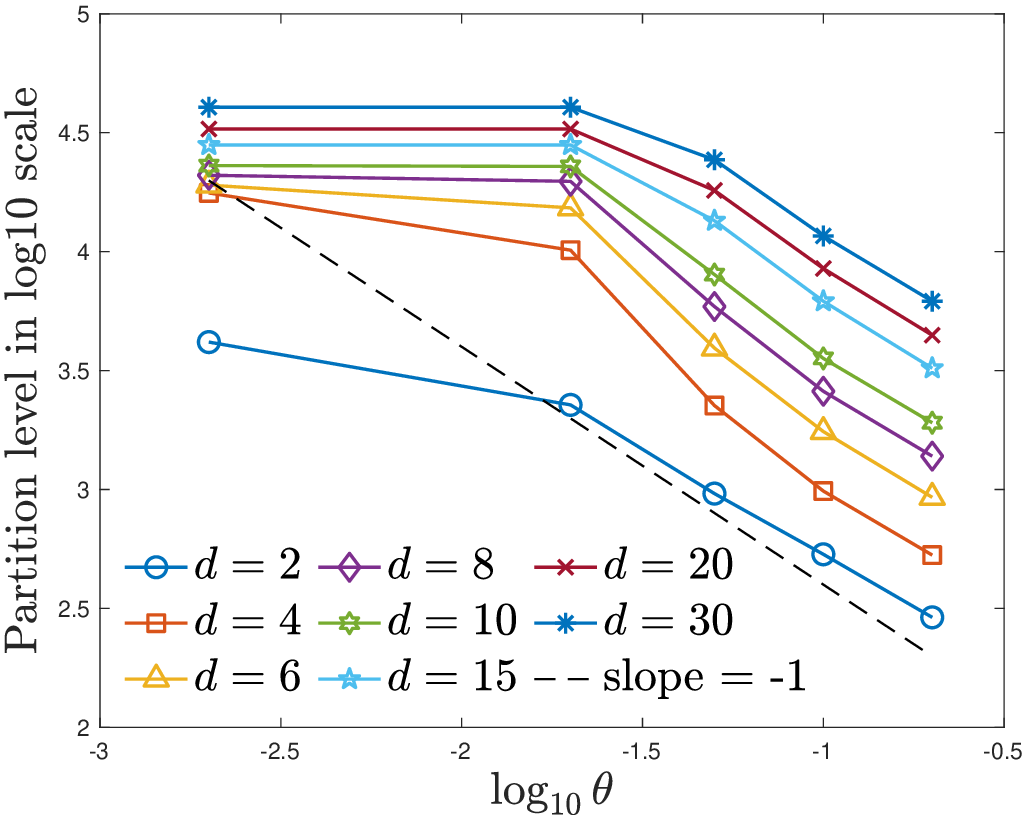}}
{\includegraphics[width=0.32\textwidth,height=0.24\textwidth]{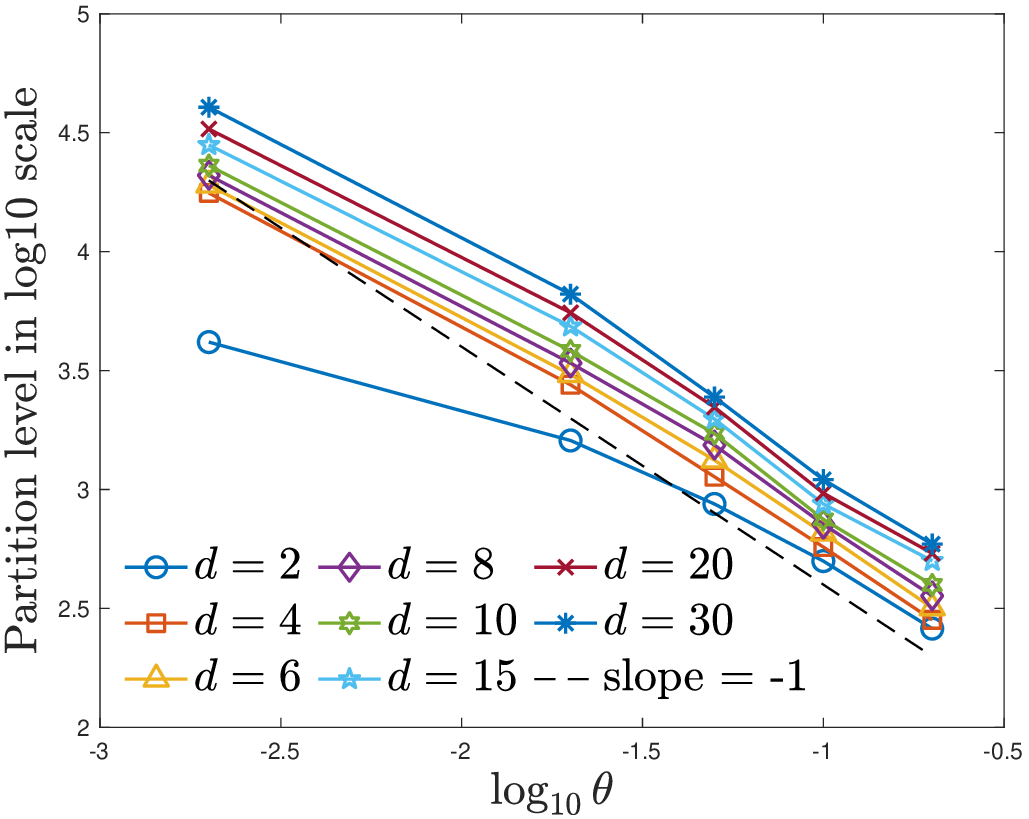}}
{\includegraphics[width=0.32\textwidth,height=0.24\textwidth]{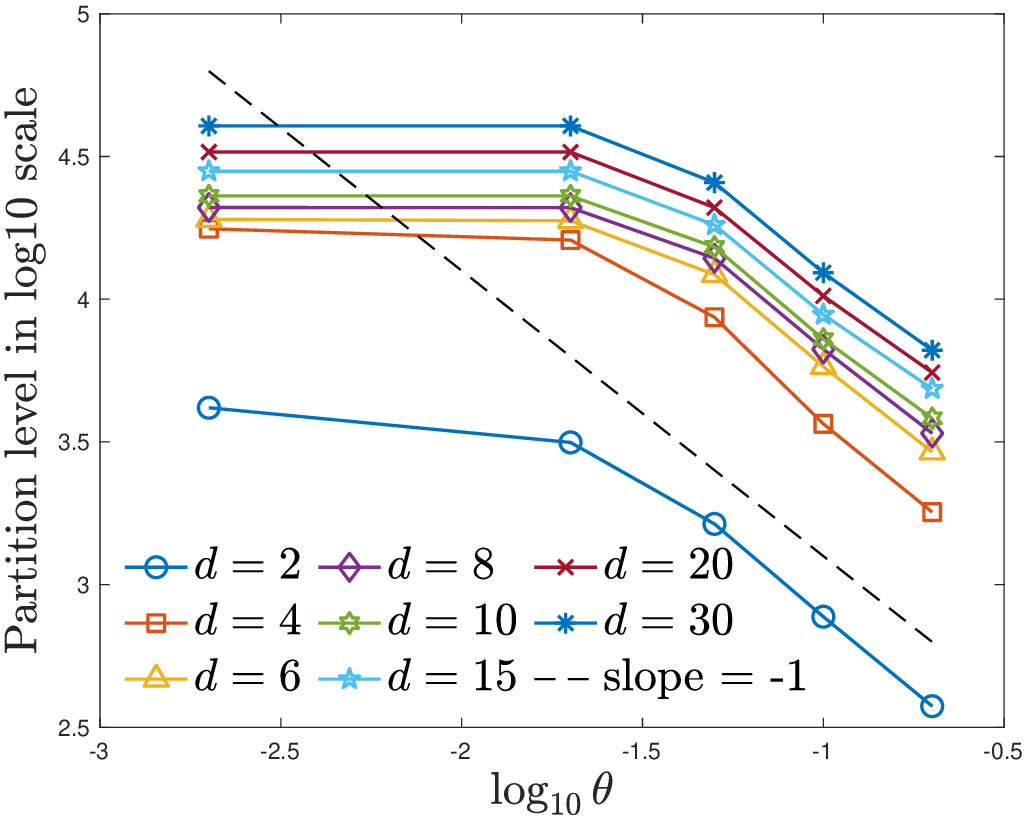}}}
\\
\centering
\subfigure[Partition level ($N = 10^6$) (left: DSP, middle: DSP-mix, right: MSP). ]{
{\includegraphics[width=0.32\textwidth,height=0.24\textwidth]{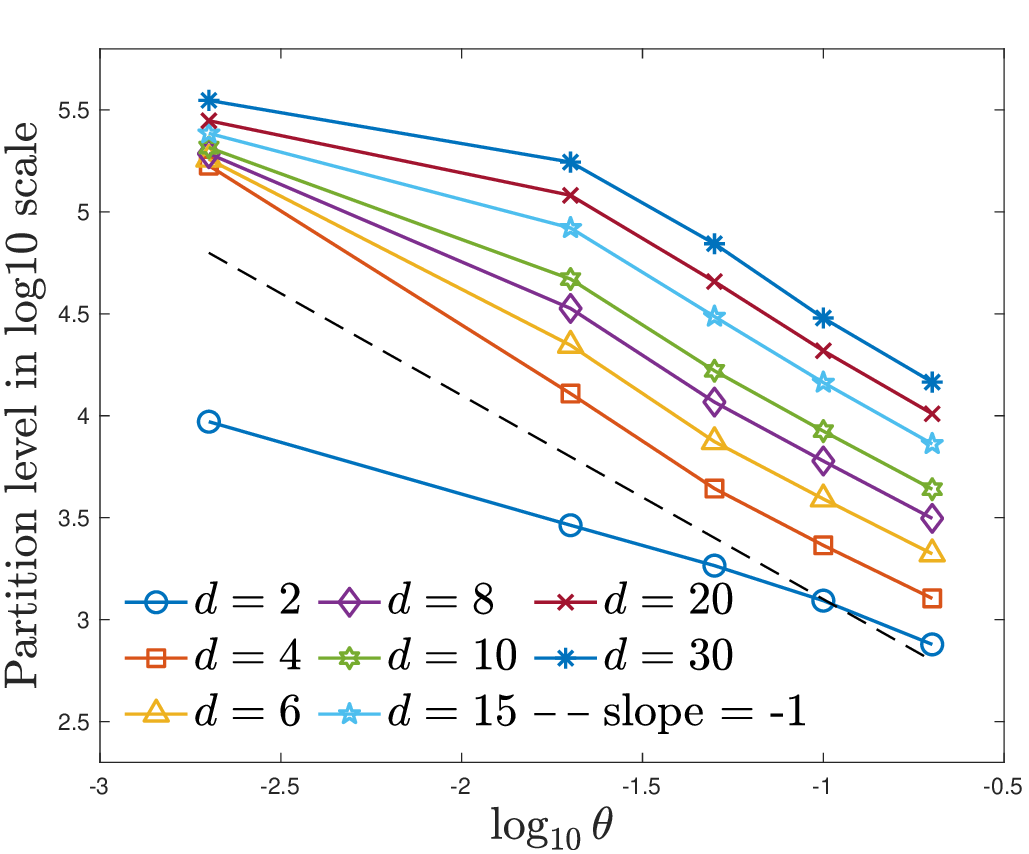}}
{\includegraphics[width=0.32\textwidth,height=0.24\textwidth]{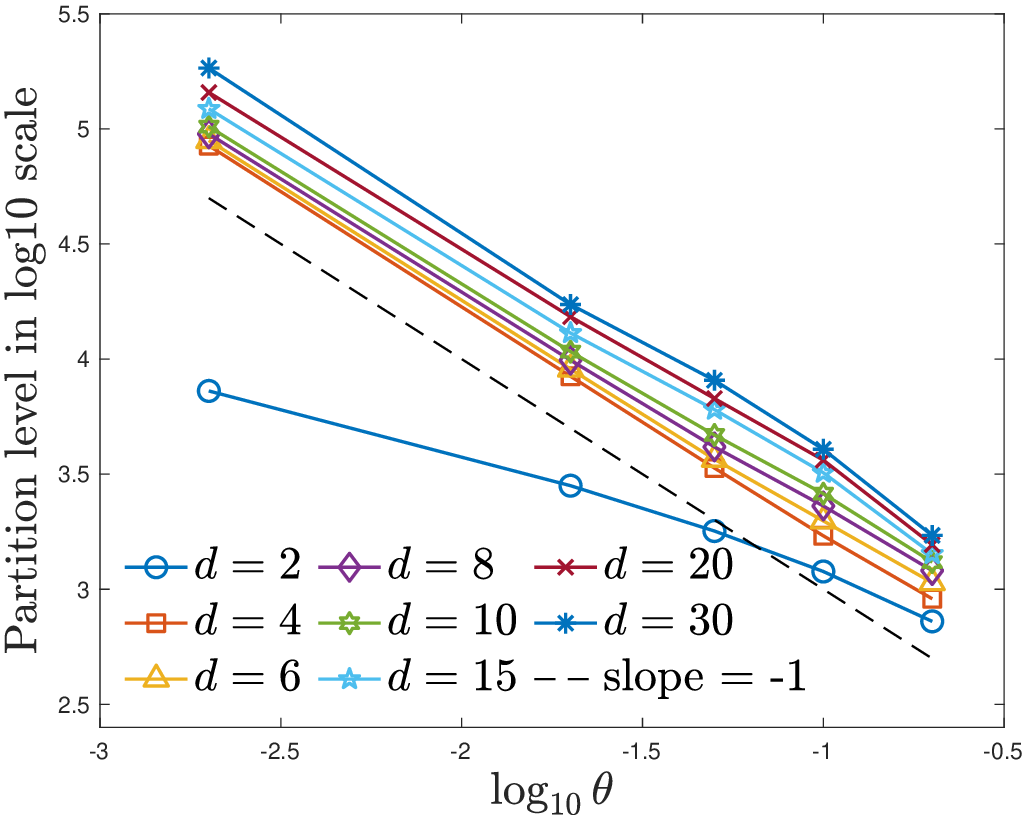}}
{\includegraphics[width=0.32\textwidth,height=0.24\textwidth]{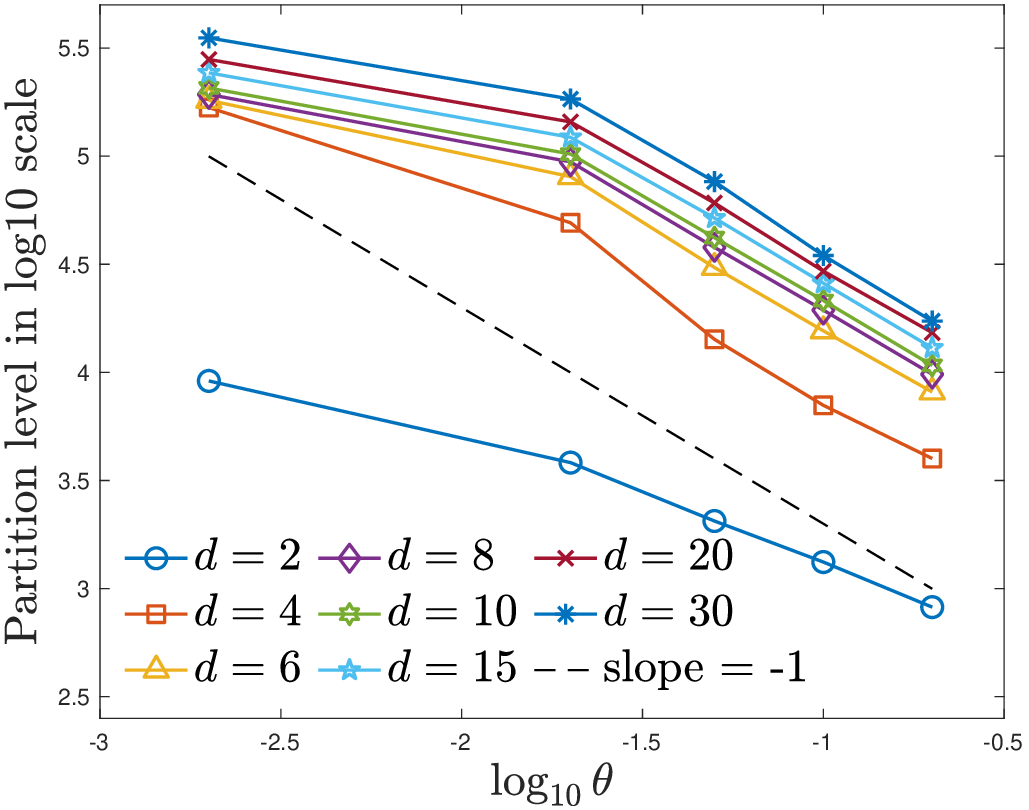}}}
\\
\centering
\subfigure[Partition level ($N = 10^7$) (left: DSP, middle: DSP-mix, right: MSP). ]{
{\includegraphics[width=0.32\textwidth,height=0.24\textwidth]{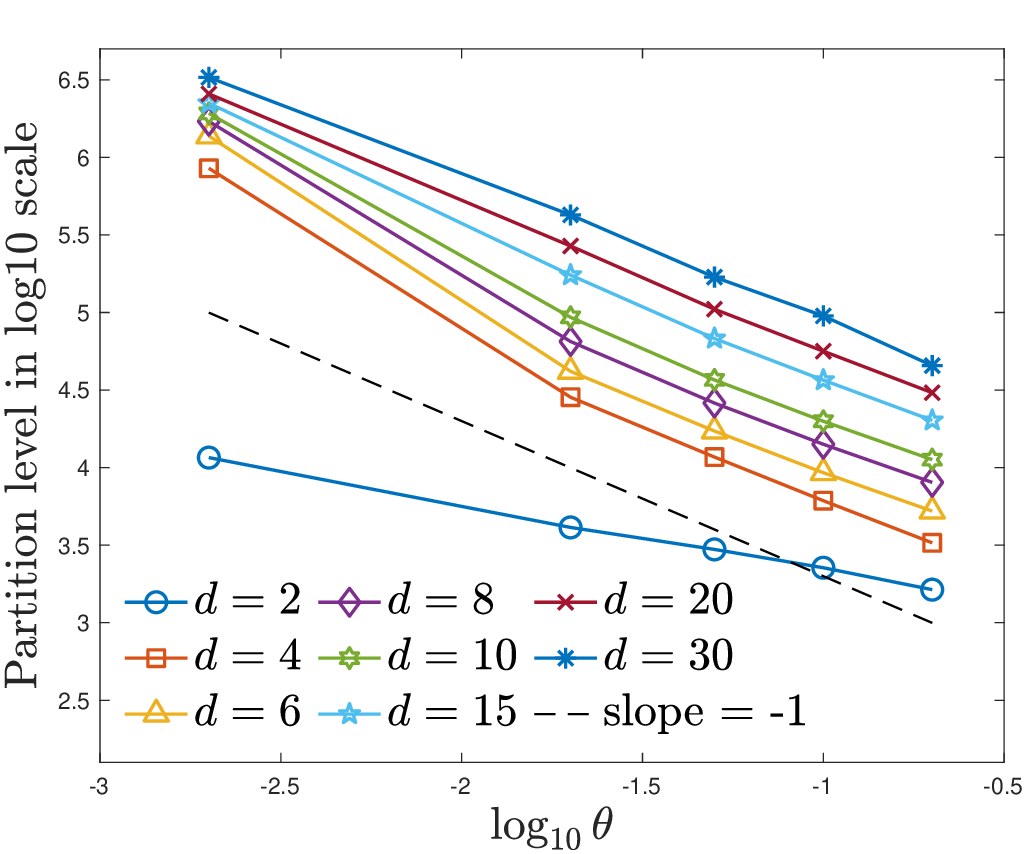}}
{\includegraphics[width=0.32\textwidth,height=0.24\textwidth]{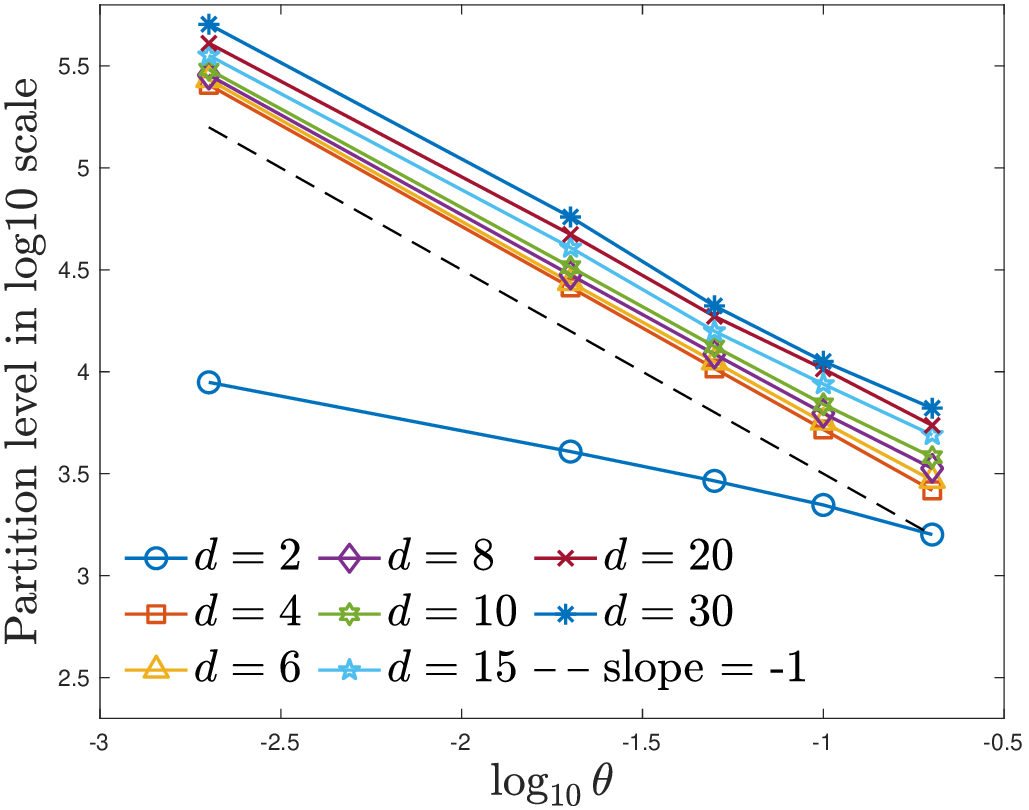}}
{\includegraphics[width=0.32\textwidth,height=0.24\textwidth]{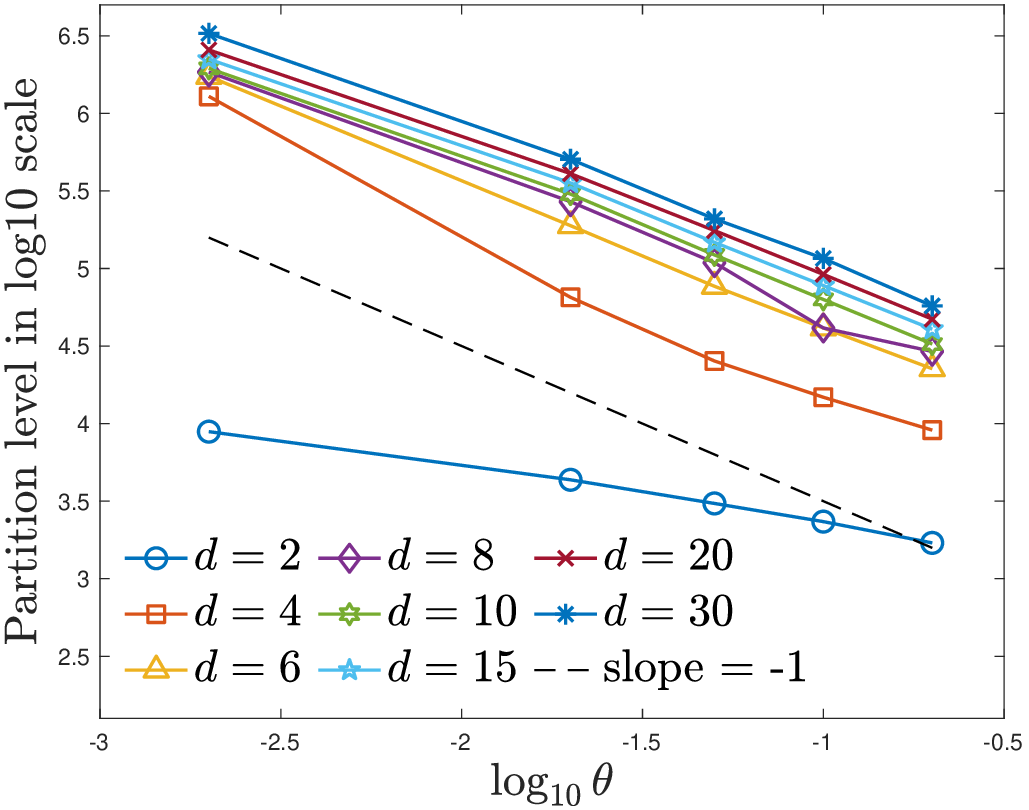}}}
\caption{\small $d$-D Beta mixtures: The partition level $L$ under different $\theta$, $N$ and $d$. It is seen that $L \sim \theta^{-1}$ when $N$ is sufficiently large.} \label{beta_partition}
\end{figure}

For high-dimensional tests with $d = 15, 20, 30$,  Figure \ref{beta_high_dimensional} presents the convergence of error metrics with respect to sample size $N = 10^4, 10^5, 10^6, 10^7, 10^8$, with the raw data collected in Tables~\ref{beta_N1e7} and \ref{beta_high}. Figure \ref{beta_partition} plots the partition level under different $d$, $N$, with $\theta$ fixed to be $0.002$ with the raw data collected in Tables \ref{beta_K_time_N1e7} and \ref{beta_K_time_high}. From the above results, we make several observations.
\begin{enumerate}

\item[(1)] All three methods are very robust as the standard derivations are much smaller than the mean errors.

\item[(2)] The accuracy of all methods can be improved by increasing sample size $N$, which validates the theoretical error bound.

\item[(3)] The accuracy of tree-based density estimation diminishes as dimension $d$ increases. This is because high-dimensional probability density tends to be concentrated in a very localized region (the typical set).

\item[(4)] The performance of DSP and MSP is comparable, while MSP is significantly faster. DSP-mix works well for $d \le 6$, but becomes less accurate as dimension $d$ increases. The reason is that the mixture discrepancy is a relaxation of the star discrepancy, so that the threshold $D^{\textup{mix}} \le \frac{\theta \sqrt{N}}{n}$ is more readily to attain. As a result, the partition level decreases, which is confirmed in Figure \ref{beta_partition}.

\item[(5)] The partition level $L$ increases as a smaller $\theta$ is chosen. When $N$ is large, it is seen that $L \sim \theta^{-1}$. When $\theta$ is too small, the partition ceases to be split further.  

\item[(6)] When $L$ is too large, the overpartitioning of space dramatically increases the computational cost but hampers the efficiency, due to the overfitting problem.

\item[(7)] The overfitting problem can be alleviated by increasing sample size $N$. 

\end{enumerate}

\begin{table}[H]
	\centering
	\caption{$d$-D Beta mixtures: The KL divergence and Hellinger distance  under different $d$ and $\theta$. The sample size is fixed to be $N = 10^7$.}
	\setlength{\tabcolsep}{4pt}
	\begin{tabular}{c|c|cc|cc|cc|cc|cc}
		\toprule
		& $\theta$	& \multicolumn{2}{c}{$0.002$} & \multicolumn{2}{c}{$0.02$} & \multicolumn{2}{c}{$0.05$} &  \multicolumn{2}{c}{$0.1$} &  \multicolumn{2}{c}{$0.2$}\\ 
		\midrule
		 $d$	&	Method	& KL & $\hat{H}^2$ &  KL & $\hat{H}^2$  & KL & $\hat{H}^2$  & KL & $\hat{H}^2$  & KL & $\hat{H}^2$  \\
		 \midrule
		 \multirow{4}{*}{2}	&	DSP 		&	-0.0007 	&	0.0011 	&	0.0013 	&	0.0008 	&	0.0019 	&	0.0008 	&	0.0028 	&	0.0010 	&	0.0042 	&	0.0013 	\\
		 				&	DSP-mix	&	0.0001 	&	0.0010 	&	0.0017 	&	0.0009 	&	0.0029 	&	0.0011 	&	0.0047 	&	0.0015 	&	0.0087 	&	0.0026 	\\
						&	MSP		&	0.0001 	&	0.0010 	&	0.0011 	&	0.0008 	&	0.0016 	&	0.0008 	&	0.0023 	&	0.0009 	&	0.0033 	&	0.0011 	\\	
						&	(maxSD)	&	0.0002 	&	0.0001 	&	0.0001 	&	0.0000 	&	0.0001 	&	0.0001 	&	0.0002 	&	0.0001 	&	0.0002 	&	0.0001 	\\
		 \midrule
		 \multirow{4}{*}{4}	&	DSP 		&	-0.2145 	&	0.0542 	&	0.0133 	&	0.0075 	&	0.0320 	&	0.0105 	&	0.0485 	&	0.0140 	&	0.0695 	&	0.0192 	\\
		 				&	DSP-mix	&	-0.0546 	&	0.0168 	&	0.0246 	&	0.0101 	&	0.0501 	&	0.0151 	&	0.0802 	&	0.0220 	&	0.1201 	&	0.0319 	\\
						&	MSP		&	-0.3167 	&	0.0784 	&	-0.0043 	&	0.0076 	&	0.0155 	&	0.0078 	&	0.0274 	&	0.0096 	&	0.0404 	&	0.0123 	\\
						&	(maxSD)	&	0.0012 	&	0.0004 	&	0.0002 	&	0.0001 	&	0.0006 	&	0.0002 	&	0.0008 	&	0.0003 	&	0.0012 	&	0.0005 	\\
		 \midrule
		 \multirow{4}{*}{6}	&	DSP 		&	-0.3601 	&	0.0976 	&	0.0914 	&	0.0314 	&	0.1490 	&	0.0424 	&	0.1978 	&	0.0535 	&	0.2531 	&	0.0669 	\\
		 				&	DSP-mix	&	-0.0181 	&	0.0328 	&	0.1506 	&	0.0441 	&	0.2303 	&	0.0614 	&	0.3240 	&	0.0822 	&	0.4245 	&	0.1053 	\\
						&	MSP		&	-0.4512 	&	0.1166 	&	0.0074 	&	0.0291 	&	0.0710 	&	0.0306 	&	0.1116 	&	0.0362 	&	0.1556 	&	0.0445 	\\	
						&	(maxSD)	&	0.0003 	&	0.0005 	&	0.0010 	&	0.0003 	&	0.0015 	&	0.0004 	&	0.0028 	&	0.0009 	&	0.0052 	&	0.0013 	\\
		 \midrule
		 \multirow{4}{*}{8}	&	DSP 		&	-0.4151 	&	0.1391 	&	0.2459 	&	0.0756 	&	0.3636 	&	0.0972 	&	0.4529 	&	0.1171 	&	0.5552 	&	0.1406 	\\
		 				&	DSP-mix	&	0.0943 	&	0.0677 	&	0.3946 	&	0.1043 	&	0.5627 	&	0.1387 	&	0.7416 	&	0.1738 	&	0.8929 	&	0.2052 	\\
						&	MSP		&	-0.4521 	&	0.1458 	&	0.0996 	&	0.0670 	&	0.2242 	&	0.0752 	&	0.1116 	&	0.0362 	&	0.3953 	&	0.1043 	\\
						&	(maxSD)	&	0.0008 	&	0.0006 	&	0.0029 	&	0.0008 	&	0.0058 	&	0.0022 	&	0.0130 	&	0.0029 	&	0.0126 	&	0.0034 	\\
		 \midrule
		 \multirow{4}{*}{10}	&	DSP 		&	-0.3763 	&	0.1862 	&	0.4781 	&	0.1359 	&	0.6679 	&	0.1687 	&	0.8025 	&	0.1957 	&	0.9590 	&	0.2284 	\\
		 				&	DSP-mix	&	0.2906 	&	0.1210 	&	0.7491 	&	0.1838 	&	1.0370 	&	0.2353 	&	1.2924 	&	0.2779 	&	1.4916 	&	0.3158 	\\
						&	MSP		&	-0.3842 	&	0.1875 	&	0.2912 	&	0.1211 	&	0.4852 	&	0.1386 	&	0.6142 	&	0.1582 	&	0.7492 	&	0.1838 	\\
						&	(maxSD)	&	0.0015 	&	0.0010 	&	0.0048 	&	0.0013 	&	0.0126 	&	0.0027 	&	0.0175 	&	0.0027 	&	0.0165 	&	0.0028 	\\
		 \midrule
		 \multirow{4}{*}{15}	&	DSP 		&	0.1017 	&	0.3404 	&	1.4157 	&	0.3311 	&	1.8129 	&	0.3810 	&	2.0849 	&	0.4222 	&	2.4313 	&	0.4713 	\\
		 				&	DSP-mix	&	1.1749 	&	0.3092 	&	2.0996 	&	0.4180 	&	2.7920 	&	0.4909 	&	3.1733 	&	0.5313 	&	3.5141 	&	0.5784 	\\
						&	MSP		&	0.1017 	&	0.3406 	&	1.1749 	&	0.3097 	&	1.5681 	&	0.3443 	&	1.8150 	&	0.3765 	&	2.0996 	&	0.4179 	\\	
						&	(maxSD)	&	0.0099 	&	0.0026 	&	0.0220 	&	0.0061 	&	0.0326 	&	0.0079 	&	0.0286 	&	0.0090 	&	0.0345 	&	0.0092 	\\
		 \midrule
		 \multirow{4}{*}{20}	&	DSP 		&	1.0627 	&	0.5251 	&	2.8794 	&	0.5400 	&	3.4996 	&	0.5934 	&	3.9360 	&	0.6369 	&	4.5112 	&	0.6807 	\\
		 				&	DSP-mix	&	2.6627 	&	0.5242 	&	4.1647 	&	0.6455 	&	5.2001 	&	0.6993 	&	5.7049 	&	0.7391 	&	6.2575 	&	0.7851 	\\
						&	MSP		&	1.0627 	&	0.5257 	&	2.6627 	&	0.5237 	&	3.2329 	&	0.5605 	&	3.6396 	&	0.6003 	&	4.1647 	&	0.6452 	\\
						&	(maxSD)	&	0.0195 	&	0.0027 	&	0.0213 	&	0.0032 	&	0.0205 	&	0.0026 	&	0.0269 	&	0.0033 	&	0.0295 	&	0.0032 	\\
		 \midrule
		 \multirow{4}{*}{30}	&	DSP 		&	4.2883 	&	0.8201 	&	7.2776 	&	0.8408 	&	8.2751 	&	0.8690 	&	8.9656 	&	0.8886 	&	10.1624 	&	0.9139 	\\
		 				&	DSP-mix	&	7.1310 	&	0.8337 	&	9.8885 	&	0.8990 	&	11.3890 	&	0.9258 	&	12.1404 	&	0.9391 	&	12.9147 	&	0.9523 	\\
						&	MSP		&	4.2883 	&	0.8195 	&	7.1310 	&	0.8342 	&	8.0927 	&	0.8590 	&	8.7329 	&	0.8761 	&	9.8885 	&	0.8989 	\\	
						&	(maxSD)	&	0.0382 	&	0.0027 	&	0.0289 	&	0.0017 	&	0.0417 	&	0.0017 	&	0.0296 	&	0.0018 	&	0.0252 	&	0.0012 	\\
		\bottomrule
	\end{tabular}
	\label{beta_N1e7}
\end{table}

\begin{table}[H]
	\centering
	\caption{$d$-D Beta mixtures: The partition level $L$ and computational time (in seconds) under different $d$ and $\theta$. The sample size is fixed to be $N  = 10^7$.}
	\setlength{\tabcolsep}{4pt}
	\begin{tabular}{c|c|cc|cc|cc|cc|cc}
		\toprule
		& $\theta$	& \multicolumn{2}{c}{$0.002$} & \multicolumn{2}{c}{$0.02$} & \multicolumn{2}{c}{$0.05$} &  \multicolumn{2}{c}{$0.1$} &  \multicolumn{2}{c}{$0.2$}\\ 
		\midrule
		 $d$	&	Method	& $L$ & time & $L$ & time  & $L$ & time  & $L$ & time  & $L$ & time \\
		 \midrule
		 \multirow{3}{*}{2}	&	DSP 		&	11601 	&	19.34 	&	4115 	&	20.62 	&	2969 	&	27.54 	&	2262 	&	39.84 	&	1634 	&	66.47 	\\
		 				&	DSP-mix	&	8874 	&	15.35 	&	4059 	&	15.23 	&	2915 	&	15.75 	&	2221 	&	19.80 	&	1590 	&	31.96 	\\
						&	MSP		&	8874 	&	14.08 	&	4342 	&	14.31 	&	3057 	&	15.00 	&	2334 	&	14.74 	&	1702 	&	14.61 	\\					
		 \midrule
		 \multirow{3}{*}{4}	&	DSP 		&	850435 	&	947.49 	&	28431 	&	291.64 	&	11696 	&	292.27 	&	6119 	&	306.78 	&	3275 	&	329.77 	\\
		 				&	DSP-mix	&	255326 	&	41.38 	&	25936 	&	43.71 	&	10381 	&	62.42 	&	5213 	&	96.96 	&	2619 	&	165.91 	\\
						&	MSP		&	1287088 	&	49.38 	&	65181 	&	28.12 	&	25332 	&	27.64 	&	14790 	&	26.30 	&	9079 	&	25.90 	\\
		 \midrule
		 \multirow{3}{*}{6}	&	DSP 		&	1373682 	&	1146.49 	&	41757 	&	430.30 	&	17211 	&	392.71 	&	9272 	&	393.10 	&	5246 	&	420.23 	\\
		 				&	DSP-mix	&	270966 	&	56.15 	&	27402 	&	62.56 	&	11188 	&	89.92 	&	5656 	&	140.38 	&	2935 	&	239.13 	\\
						&	MSP		&	1731318 	&	71.70 	&	188819 	&	42.25 	&	76807 	&	42.46 	&	41359 	&	40.14 	&	22581 	&	38.46 	\\		
		 \midrule
		 \multirow{3}{*}{8}	&	DSP 		&	1707927 	&	970.22 	&	65024 	&	595.31 	&	26072 	&	514.57 	&	14138 	&	489.84 	&	8032 	&	519.53 	\\
		 				&	DSP-mix	&	285207 	&	68.54 	&	29847 	&	80.34 	&	12204 	&	117.71 	&	6260 	&	179.82 	&	3359 	&	303.59 	\\
						&	MSP		&	1845215 	&	91.44 	&	269574 	&	59.93 	&	108963 	&	55.74 	&	41359 	&	40.14 	&	29197 	&	50.70 	\\
		 \midrule
		 \multirow{3}{*}{10}	&	DSP 		&	1921213 	&	846.16 	&	93104 	&	751.74 	&	36728 	&	632.04 	&	19901 	&	585.76 	&	11259 	&	605.56 	\\
		 				&	DSP-mix	&	303408 	&	83.33 	&	32637 	&	96.91 	&	13314 	&	142.84 	&	6924 	&	216.97 	&	3811 	&	362.10 	\\
						&	MSP		&	1949089 	&	110.36 	&	301774 	&	72.19 	&	122332 	&	69.06 	&	63136 	&	65.12 	&	32591 	&	62.86 	\\
		 \midrule
		 \multirow{3}{*}{15}	&	DSP 		&	2235855 	&	698.09 	&	174225 	&	1090.08 	&	67961 	&	875.88 	&	36638 	&	809.21 	&	20170 	&	859.77 	\\
		 				&	DSP-mix	&	356114 	&	121.08 	&	40629 	&	140.78 	&	15801 	&	204.62 	&	8697 	&	300.24 	&	4900 	&	507.12 	\\
						&	MSP		&	2235891 	&	166.61 	&	356111 	&	108.13 	&	147379 	&	100.36 	&	77649 	&	94.37 	&	40629 	&	91.47 	\\
		 \midrule
		 \multirow{3}{*}{20}	&	DSP 		&	2565816 	&	723.45 	&	268238 	&	1436.26 	&	105010 	&	1174.55 	&	56290 	&	1083.46 	&	30392 	&	2503.68 	\\
		 				&	DSP-mix	&	408780 	&	160.17 	&	47167 	&	184.88 	&	18688 	&	260.86 	&	10325 	&	385.46 	&	5456 	&	652.46 	\\
						&	MSP		&	2565816 	&	226.91 	&	408780 	&	143.17 	&	175550 	&	134.04 	&	91643 	&	125.57 	&	47167 	&	121.13 	\\
		 \midrule
		 \multirow{3}{*}{30}	&	DSP 		&	3277027 	&	898.88 	&	425546 	&	2079.65 	&	169021 	&	1626.90 	&	95058 	&	1825.46 	&	45432 	&	2000.92 	\\
		 				&	DSP-mix	&	506018 	&	236.90 	&	57427 	&	269.54 	&	21036 	&	379.71 	&	11243 	&	548.21 	&	6627 	&	914.78 	\\
						&	MSP		&	3277027 	&	359.31 	&	506018 	&	218.18 	&	208819 	&	196.87 	&	116137 	&	184.91 	&	57427 	&	178.54 	\\																		
		\bottomrule
	\end{tabular}
	\label{beta_K_time_N1e7}
\end{table}

\begin{table}[H]
	\centering
	\caption{$d$-D Beta mixtures: The KL divergence and Hellinger distance  under different $d$ and $N$. The parameter $\theta$ is fixed to be $0.002$.}
	\setlength{\tabcolsep}{4pt}
	\begin{tabular}{c|c|cc|cc|cc|cc|cc}
		\toprule
		& $N$	& \multicolumn{2}{c}{$1\times10^4$} & \multicolumn{2}{c}{$1\times10^5$} & \multicolumn{2}{c}{$1\times10^6$} &  \multicolumn{2}{c}{$1\times10^7$} &  \multicolumn{2}{c}{$1\times10^8$}\\ 
		\midrule
		 $d$	&	Method	& KL & $\hat{H}^2$ &  KL & $\hat{H}^2$  & KL & $\hat{H}^2$  & KL & $\hat{H}^2$  & KL & $\hat{H}^2$  \\
		 \midrule
		 \multirow{3}{*}{15}	&	DSP 		&	2.0391 	&	0.6561 	&	1.2169 	&	0.5513 	&	1.4069 	&	0.4060 	&	0.1017 	&	0.3404 	&	0.4820 	&	0.2105 	\\
		 				&	DSP-mix	&	2.0391 	&	0.6554 	&	1.2169 	&	0.5516 	&	1.1811 	&	0.4085 	&	1.1749 	&	0.3092 	&	1.0136 	&	0.2461 	\\
						&	MSP		&	2.0391 	&	0.6558 	&	1.2169 	&	0.5508 	&	0.5660 	&	0.4338 	&	0.1017 	&	0.3406 	&	0.2728 	&	0.2262 	\\
		 \midrule
		 \multirow{3}{*}{20}	&	DSP 		&	4.0528 	&	0.8026 	&	2.9441 	&	0.7384 	&	1.9095 	&	0.6322 	&	1.0627 	&	0.5251 	&	1.2661 	&	0.3904 	\\
		 				&	DSP-mix	&	4.0528 	&	0.8027 	&	2.9441 	&	0.7382 	&	2.7194 	&	0.6178 	&	2.6627 	&	0.5242 	&	2.2509 	&	0.4465 	\\
						&	MSP		&	4.0528 	&	0.8030 	&	2.9441 	&	0.7382 	&	1.9095 	&	0.6320 	&	1.0627 	&	0.5257 	&	1.0996 	&	0.3998 	\\
		 \midrule
		 \multirow{3}{*}{30}	&	DSP 		&	9.0334 	&	0.9385 	&	7.6243 	&	0.9253 	&	5.8924 	&	0.8828 	&	4.2883 	&	0.8201 	&	3.6955 	&	0.7627 	\\
		 				&	DSP-mix	&	9.0334 	&	0.9377 	&	7.6243 	&	0.9254 	&	7.1100 	&	0.8798 	&	7.1310 	&	0.8337 	&	6.2093 	&	0.7810 	\\
						&	MSP		&	9.0334 	&	0.9380 	&	7.6243 	&	0.9253 	&	5.8924 	&	0.8828 	&	4.2883 	&	0.8195 	&	4.0776 	&	0.7293 	\\																		
		\bottomrule
	\end{tabular}
	\label{beta_high}
\end{table}

\begin{table}[H]
	\centering
	\caption{$d$-D Beta mixtures: The partition level $L$ and computational time (in seconds) under different $d$ and $N$. The parameter $\theta$ is fixed to be $0.002$.}
	\setlength{\tabcolsep}{4pt}
	\begin{tabular}{c|c|cc|cc|cc|cc|cc}
		\toprule
		& $N$	& \multicolumn{2}{c}{$1\times10^4$} & \multicolumn{2}{c}{$1\times10^5$} & \multicolumn{2}{c}{$1\times10^6$} &  \multicolumn{2}{c}{$1\times10^7$} &  \multicolumn{2}{c}{$1\times10^8$}\\ 
		\midrule
		 $d$	&	Method	& $L$ & time & $L$ & time  & $L$ & time  & $L$ & time  & $L$ & time \\
		 \midrule
		 \multirow{3}{*}{15}	&	DSP 		&	3289 	&	0.14 	&	28088 	&	1.37 	&	83373 	&	161.28 	&	2235855 	&	698.09 	&	5850719 	&	29904.56 	\\
		 				&	DSP-mix	&	3289 	&	0.15 	&	28088 	&	1.47 	&	122187 	&	13.37 	&	356114 	&	121.08 	&	1058493 	&	1335.90 	\\
						&	MSP		&	3289 	&	0.15 	&	28088 	&	1.35 	&	242887 	&	15.29 	&	2235891 	&	166.61 	&	10274742 	&	1434.56 	\\	
		 \midrule
		 \multirow{3}{*}{20}	&	DSP 		&	3782 	&	0.20 	&	32780 	&	1.95 	&	280051 	&	20.97 	&	2565816 	&	723.45 	&	8624426 	&	37177.42 	\\
		 				&	DSP-mix	&	3782 	&	0.21 	&	32780 	&	2.09 	&	143972 	&	18.12 	&	408780 	&	160.17 	&	1249503 	&	1720.63 	\\
						&	MSP		&	3782 	&	0.21 	&	32780 	&	1.98 	&	280051 	&	21.12 	&	2565816 	&	226.91 	&	11640399 	&	1854.46 	\\	
		 \midrule
		 \multirow{3}{*}{30}	&	DSP 		&	4646 	&	0.34 	&	40468 	&	3.27 	&	352382 	&	34.25 	&	3277027 	&	898.88 	&	11265426 	&	38853.64 	\\
		 				&	DSP-mix	&	4646 	&	0.35 	&	40468 	&	3.61 	&	183723 	&	28.50 	&	506018 	&	236.90 	&	1644069 	&	2407.87 	\\
						&	MSP		&	4646 	&	0.35 	&	40468 	&	3.44 	&	352382 	&	34.30 	&	3277027 	&	359.31 	&	14813816 	&	2909.42 	\\																		
		\bottomrule
	\end{tabular}
	\label{beta_K_time_high}
\end{table}

\subsection{$d$-D Gaussian mixtures}


Now consider the $d$-D Gaussian mixtures \cite{Li2016},
\begin{equation}
	\bx_i \sim\left(\sum_{i=1}^4 \vec{\alpha}_i \mathcal{N}\left(\vec{\mu}_i, \vec{\Sigma}_i\right)\right),\,i = 1,\dots,N, 
\end{equation}
where $\vec{\alpha} = (0.4,0.3,0.2,0.1)$, $\vec{\mu}_1 = 0.3 \mone, \vec{\mu}_2 = 0.4 \mone, \vec{\mu}_3 = 0.5\mone, \vec{\mu}_4 = 0.6\mone$, $\vec{\Sigma}_1 = 0.01 \vec{I}, \vec{\Sigma}_2 = 0.02\vec{I}, \vec{\Sigma}_3 = 0.01 \vec{I}, \vec{\Sigma}_4 = 0.02 \vec{I}$. Here $\mone$ is a $1\times d$ all-one vector, and $\vec{I}$ is a $d\times d$ identity matrix.  

\begin{figure}[!h]
\centering
\subfigure[KL divergence ($d=2$) (left: DSP, middle: DSP-mix, right: MSP). ]{
{\includegraphics[width=0.32\textwidth,height=0.24\textwidth]{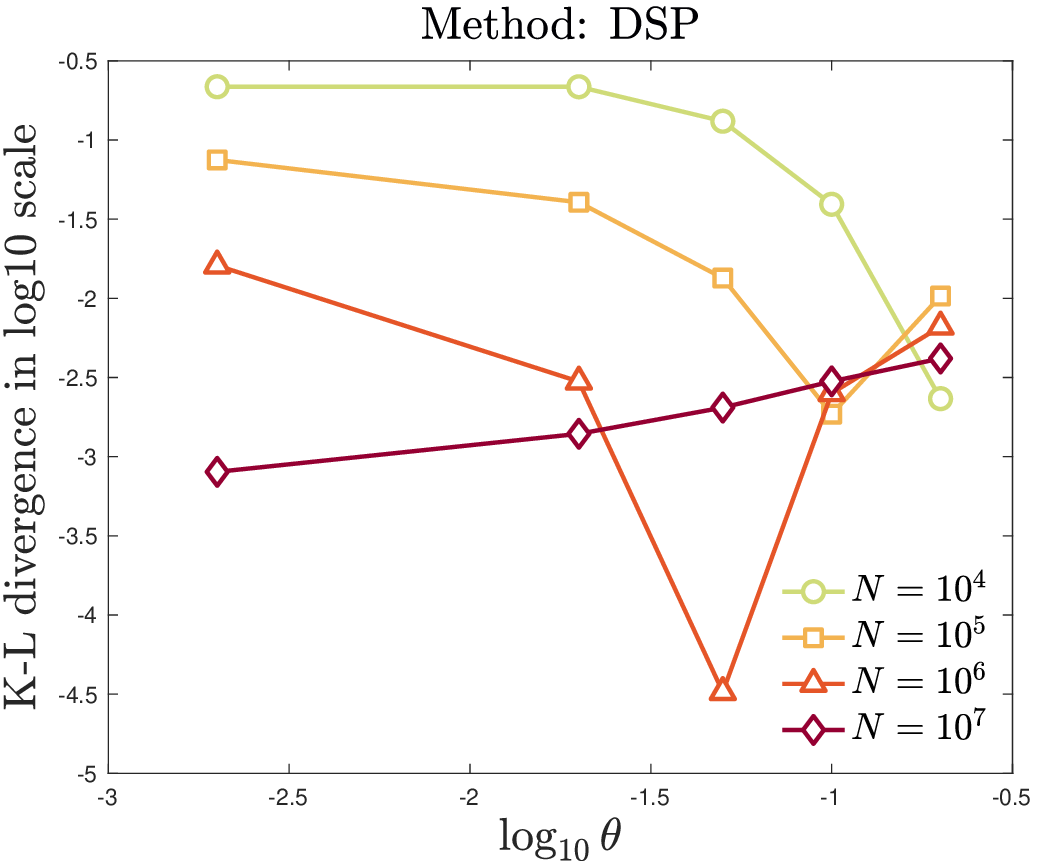}}
{\includegraphics[width=0.32\textwidth,height=0.24\textwidth]{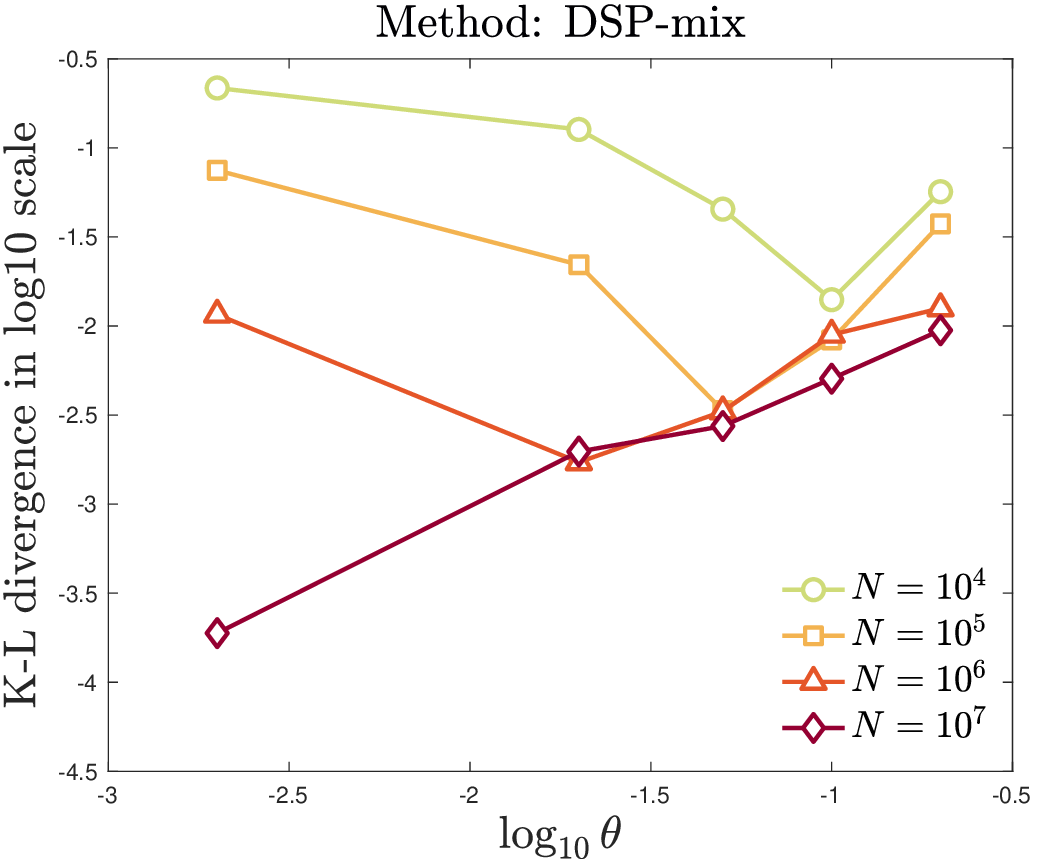}}
{\includegraphics[width=0.32\textwidth,height=0.24\textwidth]{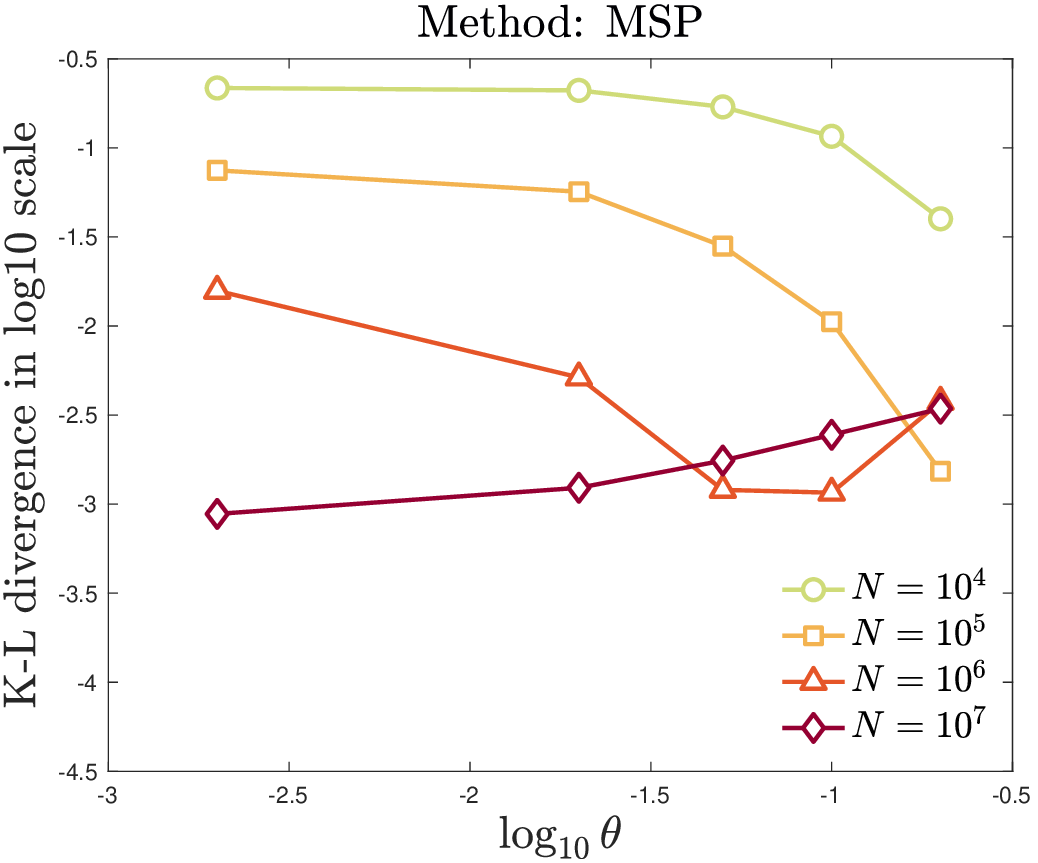}}}
\\
\centering
\subfigure[Hellinger distance ($d=2$) (left: DSP, middle: DSP-mix, right: MSP). ]{
{\includegraphics[width=0.32\textwidth,height=0.24\textwidth]{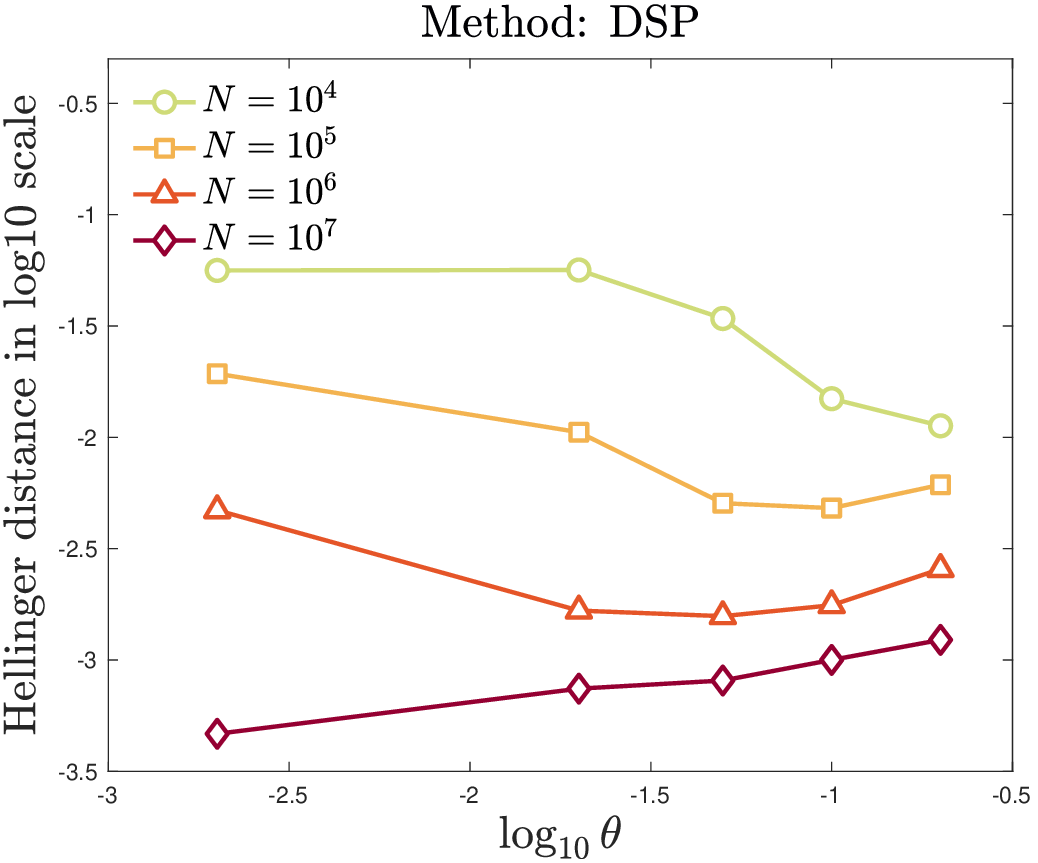}}
{\includegraphics[width=0.32\textwidth,height=0.24\textwidth]{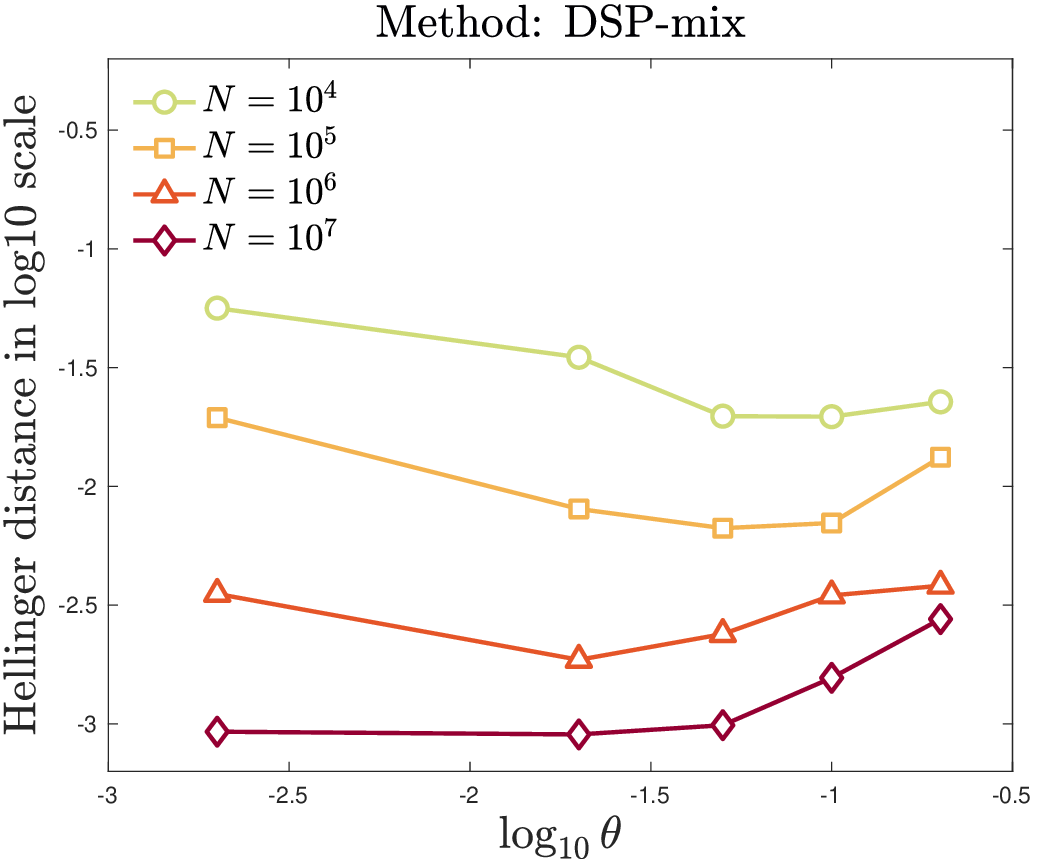}}
{\includegraphics[width=0.32\textwidth,height=0.24\textwidth]{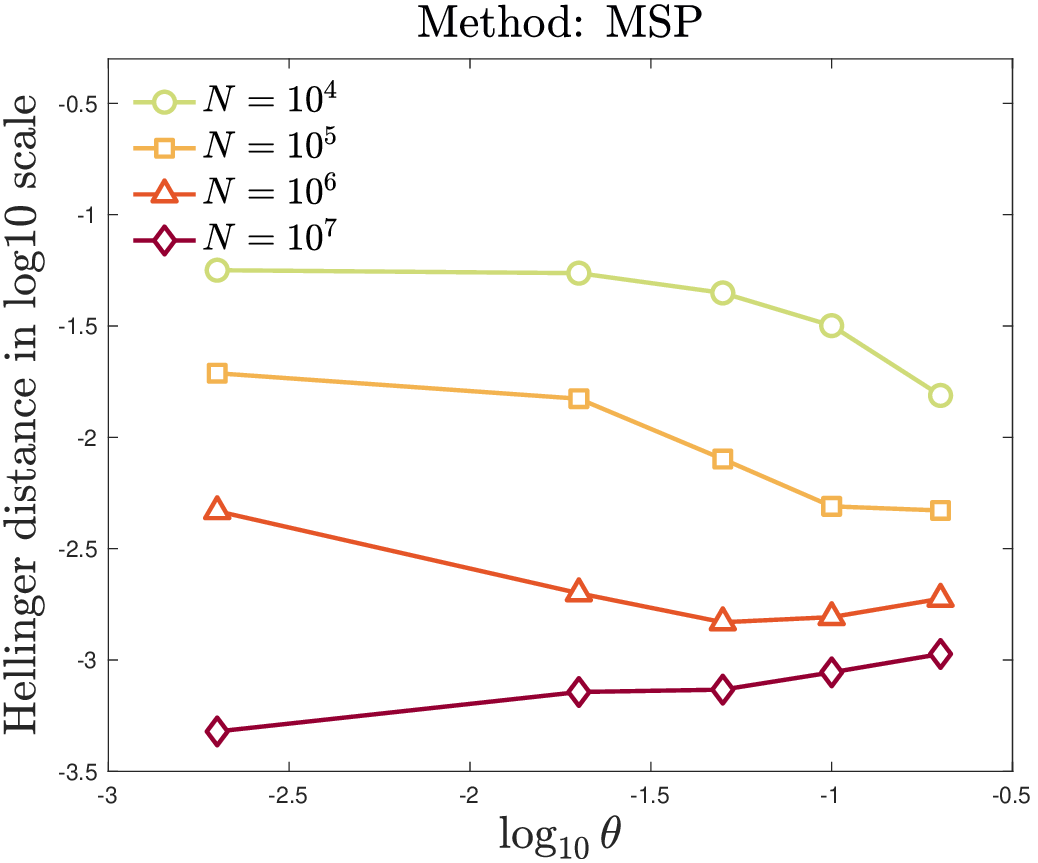}}}
\caption{\small $2$-D Gaussian mixtures: The KL divergence and Hellinger distance under different $N$ and $\theta$.}
\label{gauss_low}
\end{figure}

\begin{figure}[!h]
\centering
\subfigure[True density and samples.]{
\includegraphics[width=0.4\textwidth,height=0.3\textwidth]{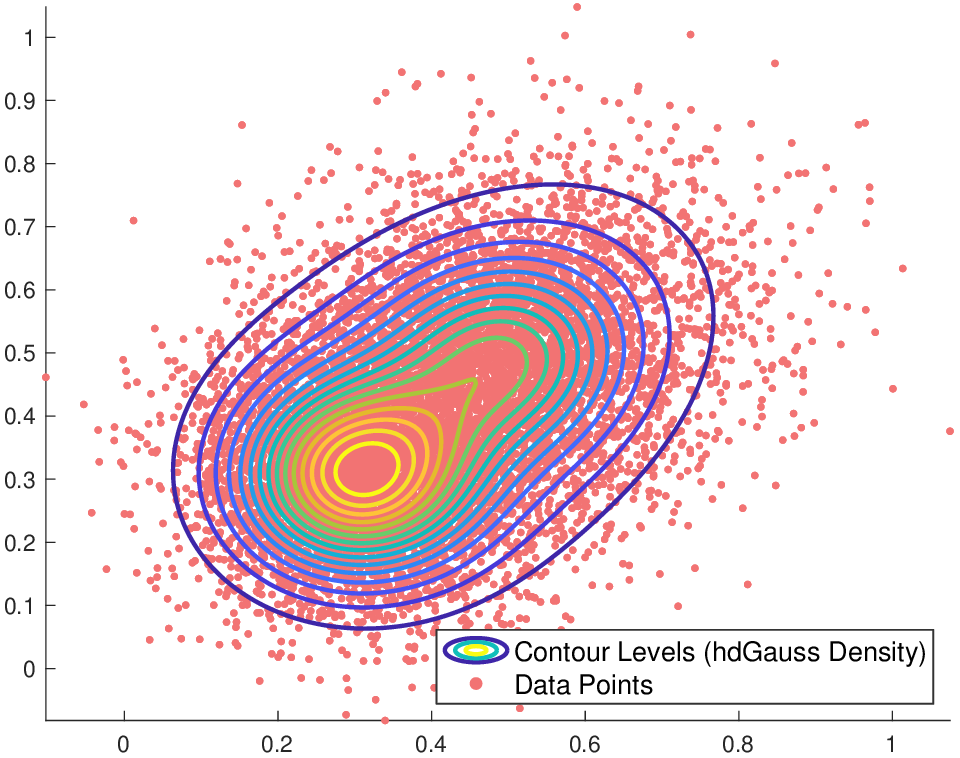}}
\\
\centering
\subfigure[Adaptive partition (left: DSP, middle: DSP-mix, right: MSP).]{
\includegraphics[width=0.32\textwidth,height=0.24\textwidth]{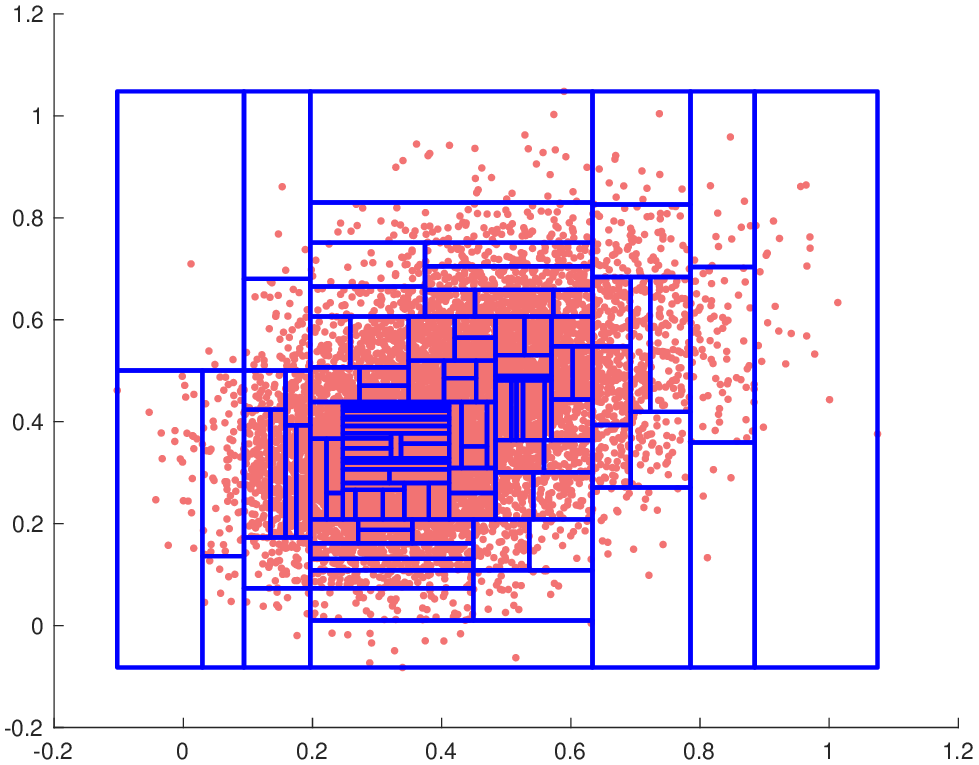}
\includegraphics[width=0.32\textwidth,height=0.24\textwidth]{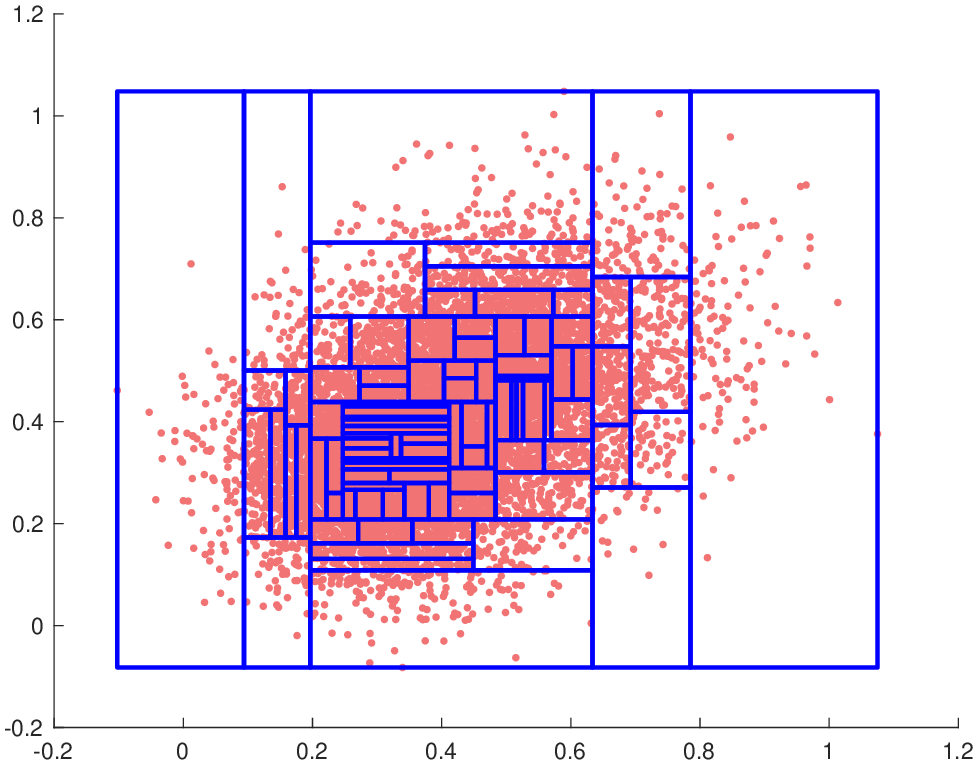}
\includegraphics[width=0.32\textwidth,height=0.24\textwidth]{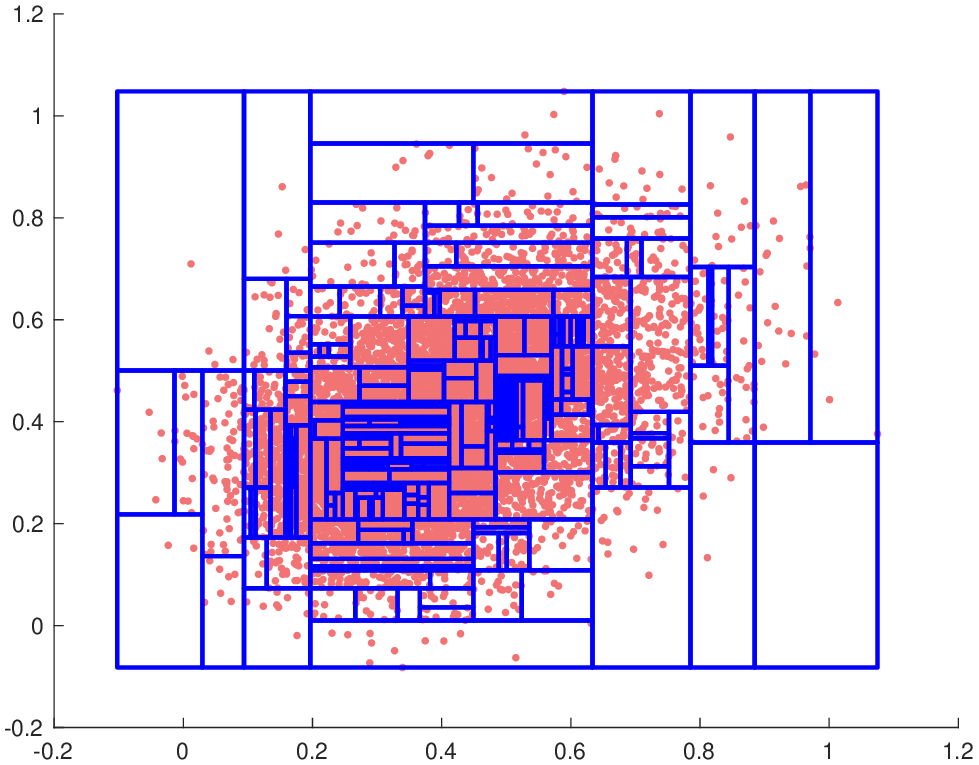}}
\\
\centering
\subfigure[Density estimation (left: DSP, middle: DSP-mix, right: MSP).]{
\includegraphics[width=0.32\textwidth,height=0.24\textwidth]{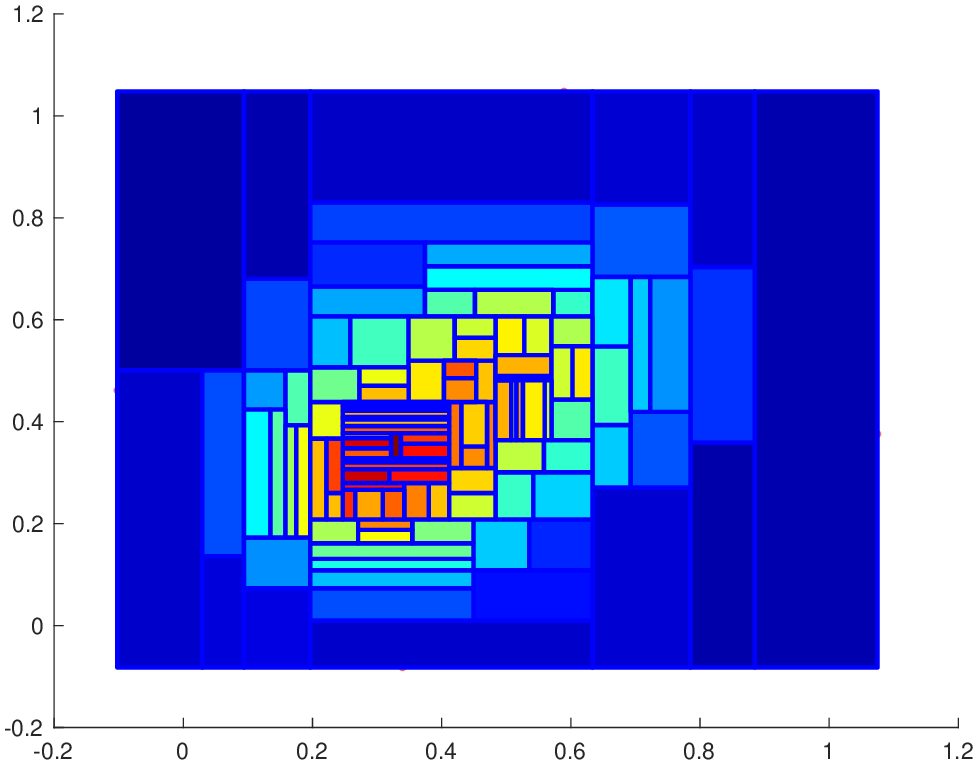}
\includegraphics[width=0.32\textwidth,height=0.24\textwidth]{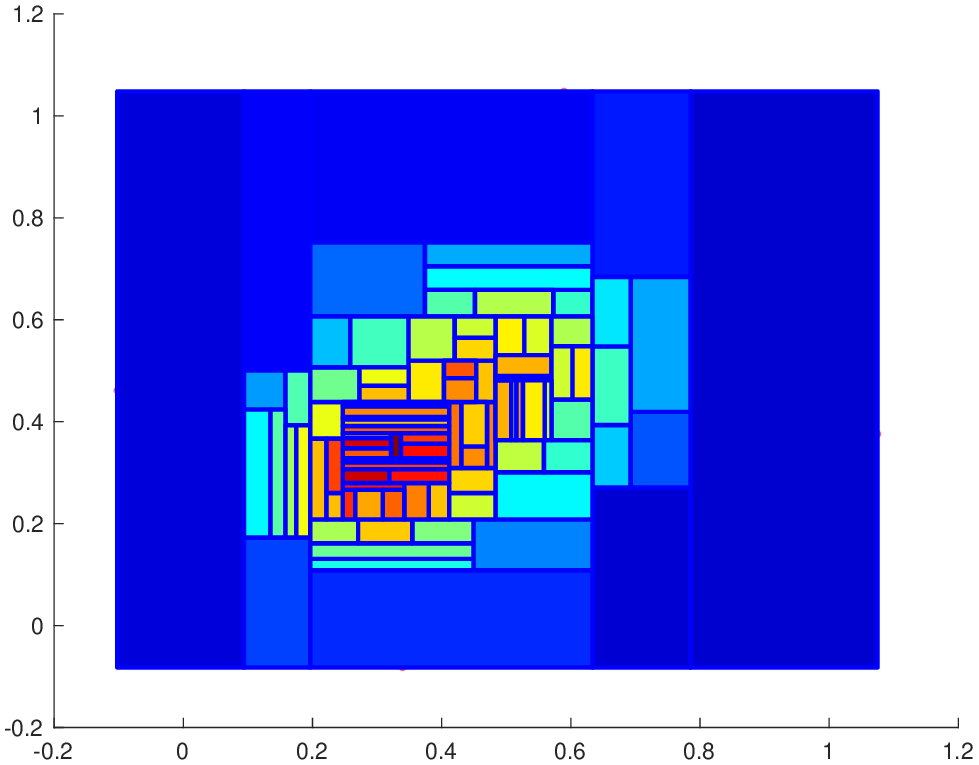}
\includegraphics[width=0.32\textwidth,height=0.24\textwidth]{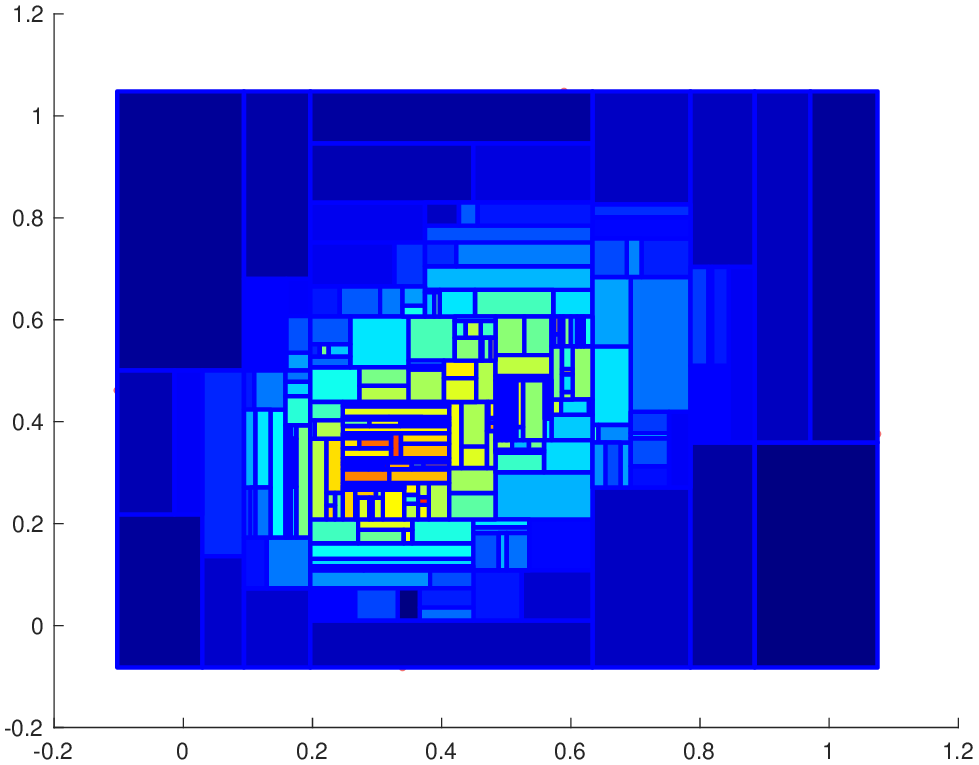}}
	\caption{2-D Gaussian mixtures: Adaptive partitions and density estimators produced by DSP-mix, MSP and DSP with $N = 1\times 10^4$ and $\theta=0.2$.}
	\label{2dGaussmix fig}
\end{figure}

Figure \ref{2dGaussmix fig} presents the numerical solutions of DSP, DSP-mix and MSP. All of them possess the capability to capture the distribution structure. They feature fine grids at positions where the probability density changes significantly and sparse grids in unimportant regions. 

From Figure \ref{gauss_low}, the V-shape curves with respect to $\theta$ are still observed in both error metrics. Therefore, it is suggested to avoid using too small parameter $\theta$ unless the sample size is large enough.

\begin{figure}[H]
\centering
\subfigure[KL divergence (left: DSP, middle: DSP-mix, right: MSP). ]{
{\includegraphics[width=0.32\textwidth,height=0.24\textwidth]{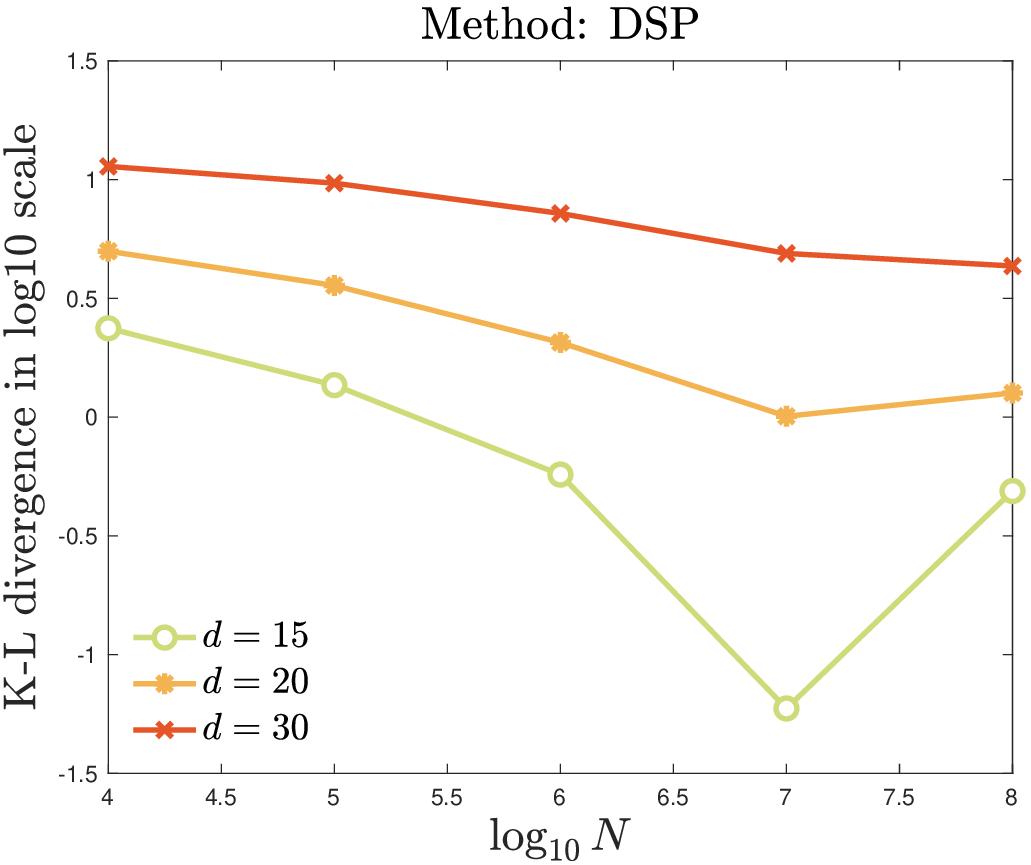}}
{\includegraphics[width=0.32\textwidth,height=0.24\textwidth]{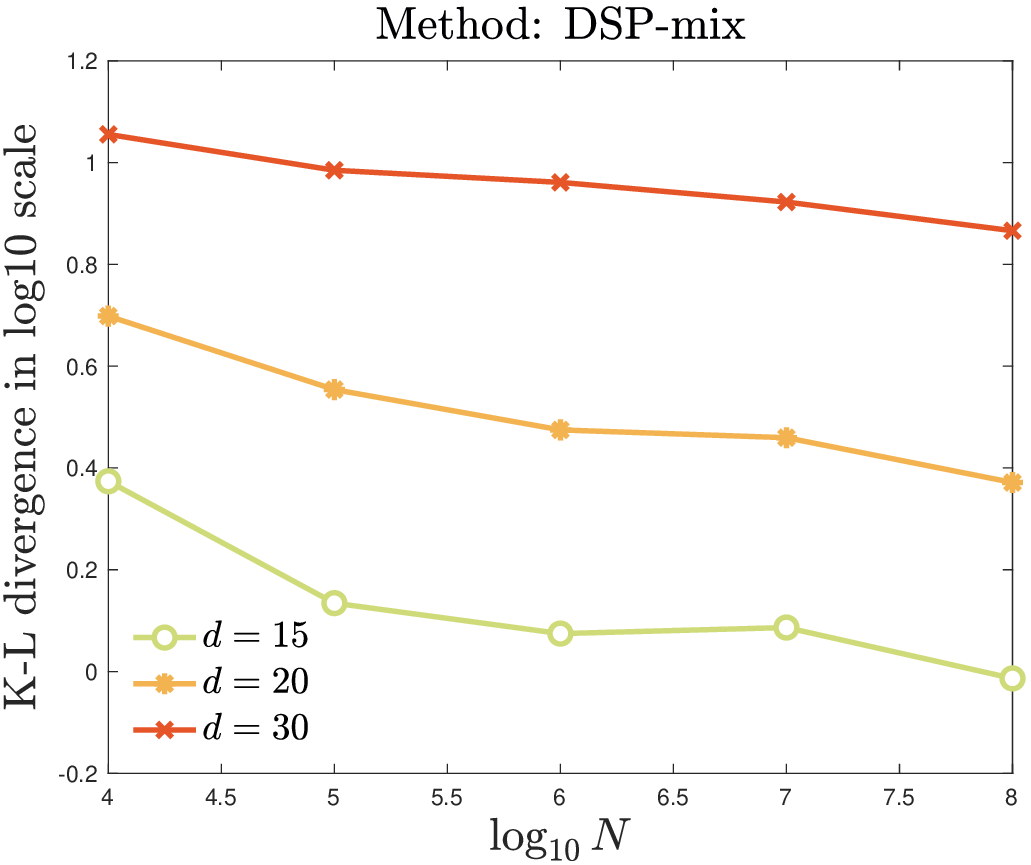}}
{\includegraphics[width=0.32\textwidth,height=0.24\textwidth]{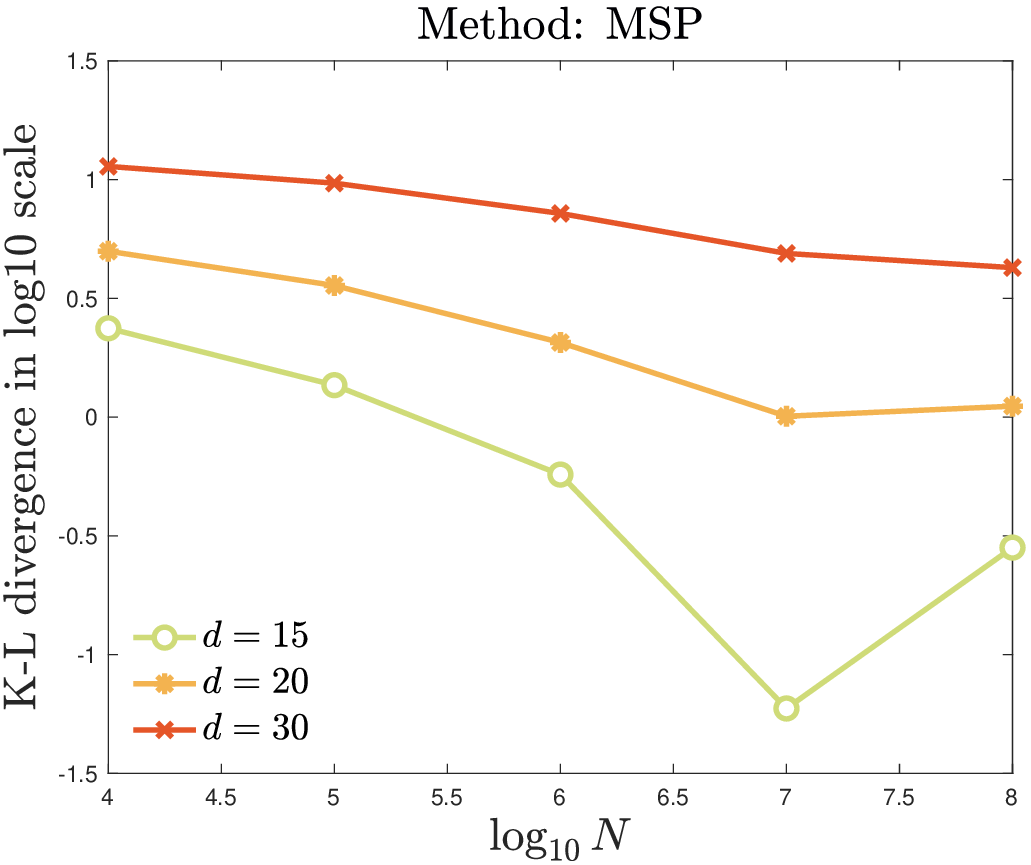}}}
\\
\centering
\subfigure[Hellinger distance (left: DSP, middle: DSP-mix, right: MSP). ]{
{\includegraphics[width=0.32\textwidth,height=0.24\textwidth]{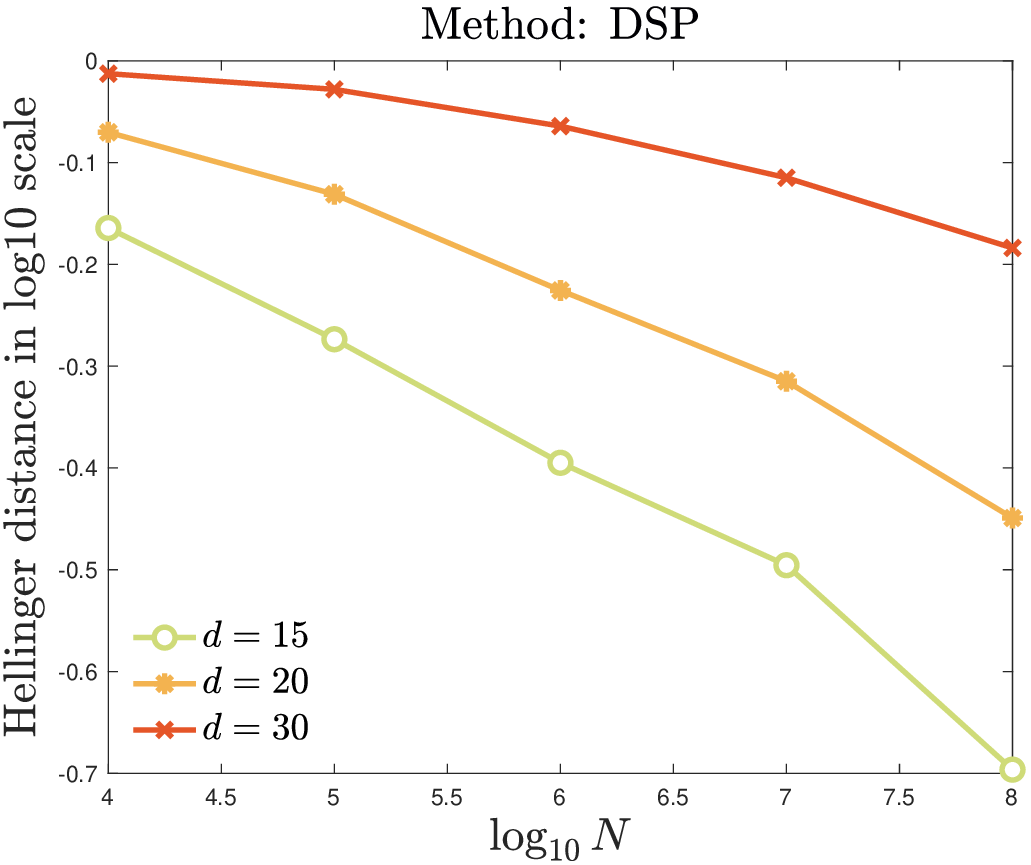}}
{\includegraphics[width=0.32\textwidth,height=0.24\textwidth]{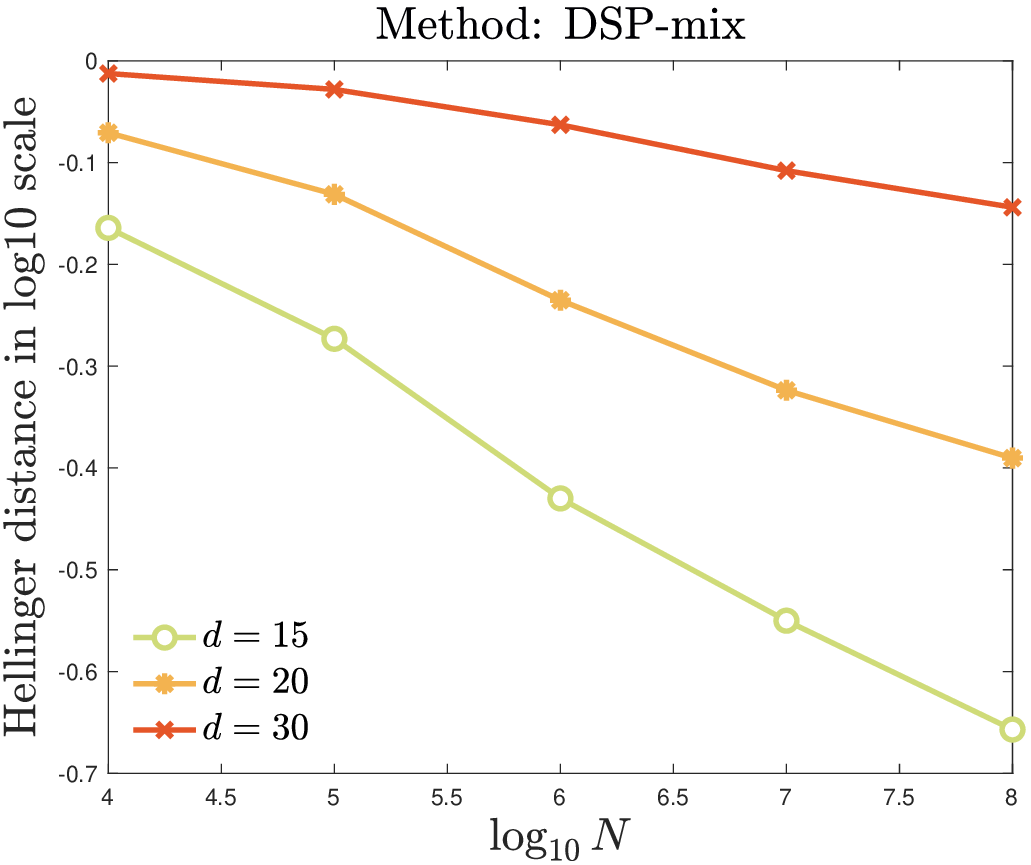}}
{\includegraphics[width=0.32\textwidth,height=0.24\textwidth]{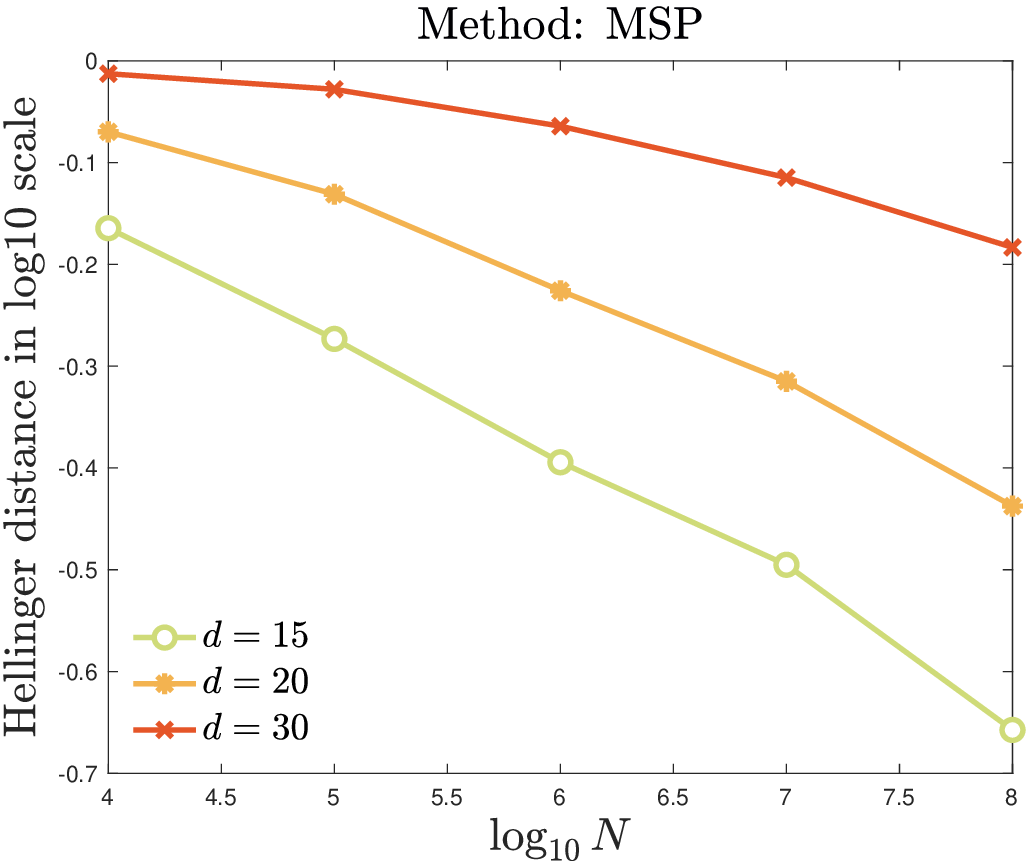}}}
\\
\centering
\subfigure[Partition level (left: DSP, middle: DSP-mix, right: MSP). ]{
{\includegraphics[width=0.32\textwidth,height=0.24\textwidth]{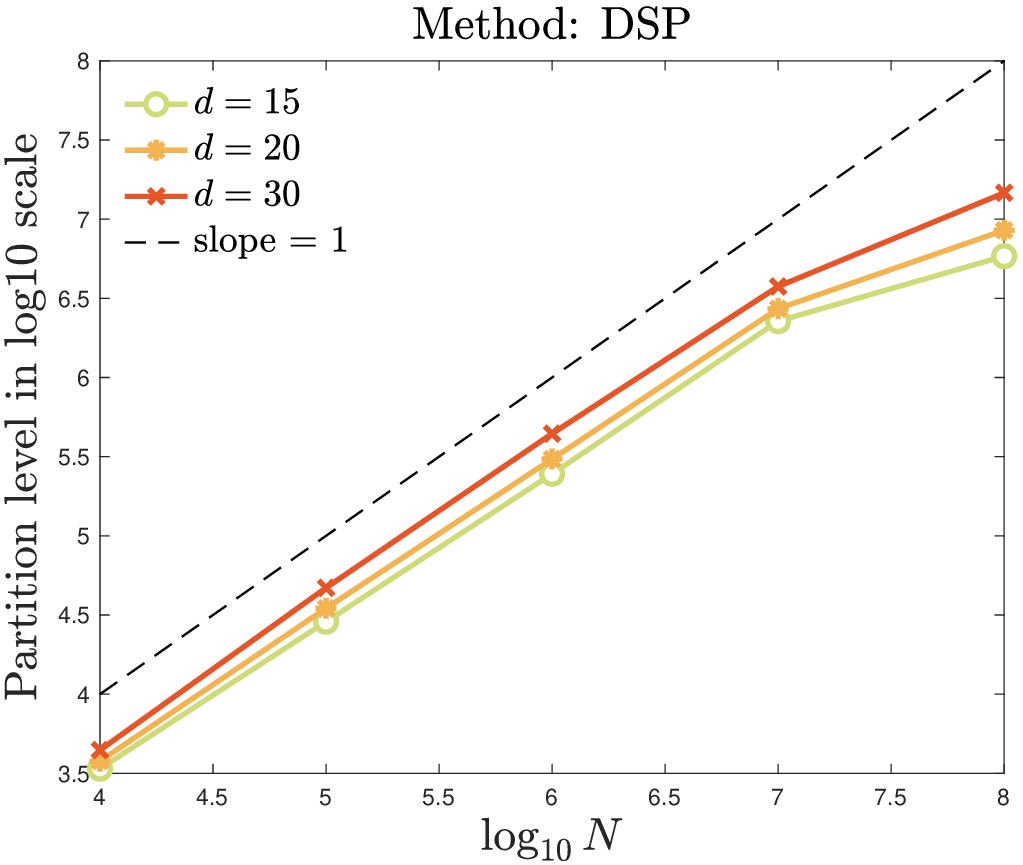}}
{\includegraphics[width=0.32\textwidth,height=0.24\textwidth]{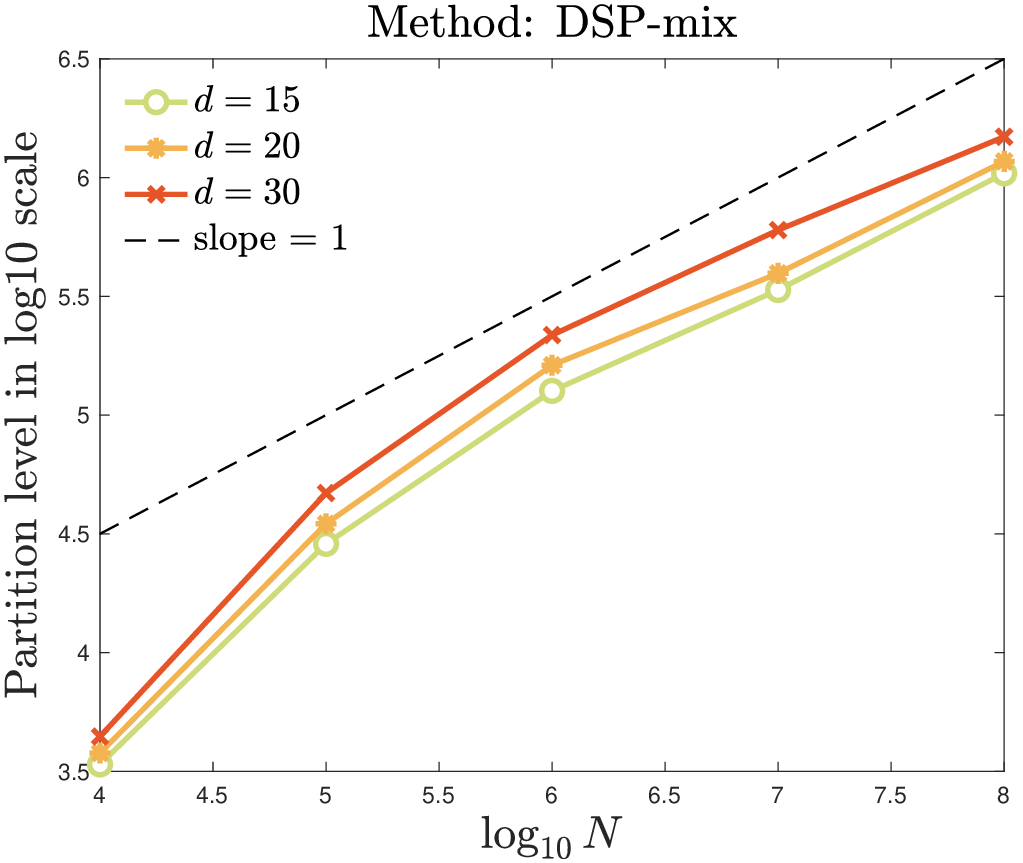}}
{\includegraphics[width=0.32\textwidth,height=0.24\textwidth]{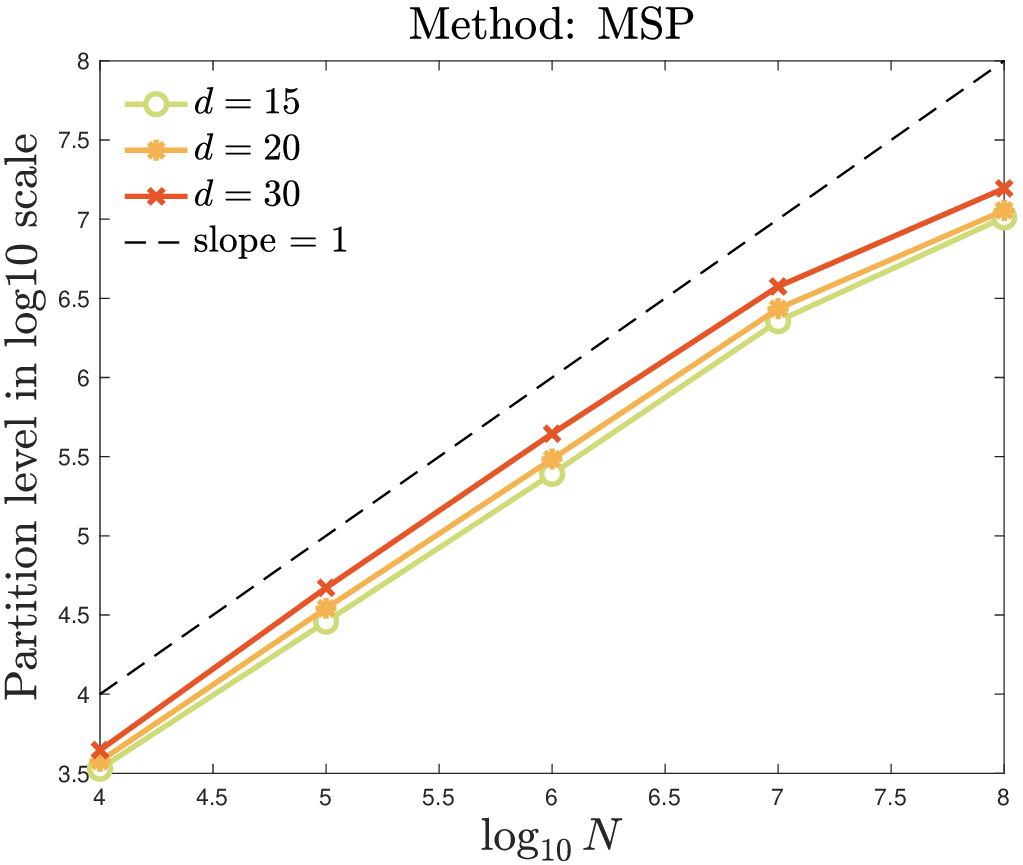}}}
\caption{\small $d$-D Gaussian mixtures: Convergence of DSP, DSP-mix and MSP with respect to $N$ for $d=15, 20, 30$.} \label{hdgauss_high_dimensional}
\end{figure}

\begin{figure}[H]
\centering
\subfigure[Partition level ($N = 10^4$) (left: DSP, middle: DSP-mix, right: MSP). ]{
{\includegraphics[width=0.32\textwidth,height=0.24\textwidth]{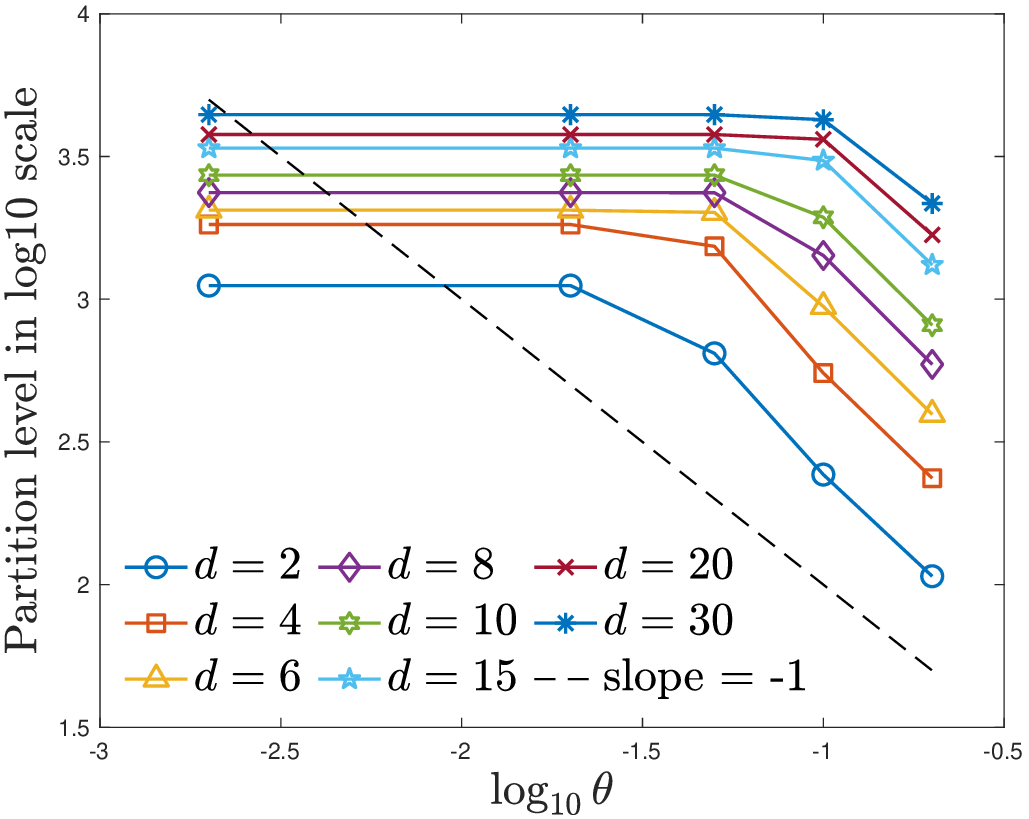}}
{\includegraphics[width=0.32\textwidth,height=0.24\textwidth]{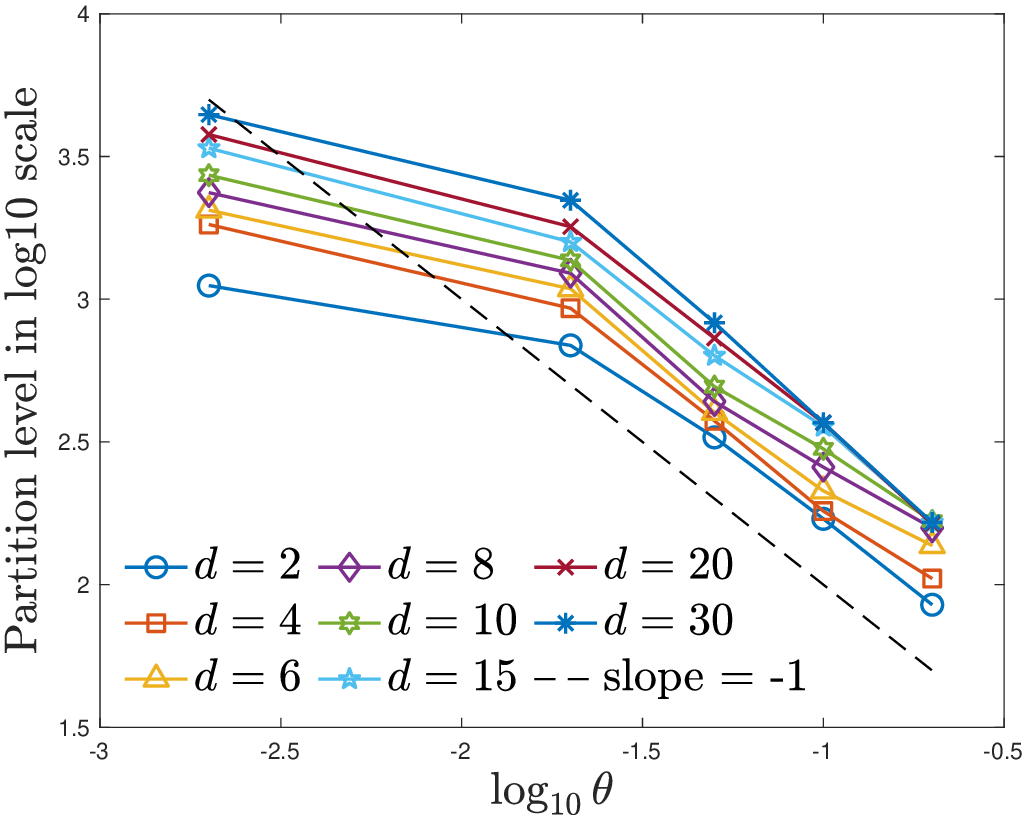}}
{\includegraphics[width=0.32\textwidth,height=0.24\textwidth]{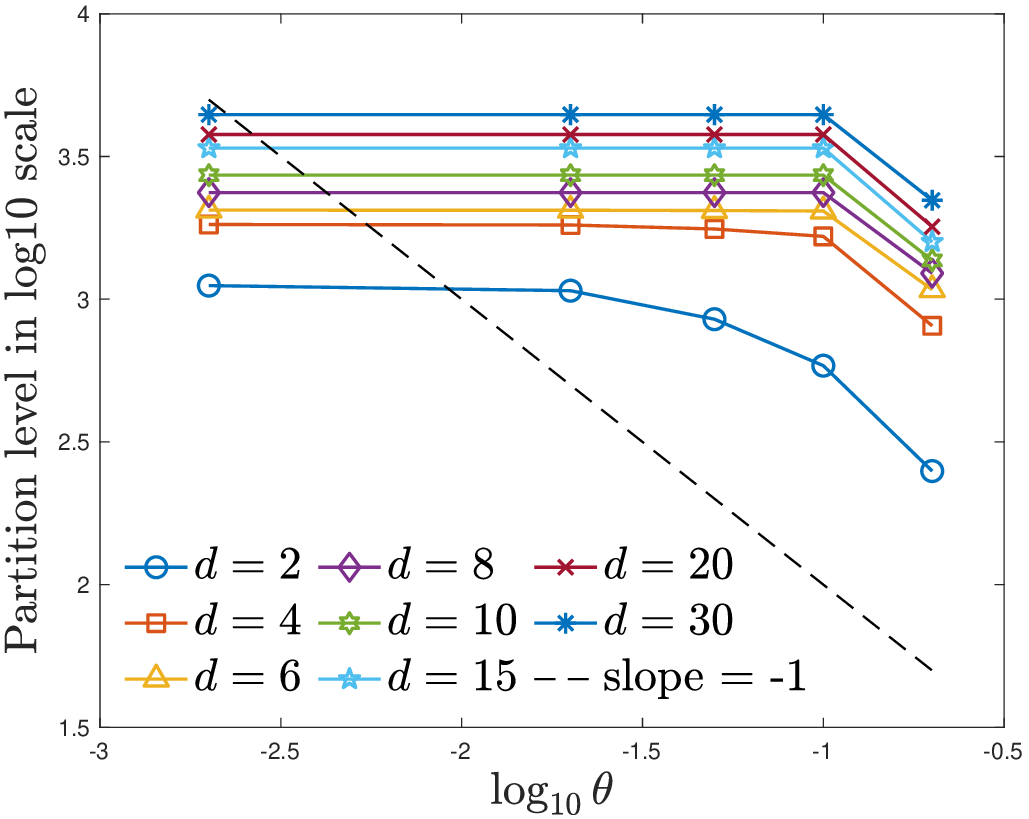}}}
\\
\centering
\subfigure[Partition level ($N = 10^5$) (left: DSP, middle: DSP-mix, right: MSP). ]{
{\includegraphics[width=0.32\textwidth,height=0.24\textwidth]{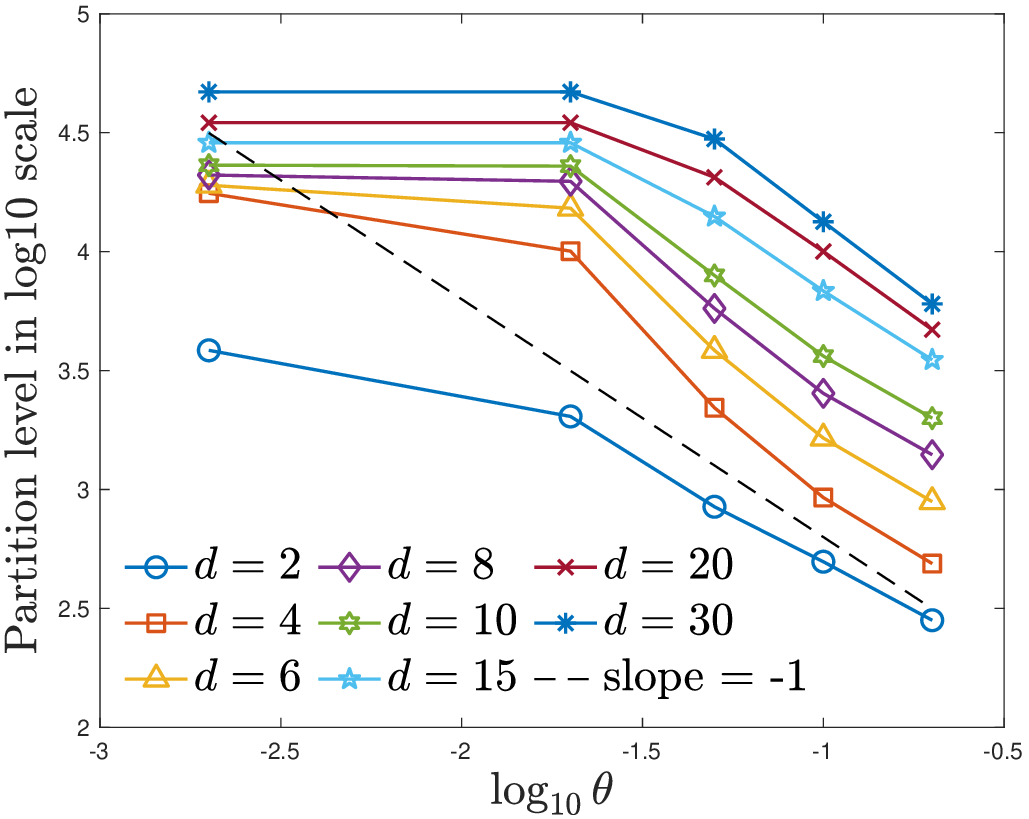}}
{\includegraphics[width=0.32\textwidth,height=0.24\textwidth]{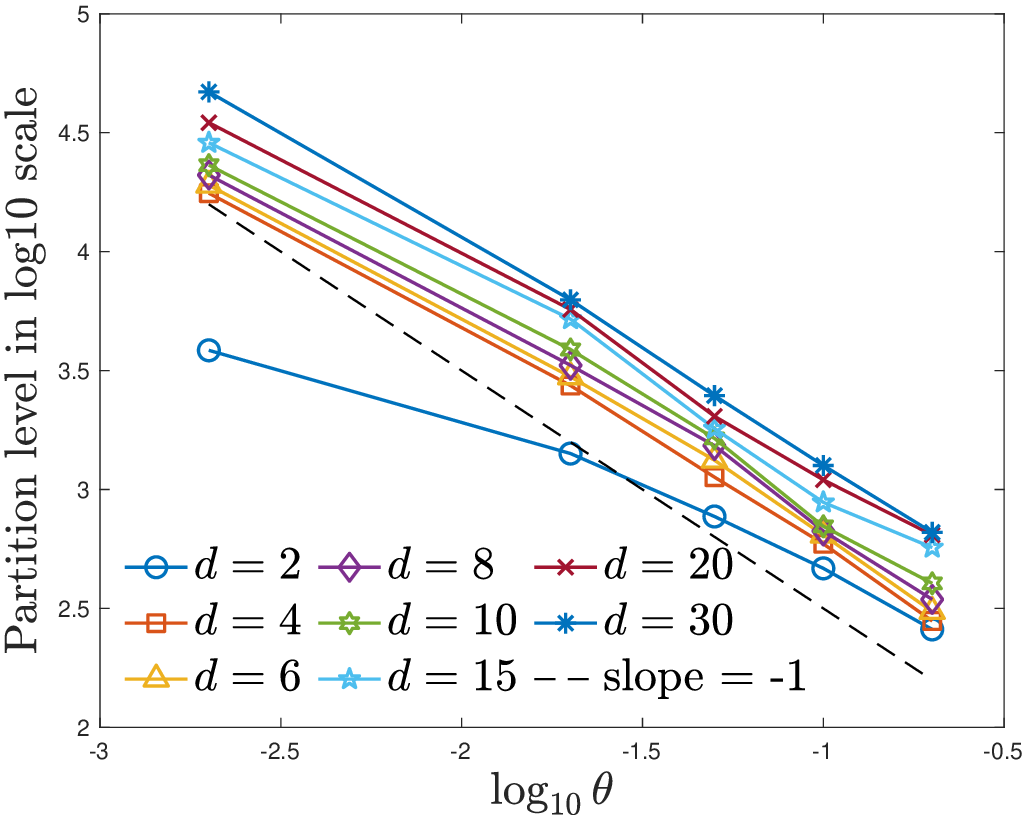}}
{\includegraphics[width=0.32\textwidth,height=0.24\textwidth]{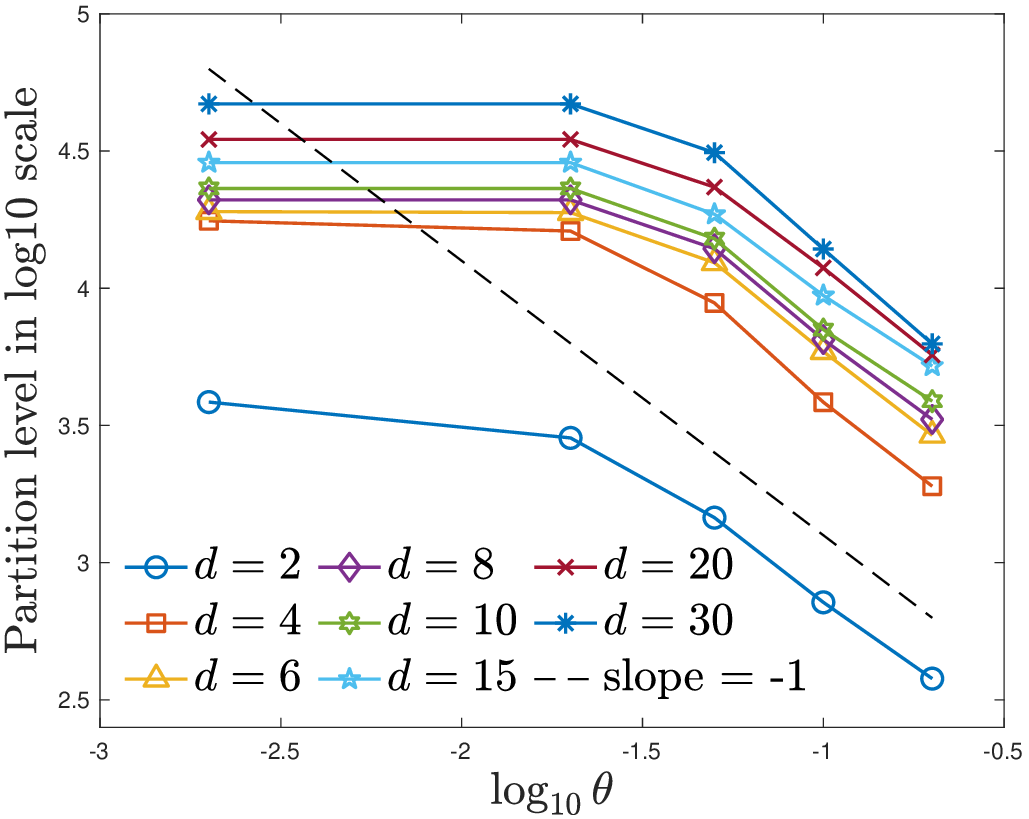}}}
\\
\centering
\subfigure[Partition level ($N = 10^6$) (left: DSP, middle: DSP-mix, right: MSP). ]{
{\includegraphics[width=0.32\textwidth,height=0.24\textwidth]{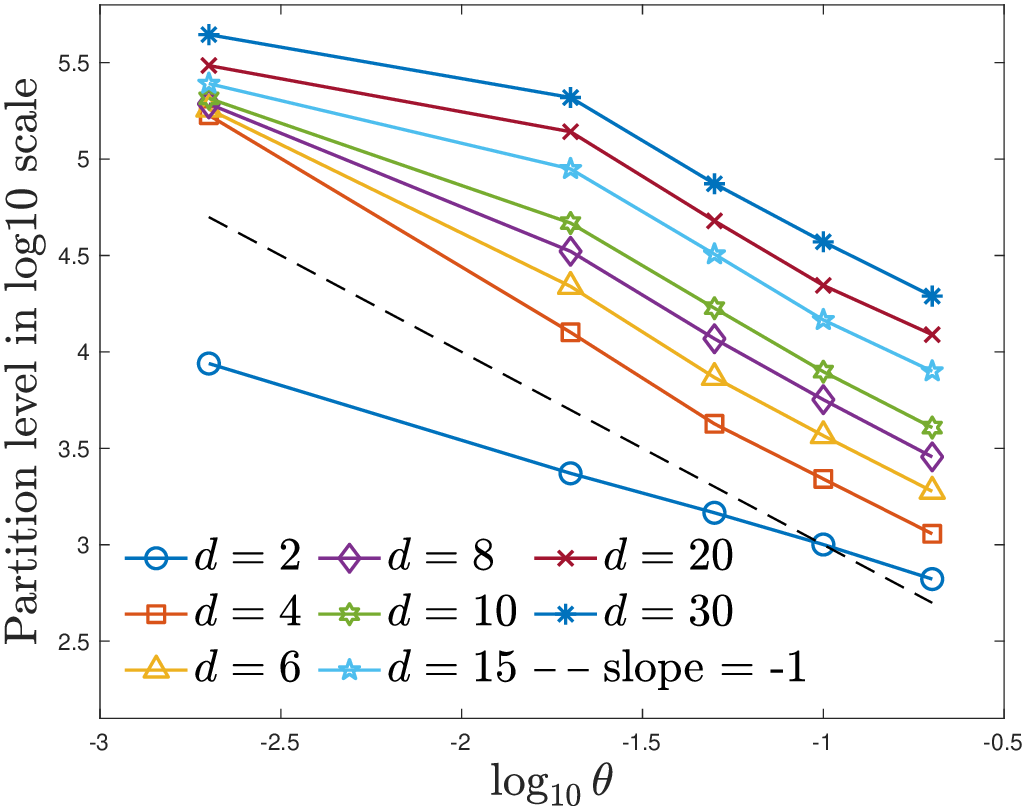}}
{\includegraphics[width=0.32\textwidth,height=0.24\textwidth]{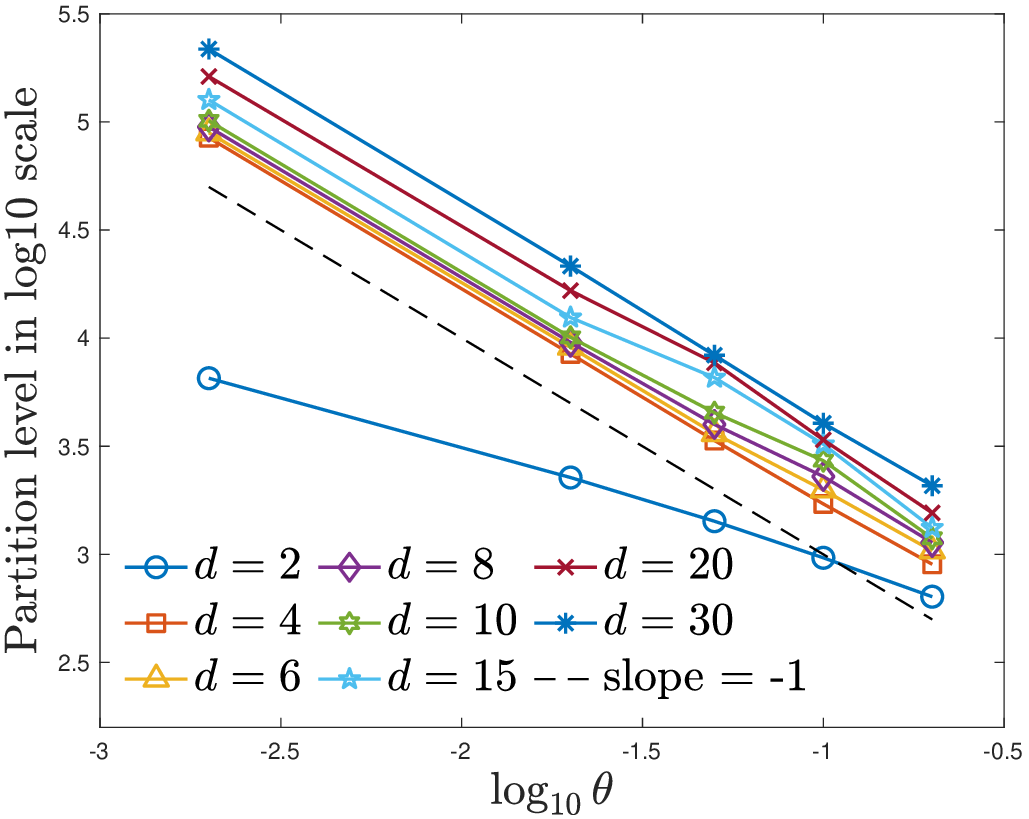}}
{\includegraphics[width=0.32\textwidth,height=0.24\textwidth]{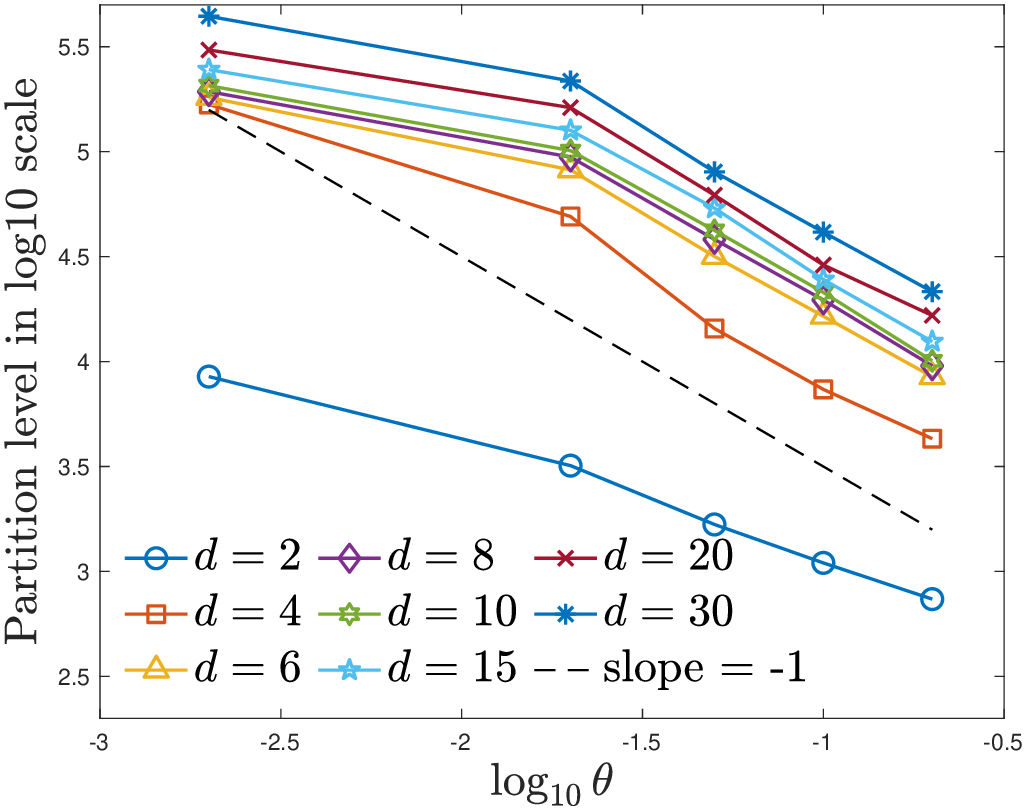}}}
\\
\centering
\subfigure[Partition level ($N = 10^7$) (left: DSP, middle: DSP-mix, right: MSP). ]{
{\includegraphics[width=0.32\textwidth,height=0.24\textwidth]{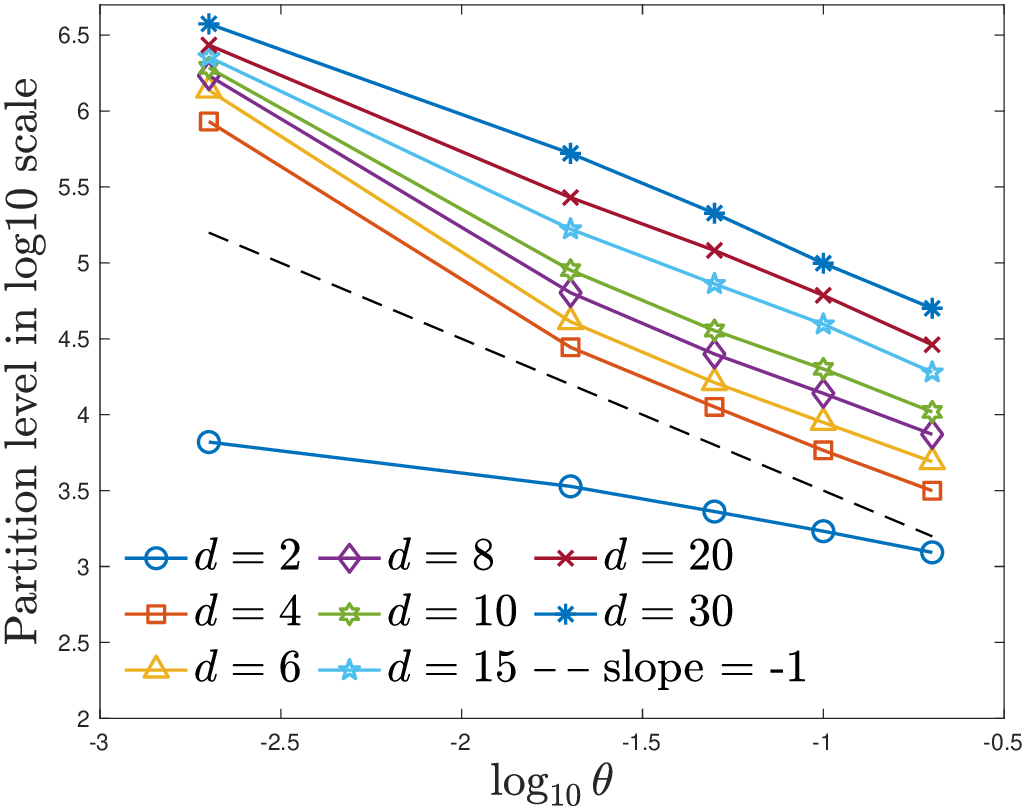}}
{\includegraphics[width=0.32\textwidth,height=0.24\textwidth]{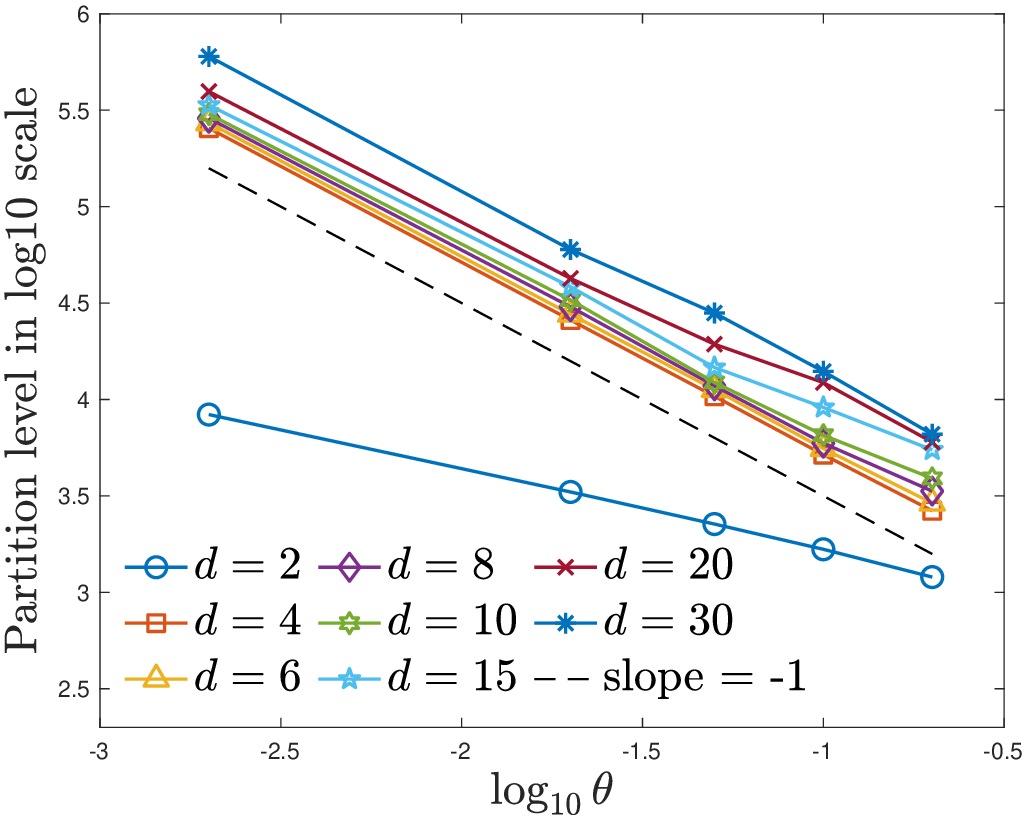}}
{\includegraphics[width=0.32\textwidth,height=0.24\textwidth]{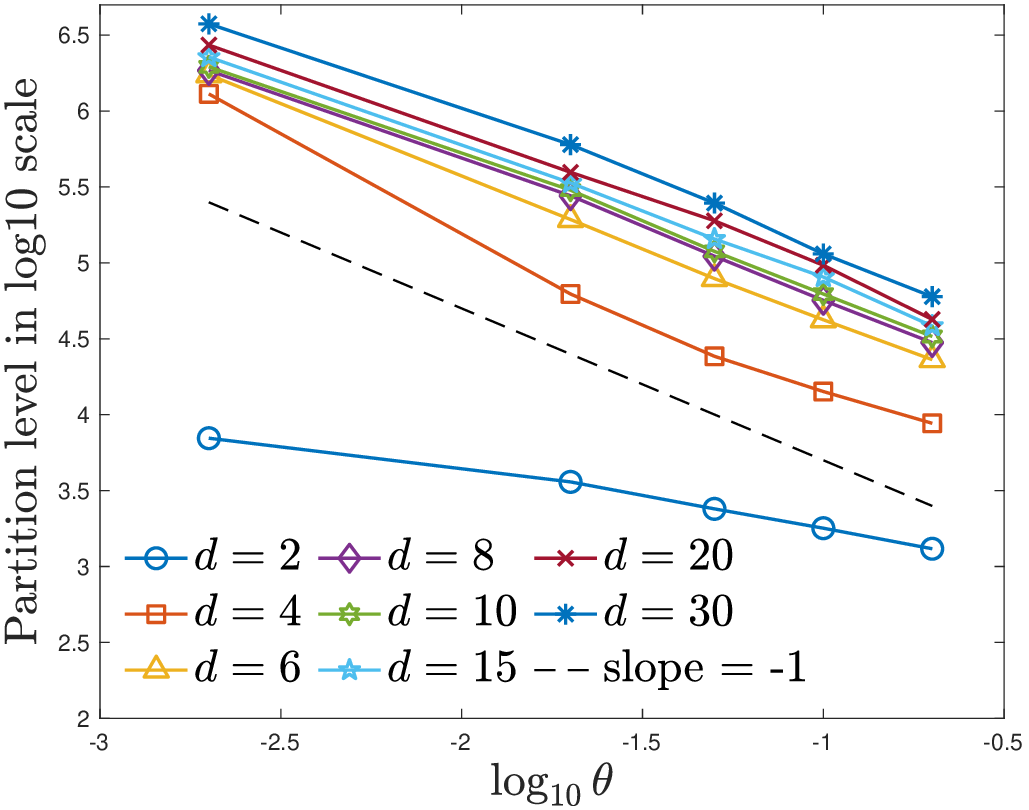}}}
\caption{\small $d$-D Gaussian mixture: The partition level $L$ under different $\theta$, $N$ and $d$. \label{hdgauss_partition}}
\end{figure}

For high-dimensional tests under $d = 15, 20, 30$,  Figure \ref{hdgauss_high_dimensional} presents the convergence of error metrics with respect to sample size $N = 10^4, 10^5, 10^6, 10^7, 10^8$, with the raw data collected in Tables \ref{hdGauss_error_N1e7} and \ref{hdGauss_high}. Figure \ref{hdgauss_partition} plots the partition level under different $d$, $N$, with $\theta$ fixed to be $0.002$, with the raw data collected in Tables \ref{hdGauss_K_time_N1e7} and \ref{hdGauss_K_time_high}.

For the rapidly decreasing Gaussian distribution, the performance of DSP, DSP-mix and MSP is very similar to those for the localized Beta mixtures under  different dimensions. It is still observed that the accuracy of all density estimators can be improved by increasing either sample size $N$ or increasing the partition level $L$. In terms of running time, DSP-mix and MSP are significantly faster and can generally achieve a tenfold or even twenty-fold speedup for large sample size ($N \ge 10^7$). Moreover, the trend $L \sim \theta^{-1}$ is still observed in all cases with only exception $d=2$. 

\begin{table}[H]
	\centering
	\caption{$d$-D Gaussian mixtures: The KL divergence and Hellinger distance  under different $d$ and $\theta$. The sample size is fixed to be $N = 10^7$.}
	\setlength{\tabcolsep}{4pt}
	\begin{tabular}{c|c|cc|cc|cc|cc|cc}
		\toprule
		& $\theta$	& \multicolumn{2}{c}{$0.002$} & \multicolumn{2}{c}{$0.02$} & \multicolumn{2}{c}{$0.05$} &  \multicolumn{2}{c}{$0.1$} &  \multicolumn{2}{c}{$0.2$}\\ 
		\midrule
		 $d$	&	Method	& KL & $\hat{H}^2$ &  KL & $\hat{H}^2$  & KL & $\hat{H}^2$  & KL & $\hat{H}^2$  & KL & $\hat{H}^2$  \\
		 \midrule
		 \multirow{3}{*}{2}	&	DSP	&	-0.0008 	&	0.0005 	&	0.0014 	&	0.0007 	&	0.0020 	&	0.0008 	&	0.0030 	&	0.0010 	&	0.0042 	&	0.0012 	\\
&	DSP-mix	&	0.0002 	&	0.0009 	&	0.0020 	&	0.0009 	&	0.0027 	&	0.0010 	&	0.0051 	&	0.0016 	&	0.0095 	&	0.0028 	\\
&	MSP	&	-0.0009 	&	0.0005 	&	0.0012 	&	0.0007 	&	0.0018 	&	0.0007 	&	0.0025 	&	0.0009 	&	0.0034 	&	0.0011 	\\
						& (maxSD) &      0.0001 	&	0.0000 	&	0.0001 	&	0.0000 	&	0.0001 	&	0.0000 	&	0.0002 	&	0.0001 	&	0.0003 	&	0.0001 		      \\							
		 \midrule
		 \multirow{3}{*}{4}	&	DSP	&	0.9405 	&	0.0551 	&	0.0046 	&	0.0074 	&	0.0187 	&	0.0100 	&	0.0356 	&	0.0128 	&	0.0510 	&	0.0162 	\\
&	DSP-mix	&	-0.0621 	&	0.0173 	&	0.0125 	&	0.0092 	&	0.0338 	&	0.0128 	&	0.0597 	&	0.0195 	&	0.0852 	&	0.0250 	\\
&	MSP	&	0.8370 	&	0.0798 	&	-0.0118 	&	0.0076 	&	0.0062 	&	0.0076 	&	0.0164 	&	0.0090 	&	0.0262 	&	0.0107 	\\
						& (maxSD) &       3.4708 	&	0.0026 	&	0.0189 	&	0.0026 	&	0.0284 	&	0.0026 	&	0.0294 	&	0.0026 	&	0.0292 	&	0.0027 		     \\	
		 \midrule
		 \multirow{3}{*}{6}	&	DSP	&	-0.3613 	&	0.0969 	&	0.0830 	&	0.0284 	&	0.1336 	&	0.0368 	&	0.1671 	&	0.0437 	&	0.2146 	&	0.0547 	\\
&	DSP-mix	&	-0.0253 	&	0.0304 	&	0.1213 	&	0.0348 	&	0.2003 	&	0.0510 	&	0.2791 	&	0.0669 	&	0.3264 	&	0.0765 	\\
&	MSP	&	-0.4532 	&	0.1162 	&	-0.0007 	&	0.0271 	&	0.0608 	&	0.0273 	&	0.0960 	&	0.0308 	&	0.1261 	&	0.0355 	\\	
						& (maxSD) &        0.0003 	&	0.0004 	&	0.0008 	&	0.0004 	&	0.0023 	&	0.0006 	&	0.0029 	&	0.0008 	&	0.0034 	&	0.0008 		    \\		
		 \midrule
		 \multirow{3}{*}{8}	&	DSP	&	-0.4183 	&	0.1383 	&	0.2277 	&	0.0676 	&	0.3186 	&	0.0817 	&	0.3812 	&	0.0935 	&	0.4916 	&	0.1177 	\\
&	DSP-mix	&	0.0799 	&	0.0626 	&	0.3278 	&	0.0832 	&	0.5159 	&	0.1182 	&	0.6375 	&	0.1377 	&	0.7058 	&	0.1510 	\\
&	MSP	&	-0.4558 	&	0.1455 	&	0.0833 	&	0.0623 	&	0.2030 	&	0.0667 	&	0.2674 	&	0.0734 	&	0.3281 	&	0.0831 	\\
						& (maxSD) &       0.0011 	&	0.0006 	&	0.0025 	&	0.0008 	&	0.0036 	&	0.0006 	&	0.0047 	&	0.0010 	&	0.0058 	&	0.0013 		     \\	
		 \midrule
		 \multirow{3}{*}{10}	&	DSP	&	-0.3841 	&	0.1836 	&	0.4510 	&	0.1210 	&	0.5888 	&	0.1395 	&	0.7002 	&	0.1602 	&	0.9045 	&	0.1979 	\\
&	DSP-mix	&	0.2744 	&	0.1115 	&	0.6628 	&	0.1523 	&	0.9963 	&	0.2029 	&	1.1433 	&	0.2224 	&	1.2483 	&	0.2418 	\\
&	MSP	&	-0.3919 	&	0.1848 	&	0.2745 	&	0.1113 	&	0.4514 	&	0.1217 	&	0.5395 	&	0.1313 	&	0.6628 	&	0.1523 	\\
						& (maxSD) &       0.0021 	&	0.0006 	&	0.0036 	&	0.0008 	&	0.0061 	&	0.0014 	&	0.0122 	&	0.0015 	&	0.0118 	&	0.0017 		     \\	
		 \midrule
		 \multirow{3}{*}{15}	&	DSP	&	0.0593 	&	0.3195 	&	1.4166 	&	0.2948 	&	1.7380 	&	0.3296 	&	2.1080 	&	0.3816 	&	2.5918 	&	0.4280 	\\
&	DSP-mix	&	1.2203 	&	0.2818 	&	2.2709 	&	0.3894 	&	2.8796 	&	0.4337 	&	3.1056 	&	0.4576 	&	3.5545 	&	0.5211 	\\
&	MSP	&	0.0593 	&	0.3198 	&	1.2203 	&	0.2817 	&	1.5263 	&	0.3001 	&	1.7615 	&	0.3296 	&	2.2709 	&	0.3894 	\\
						& (maxSD) &     0.0053 	&	0.0015 	&	0.0081 	&	0.0014 	&	0.0205 	&	0.0027 	&	0.0264 	&	0.0022 	&	0.0295 	&	0.0028 		       \\	
		 \midrule
		 \multirow{3}{*}{20}	&	DSP	&	1.0081 	&	0.4844 	&	3.0613 	&	0.4864 	&	3.7281 	&	0.5460 	&	4.4446 	&	0.5987 	&	5.1962 	&	0.6306 	\\
&	DSP-mix	&	2.8806 	&	0.4746 	&	4.9634 	&	0.6092 	&	5.6702 	&	0.6371 	&	6.1985 	&	0.6827 	&	7.3848 	&	0.7647 	\\
&	MSP	&	1.0081 	&	0.4843 	&	2.8806 	&	0.4748 	&	3.3849 	&	0.5061 	&	4.0691 	&	0.5638 	&	4.9634 	&	0.6094 	\\	
						& (maxSD) &      0.0200 	&	0.0017 	&	0.0311 	&	0.0024 	&	0.0483 	&	0.0029 	&	0.0591 	&	0.0027 	&	0.0668 	&	0.0028 		      \\	
		 \midrule
		 \multirow{3}{*}{30}	&	DSP	&	4.8871 	&	0.7676 	&	8.5323 	&	0.7894 	&	10.1693 	&	0.8359 	&	11.5031 	&	0.8534 	&	12.6868 	&	0.8754 	\\
&	DSP-mix	&	8.3667 	&	0.7800 	&	12.4234 	&	0.8571 	&	13.8661 	&	0.8963 	&	15.4386 	&	0.9284 	&	17.0640 	&	0.9359 	\\
&	MSP	&	4.8871 	&	0.7679 	&	8.3667 	&	0.7805 	&	9.9671 	&	0.8262 	&	11.3469 	&	0.8446 	&	12.4234 	&	0.8570 	\\
						& (maxSD) &      0.0320 	&	0.0012 	&	0.0418 	&	0.0014 	&	0.0426 	&	0.0017 	&	0.0555 	&	0.0015 	&	0.0612 	&	0.0013 		      \\																			
		\bottomrule
	\end{tabular}
	\label{hdGauss_error_N1e7}
\end{table}

\begin{table}[H]
	\centering
	\caption{$d$-D Gaussian mixtures: The partition level $L$ and computational time (in seconds) under different $d$ and $\theta$. The sample size is fixed to be $N  = 10^7$.}
	\setlength{\tabcolsep}{4pt}
	\begin{tabular}{c|c|cc|cc|cc|cc|cc}
		\toprule
		& $\theta$	& \multicolumn{2}{c}{$0.002$} & \multicolumn{2}{c}{$0.02$} & \multicolumn{2}{c}{$0.05$} &  \multicolumn{2}{c}{$0.1$} &  \multicolumn{2}{c}{$0.2$}\\ 
		\midrule
		 $d$	&	Method	& $L$ & time & $L$ & time  & $L$ & time  & $L$ & time  & $L$ & time \\
		 \midrule
		 \multirow{3}{*}{2}	&	DSP	&	6603 	&	7.88 	&	3372 	&	13.35 	&	2302 	&	18.66 	&	1708 	&	27.79 	&	1239 	&	45.49 	\\
&	DSP-mix	&	8353 	&	6.92 	&	3315 	&	7.38 	&	2260 	&	9.03 	&	1672 	&	11.67 	&	1200 	&	20.86 	\\
&	MSP	&	7001 	&	7.28 	&	3608 	&	8.83 	&	2397 	&	9.03 	&	1788 	&	8.99 	&	1309 	&	9.18 	\\						
		 \midrule
		 \multirow{3}{*}{4}	&	DSP	&	853559 	&	882.76 	&	27793 	&	269.96 	&	11255 	&	270.44 	&	5823 	&	296.21 	&	3155 	&	323.50 	\\
&	DSP-mix	&	255266 	&	21.33 	&	25823 	&	25.91 	&	10390 	&	47.30 	&	5178 	&	81.74 	&	2649 	&	149.73 	\\
&	MSP	&	1296208 	&	32.90 	&	62370 	&	17.15 	&	24249 	&	14.46 	&	14210 	&	14.08 	&	8777 	&	13.24 	\\
		 \midrule
		 \multirow{3}{*}{6}	&	DSP	&	1372339 	&	1109.13 	&	40997 	&	426.41 	&	16269 	&	385.83 	&	8894 	&	398.37 	&	4900 	&	429.82 	\\
&	DSP-mix	&	271883 	&	25.15 	&	27674 	&	36.71 	&	11259 	&	67.65 	&	5568 	&	118.32 	&	2902 	&	215.97 	\\
&	MSP	&	1732952 	&	41.91 	&	193079 	&	26.18 	&	78787 	&	22.51 	&	42202 	&	20.70 	&	23052 	&	19.07 	\\		
		 \midrule
		 \multirow{3}{*}{8}	&	DSP	&	1711429 	&	937.34 	&	63630 	&	613.12 	&	25090 	&	523.57 	&	13821 	&	512.66 	&	7422 	&	538.91 	\\
&	DSP-mix	&	287074 	&	28.99 	&	30227 	&	47.96 	&	11765 	&	88.58 	&	5933 	&	152.11 	&	3350 	&	272.48 	\\
&	MSP	&	1850228 	&	49.01 	&	276904 	&	34.20 	&	110642 	&	30.05 	&	56644 	&	27.22 	&	29844 	&	24.78 	\\
		 \midrule
		 \multirow{3}{*}{10}	&	DSP	&	1927372 	&	798.32 	&	89458 	&	774.72 	&	35850 	&	633.31 	&	20018 	&	606.81 	&	10431 	&	634.64 	\\
&	DSP-mix	&	301812 	&	36.70 	&	32765 	&	58.30 	&	12257 	&	107.54 	&	6529 	&	181.35 	&	3899 	&	326.22 	\\
&	MSP	&	1955139 	&	56.54 	&	301284 	&	38.56 	&	119087 	&	34.70 	&	62357 	&	31.48 	&	32749 	&	29.19 	\\	
		 \midrule
		 \multirow{3}{*}{15}	&	DSP	&	2268841 	&	663.98 	&	166721 	&	988.71 	&	72700 	&	825.14 	&	39392 	&	851.34 	&	19019 	&	800.97 	\\
&	DSP-mix	&	336290 	&	55.35 	&	38372 	&	85.67 	&	14657 	&	149.58 	&	9097 	&	243.25 	&	5480 	&	464.71 	\\
&	MSP	&	2268874 	&	78.76 	&	336290 	&	54.82 	&	143598 	&	49.61 	&	80693 	&	45.85 	&	38372 	&	41.86 	\\	
		 \midrule
		 \multirow{3}{*}{20}	&	DSP	&	2721235 	&	639.02 	&	269423 	&	1184.09 	&	120788 	&	1148.42 	&	61092 	&	1131.69 	&	28893 	&	914.53 	\\
&	DSP-mix	&	395153 	&	77.47 	&	42435 	&	111.01 	&	19350 	&	185.18 	&	12237 	&	309.70 	&	6016 	&	613.64 	\\
&	MSP	&	2721235 	&	105.96 	&	395153 	&	73.37 	&	189409 	&	66.72 	&	96443 	&	60.69 	&	42435 	&	55.44 	\\	
		 \midrule
		 \multirow{3}{*}{30}	&	DSP	&	3754332 	&	734.10 	&	526282 	&	1923.58 	&	212475 	&	1829.85 	&	99414 	&	1466.28 	&	50224 	&	1479.35 	\\
&	DSP-mix	&	601142 	&	115.55 	&	59909 	&	159.59 	&	28074 	&	265.94 	&	13985 	&	458.06 	&	6604 	&	843.57 	\\
&	MSP	&	3754333 	&	167.07 	&	601142 	&	113.35 	&	247671 	&	98.94 	&	114555 	&	89.74 	&	59909 	&	85.51 	\\										
		\bottomrule
	\end{tabular}
	\label{hdGauss_K_time_N1e7}
\end{table}

\begin{table}[H]
	\centering
	\caption{$d$-D Gaussian mixtures: The KL divergence and Hellinger distance  under different dimensions $d$ and sample size $N$. The parameter $\theta$ is fixed to be $0.002$.}
	\setlength{\tabcolsep}{4pt}
	\begin{tabular}{c|c|cc|cc|cc|cc|cc}
		\toprule
		& $N$	& \multicolumn{2}{c}{$1\times10^4$} & \multicolumn{2}{c}{$1\times10^5$} & \multicolumn{2}{c}{$1\times10^6$} &  \multicolumn{2}{c}{$1\times10^7$} &  \multicolumn{2}{c}{$1\times10^8$}\\ 
		\midrule
		 $d$	&	Method	& KL & $\hat{H}^2$ &  KL & $\hat{H}^2$  & KL & $\hat{H}^2$  & KL & $\hat{H}^2$  & KL & $\hat{H}^2$  \\
		 \midrule
		 \multirow{3}{*}{15}	&	DSP 		&	2.3652 	&	0.6851 	&	1.3628 	&	0.5327 	&	0.5726 	&	0.4028 	&	0.0593 	&	0.3195 	&	0.4881 	&	0.2012 	\\
		 				&	DSP-mix 	&	2.3652 	&	0.6853 	&	1.3628 	&	0.5332 	&	1.1873 	&	0.3716 	&	1.2203 	&	0.2818 	&	0.9694 	&	0.2203 	\\
						&	MSP		&	2.3652 	&	0.6848 	&	1.3628 	&	0.5331 	&	0.5726 	&	0.4033 	&	0.0593 	&	0.3198 	&	0.2821 	&	0.2201 	\\
		 \midrule
		 \multirow{3}{*}{20}	&	DSP 		&	4.9939 	&	0.8507 	&	3.5812 	&	0.7394 	&	2.0609 	&	0.5949 	&	1.0081 	&	0.4844 	&	1.2630 	&	0.3555 	\\
		 				&	DSP-mix	&	4.9939 	&	0.8499 	&	3.5812 	&	0.7393 	&	2.9840 	&	0.5818 	&	2.8806 	&	0.4746 	&	2.3519 	&	0.4071 	\\
						&	MSP		&	4.9939 	&	0.8516 	&	3.5812 	&	0.7395 	&	2.0609 	&	0.5945 	&	1.0081 	&	0.4843 	&	1.1115 	&	0.3652 	\\
		 \midrule
		 \multirow{3}{*}{30}	&	DSP 		&	11.3623 	&	0.9714 	&	9.6601 	&	0.9375 	&	7.2038 	&	0.8627 	&	4.8871 	&	0.7676 	&	4.3272 	&	0.6549 	\\
		 				&	DSP-mix	&	11.3623 	&	0.9717 	&	9.6601 	&	0.9374 	&	9.1441 	&	0.8652 	&	8.3667 	&	0.7800 	&	7.3474 	&	0.7180 	\\
						&	MSP		&	11.3623 	&	0.9713 	&	9.6601 	&	0.9376 	&	7.2038 	&	0.8626 	&	4.8871 	&	0.7679 	&	4.2576 	&	0.6557 	\\																		
		\bottomrule
	\end{tabular}
	\label{hdGauss_high}
\end{table}

\begin{table}[H]
	\centering
	\caption{$d$-D Gaussian mixtures: The partition level $L$ and computational time (in seconds) under different $d$ and $N$. The parameter $\theta$ is fixed to be $0.002$.}
	\setlength{\tabcolsep}{4pt}
	\begin{tabular}{c|c|cc|cc|cc|cc|cc}
		\toprule
		& $N$	& \multicolumn{2}{c}{$1\times10^4$} & \multicolumn{2}{c}{$1\times10^5$} & \multicolumn{2}{c}{$1\times10^6$} &  \multicolumn{2}{c}{$1\times10^7$} &  \multicolumn{2}{c}{$1\times10^8$}\\ 
		\midrule
		 $d$	&	Method	& $L$ & time & $L$ & time  & $L$ & time  & $L$ & time  & $L$ & time \\
		 \midrule
		 \multirow{3}{*}{15}	&	DSP 		&	3383 	&	0.03 	&	28693 	&	0.37 	&	246147 	&	5.42 	&	2268841 	&	663.98 	&	5838142	&	30802.07 	\\
		 				&	DSP-mix	&	3383 	&	0.03 	&	28693 	&	0.44 	&	126396 	&	5.11 	&	336290 	&	55.35 	&	1042368	&	803.32 	\\
						&	MSP		&	3383 	&	0.03 	&	28693 	&	0.31 	&	246147 	&	5.49 	&	2268874 	&	78.76 	&	10250539	&	752.33 	\\	
		 \midrule
		 \multirow{3}{*}{20}	&	DSP 		&	3775 	&	0.04 	&	34852 	&	0.57 	&	305394 	&	7.96 	&	2721235 	&	639.02 	&	8503962	&	34682.44 	\\
		 				&	DSP-mix	&	3775 	&	0.04 	&	34852 	&	0.64 	&	162165 	&	6.76 	&	395153 	&	77.47 	&	1170087	&	1076.32 	\\
						&	MSP		&	3775 	&	0.04 	&	34852 	&	0.48 	&	305394 	&	7.84 	&	2721235 	&	105.96 	&	11362133	&	882.67 	\\
		 \midrule
		 \multirow{3}{*}{30}	&	DSP 		&	4433 	&	0.07 	&	46950 	&	0.86 	&	441887 	&	13.93 	&	3754332 	&	734.10 	&	14664912	&	42827.37 	\\
		 				&	DSP-mix	&	4433 	&	0.07 	&	46950 	&	1.04 	&	217169 	&	11.25 	&	601142 	&	115.55 	&	1485038	&	1677.59 	\\
						&	MSP		&	4433 	&	0.07 	&	46950 	&	0.98 	&	441887 	&	13.49 	&	3754333 	&	167.07 	&	15638938	&	1439.79 	\\																	
		\bottomrule
	\end{tabular}
	\label{hdGauss_K_time_high}
\end{table}

\subsection{$d$-D Cauchy Mixtures}

Finally, we consider a much more challenging problem: Learning the density of the multidimensional Cauchy mixtures \cite{RuzgasLukauskasCepkauskas2021}
\begin{equation}
\bx_i \sim \sum_{j=1}^q p_j C(\bx, \bmm_j, \bu_j), \quad C(\bx, \bmm_j, \bu_j) = \prod_{k=1}^d \frac{u_{jk}}{\pi\left[u_{jk}^2 + (x_k - m_{jk})^2\right]}.
\end{equation}
All the parameters are set to be: $q= 4$, $p_1 = p_2 = p_3 = p_4 = 0.25$, $u_{jk} = 0.32$ for all $j, k$ and the centres $m_{1k} = -3$, $m_{2k} = -1$, $m_{3k} = 1$, $m_{4k} = 3$ for $k = 1, \dots, d$.
\begin{figure}[!h]
\centering
\subfigure[KL divergence ($d=2$) (left: DSP, middle: DSP-mix, right: MSP). ]{
{\includegraphics[width=0.32\textwidth,height=0.24\textwidth]{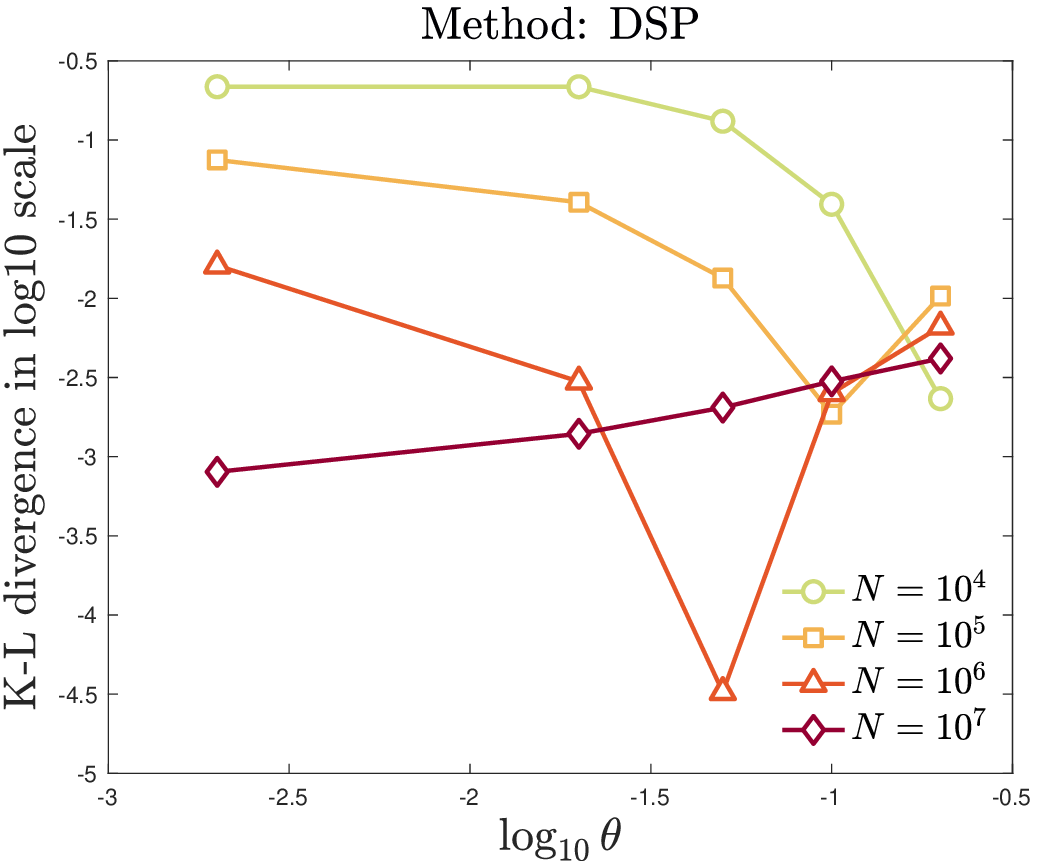}}
{\includegraphics[width=0.32\textwidth,height=0.24\textwidth]{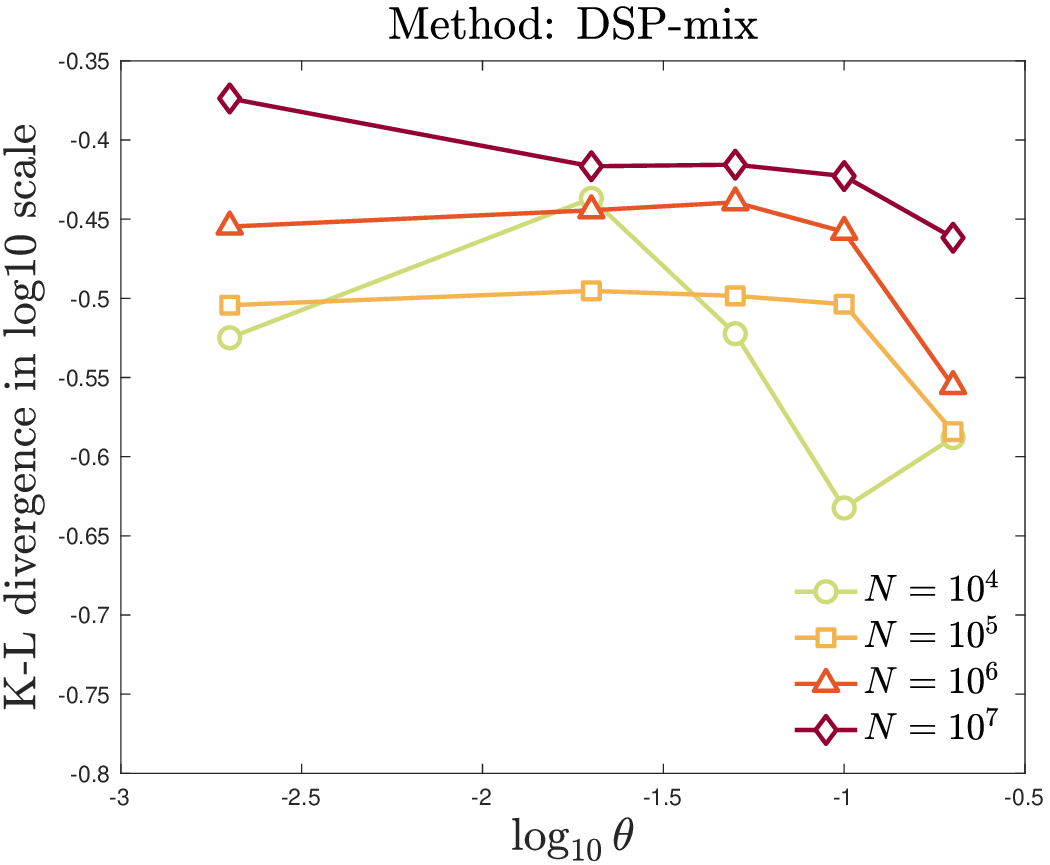}}
{\includegraphics[width=0.32\textwidth,height=0.24\textwidth]{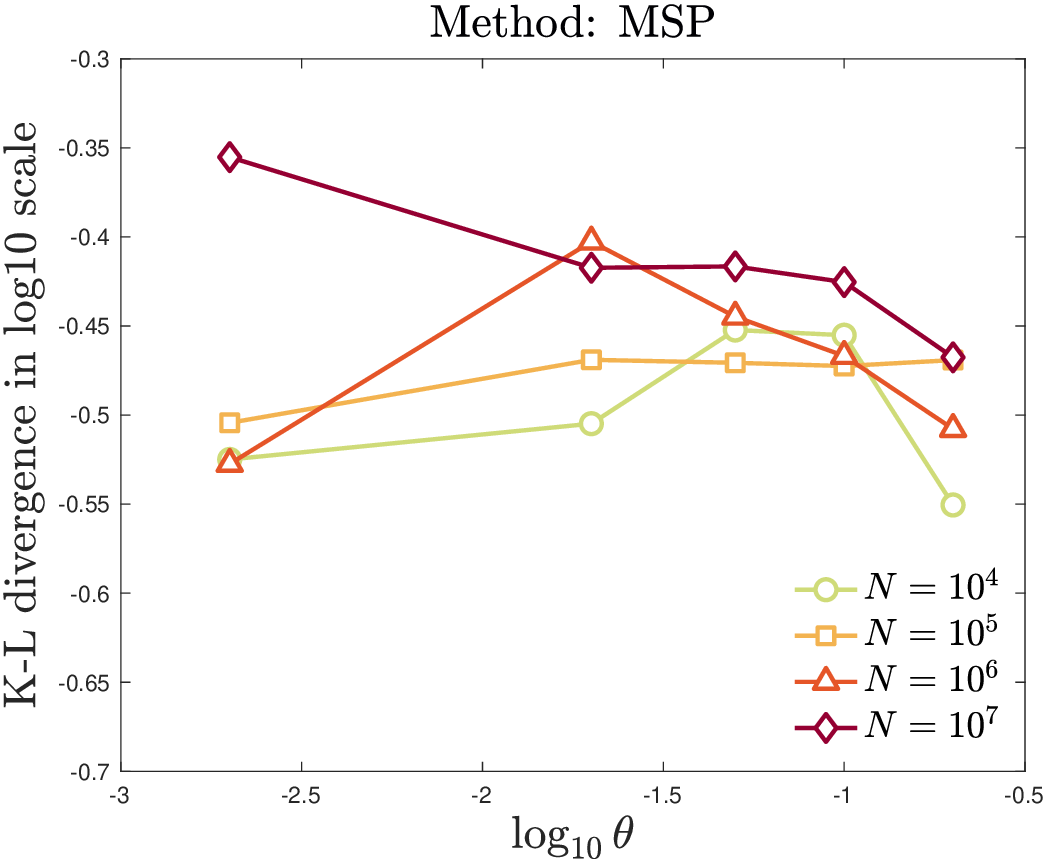}}}
\\
\centering
\subfigure[Hellinger distance ($d=2$) (left: DSP, middle: DSP-mix, right: MSP). ]{
{\includegraphics[width=0.32\textwidth,height=0.24\textwidth]{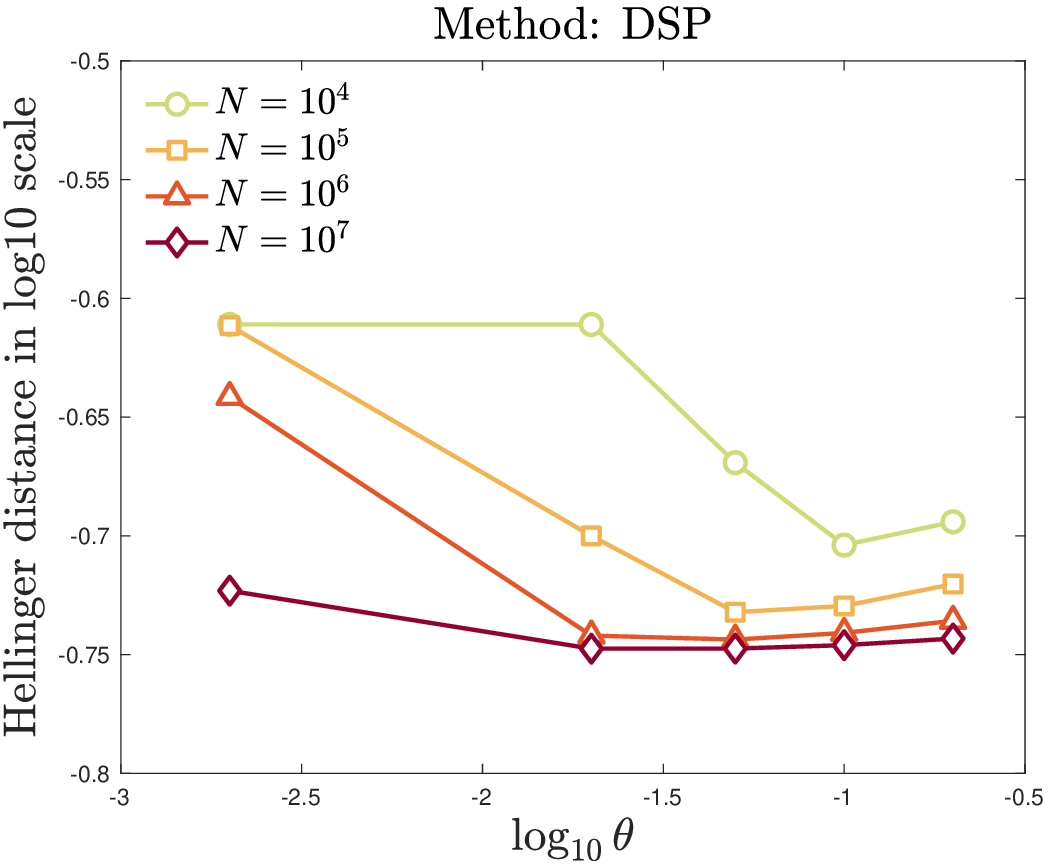}}
{\includegraphics[width=0.32\textwidth,height=0.24\textwidth]{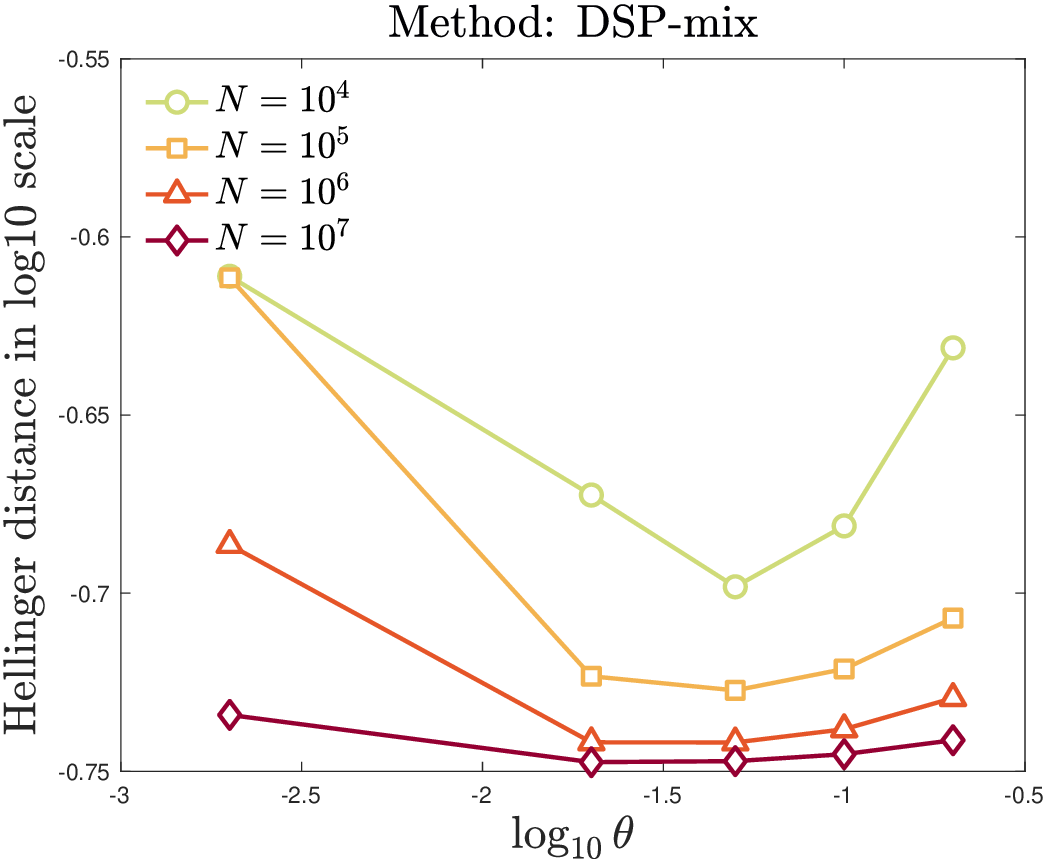}}
{\includegraphics[width=0.32\textwidth,height=0.24\textwidth]{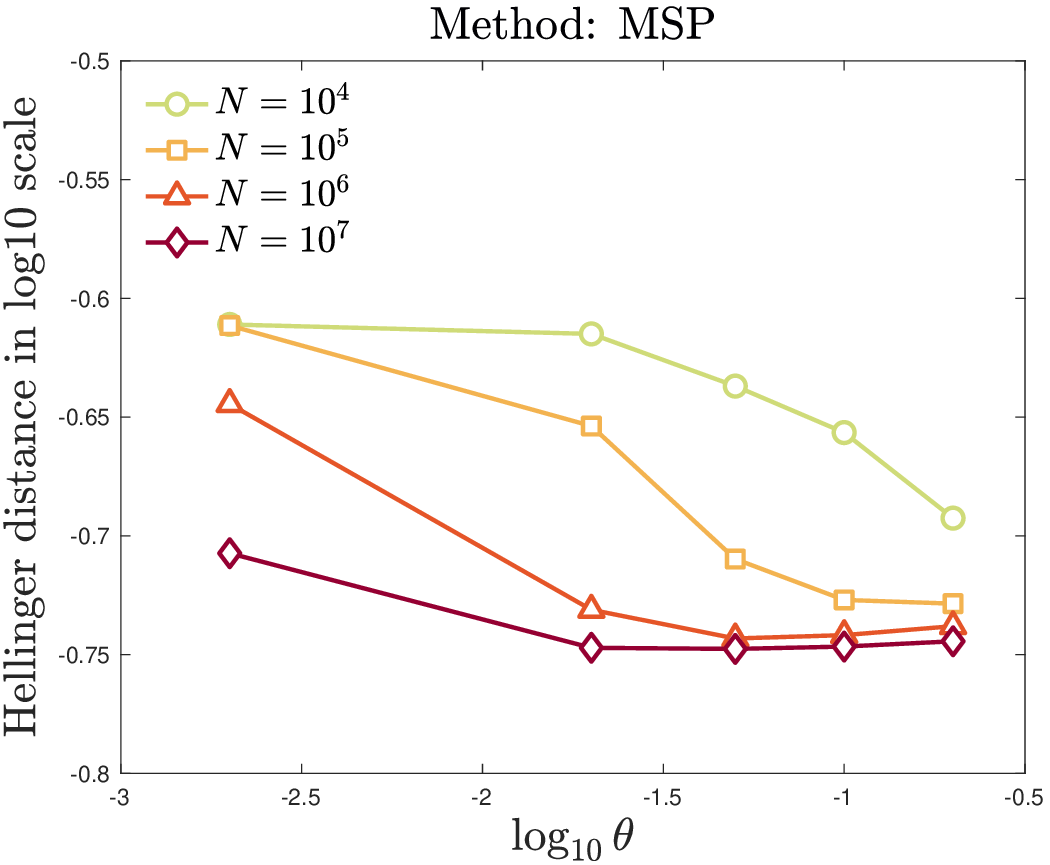}}}
\caption{\small $2$-D Cauchy mixtures: The KL divergence and Hellinger distance under different $N$ and $\theta$.}
\label{cauchy_low}
\end{figure}

\begin{figure}[!h]
\centering
\subfigure[True density and samples.]{
\includegraphics[width=0.4\textwidth,height=0.3\textwidth]{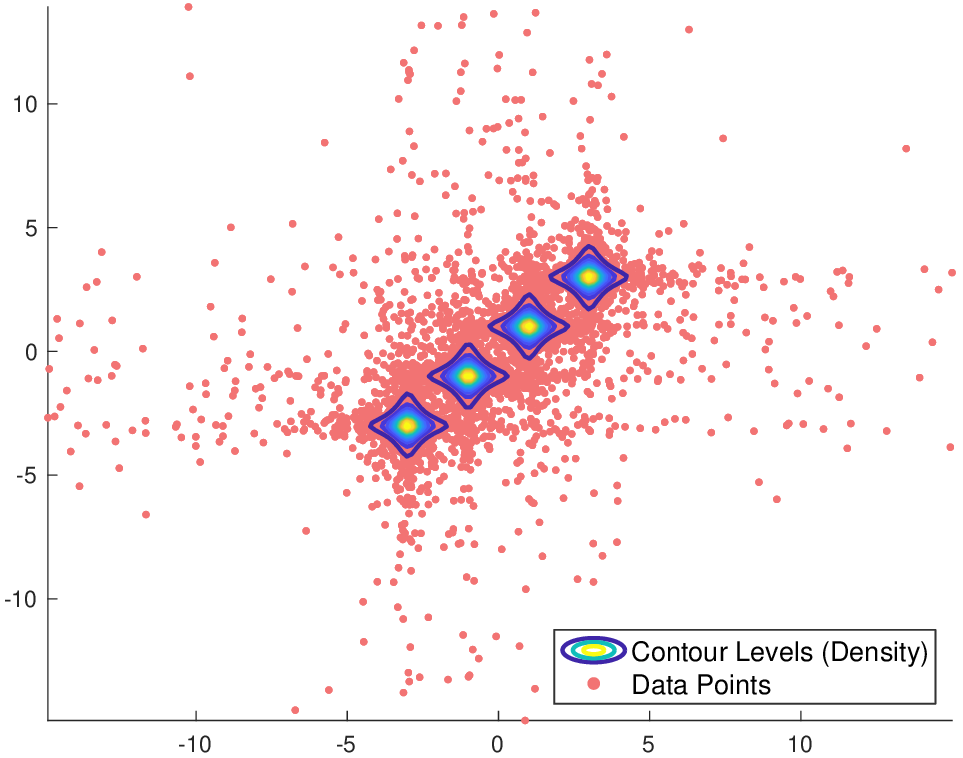}}
\subfigure[Heavy tail of Cauchy mixtures.]{
\includegraphics[width=0.4\textwidth,height=0.3\textwidth]{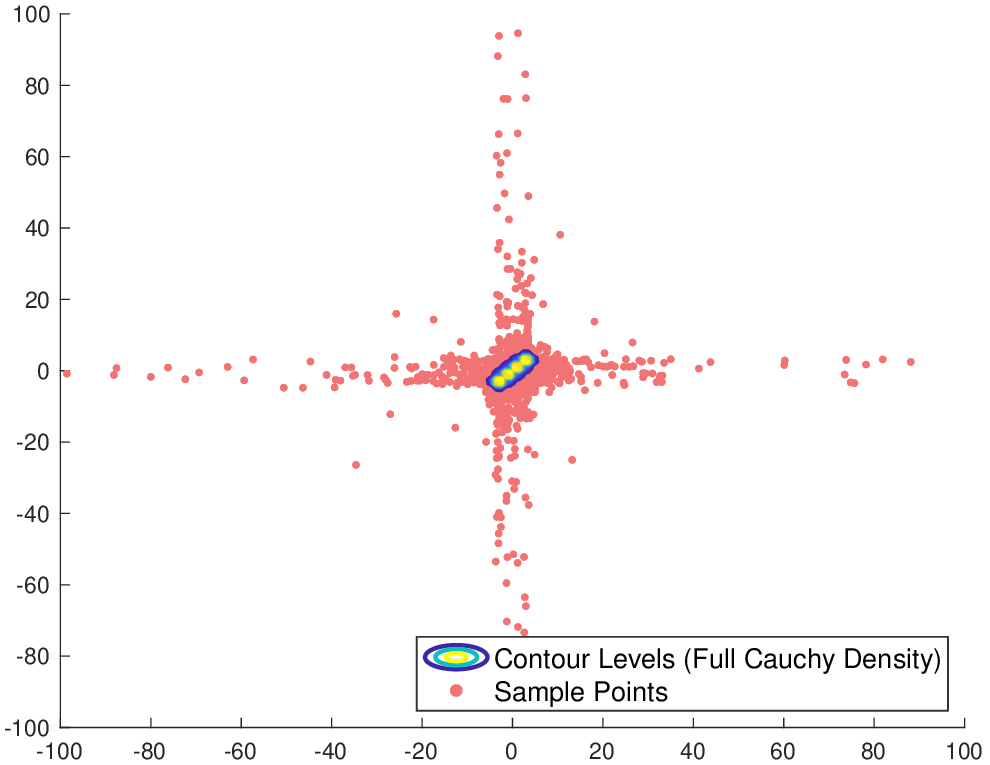}}
\\
\centering
\subfigure[Adaptive partition (left: DSP, middle: DSP-mix, right: MSP).]{
\includegraphics[width=0.32\textwidth,height=0.24\textwidth]{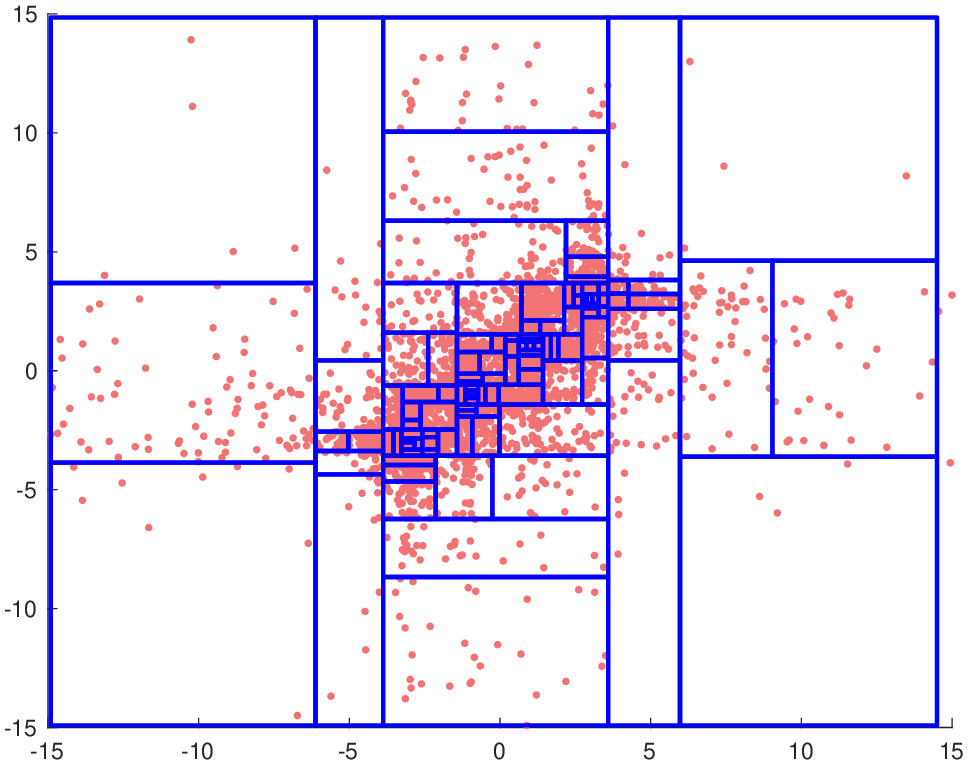}
\includegraphics[width=0.32\textwidth,height=0.24\textwidth]{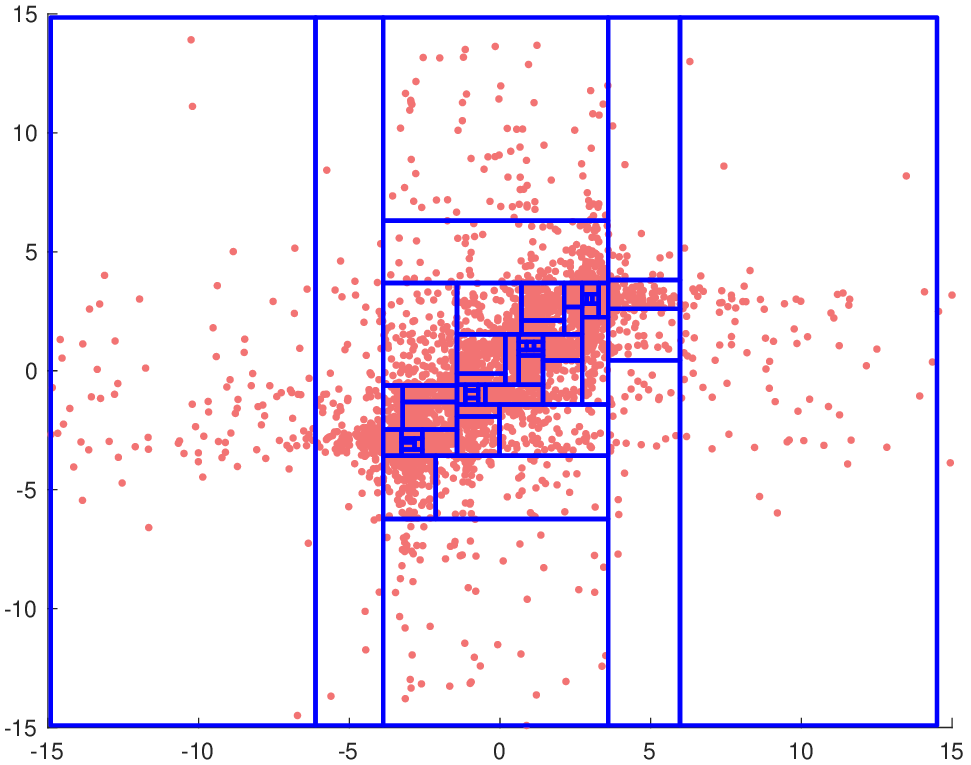}
\includegraphics[width=0.32\textwidth,height=0.24\textwidth]{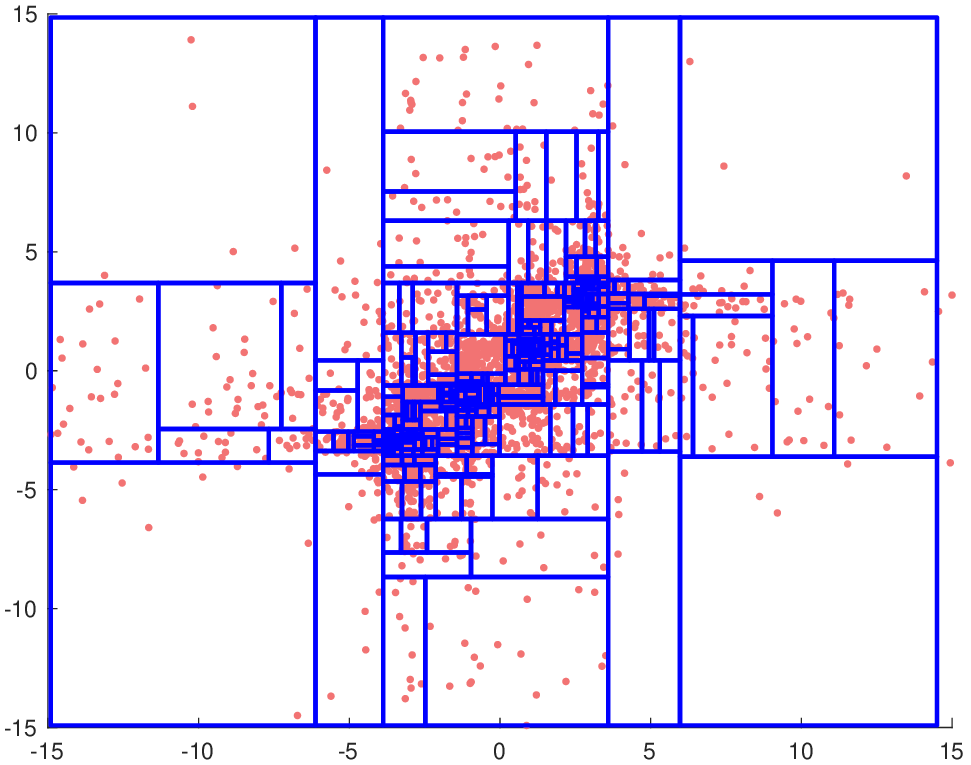}}
\\
\centering
\subfigure[Density estimation (left: DSP, middle: DSP-mix, right: MSP).]{
\includegraphics[width=0.32\textwidth,height=0.24\textwidth]{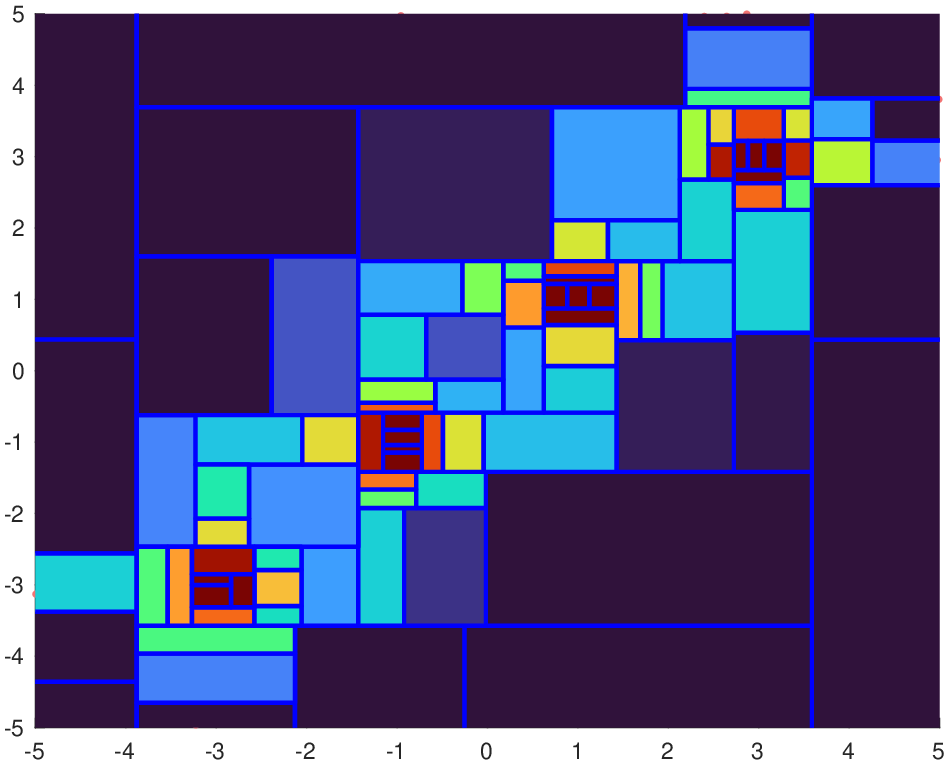}
\includegraphics[width=0.32\textwidth,height=0.24\textwidth]{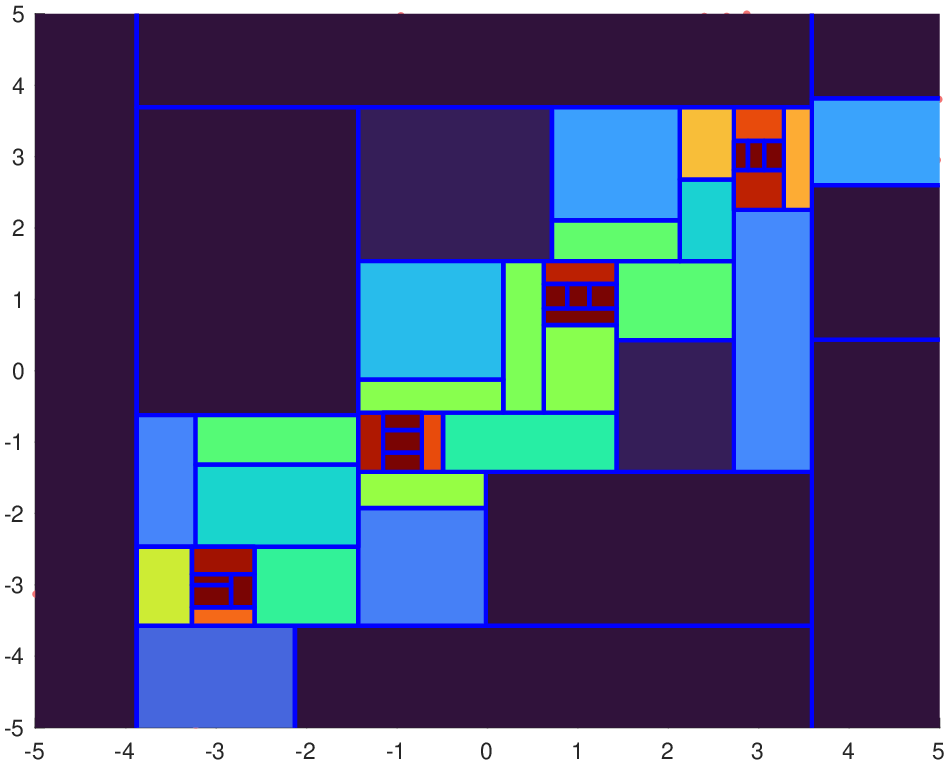}
\includegraphics[width=0.32\textwidth,height=0.24\textwidth]{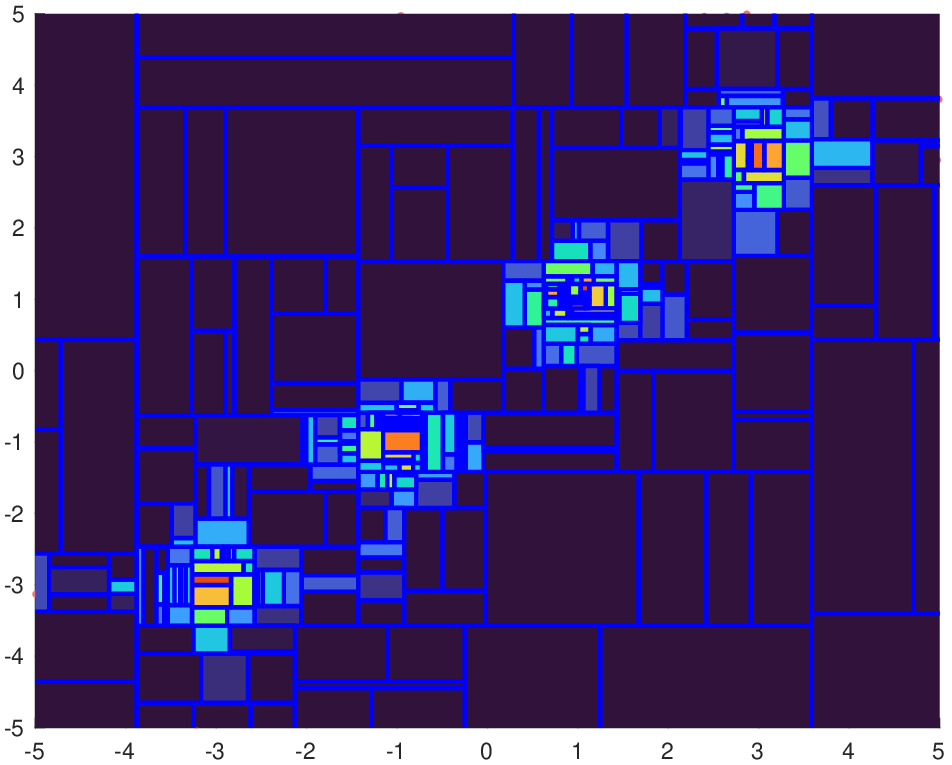}}
	\caption{2-D Cauchy mixtures: Adaptive partitions and density estimators produced by DSP-mix, MSP and DSP with $N = 1\times 10^4$ and $\theta=0.2$.}
	\label{2dcauchymix fig}
\end{figure}

One can sample the Cauchy distribution by using its cumulative function 
\begin{equation}
F\left(\frac{x}{u_j}\right) = \frac{1}{\pi} \left[\arctan\left(\frac{x}{u_j} - \frac{m_j}{u_j}\right) + \frac{\pi}{2}\right].
\end{equation}
As shown in Figure \ref{2dcauchymix fig}, the Cauchy distribution has a very heavy tail, although the typical set lies in a bounded domain $[-15, 15]^d$. In fact, the Cauchy distribution has infinite mean and variance, so that direct Monte Carlo sampling may fail to converge. Therefore, we only try to reconstruct the density in the typical set $\Omega = [-15, 15]^d$ and discard the particles outside $\Omega$.

Indeed, the heavy tail poses a significant challenge for the tree-based density estimation. Nevertheless,  DSP, DSP-mix and MSP are able to capture the four-peak structure in the typical set, as clearly visualized in Figure \ref{2dcauchymix fig}.  Figure \ref{cauchy_low} shows that the behavior of the KL divergence is a bit strange compared with Figures~\ref{beta_low} and \ref{gauss_low} for 2-D Beta mixtures and 2-D Gaussian mixtures.
It may be also related to the heavy tail and we guess that, when the domain is enlarged, the performance of tree-based density estimator should be more sensitive to the partition level. As $\theta$ increases, the decision tree stops at a  shallow level, producing coarser but more stable regions. This coarsening effectively suppresses the influence of isolated Cauchy outliers, causing the KL divergence to be smaller. That is, the KL divergence no longer serves as a satisfactory criterion for such heavy tail distribution as also pointed out in \cite{WuLindquist2024}. By contrast, the Hellinger distance always displays a similar behavior for the three mixtures. 

\begin{figure}[H]
\centering
\subfigure[KL divergence (left: DSP, middle: DSP-mix, right: MSP). ]{
{\includegraphics[width=0.32\textwidth,height=0.24\textwidth]{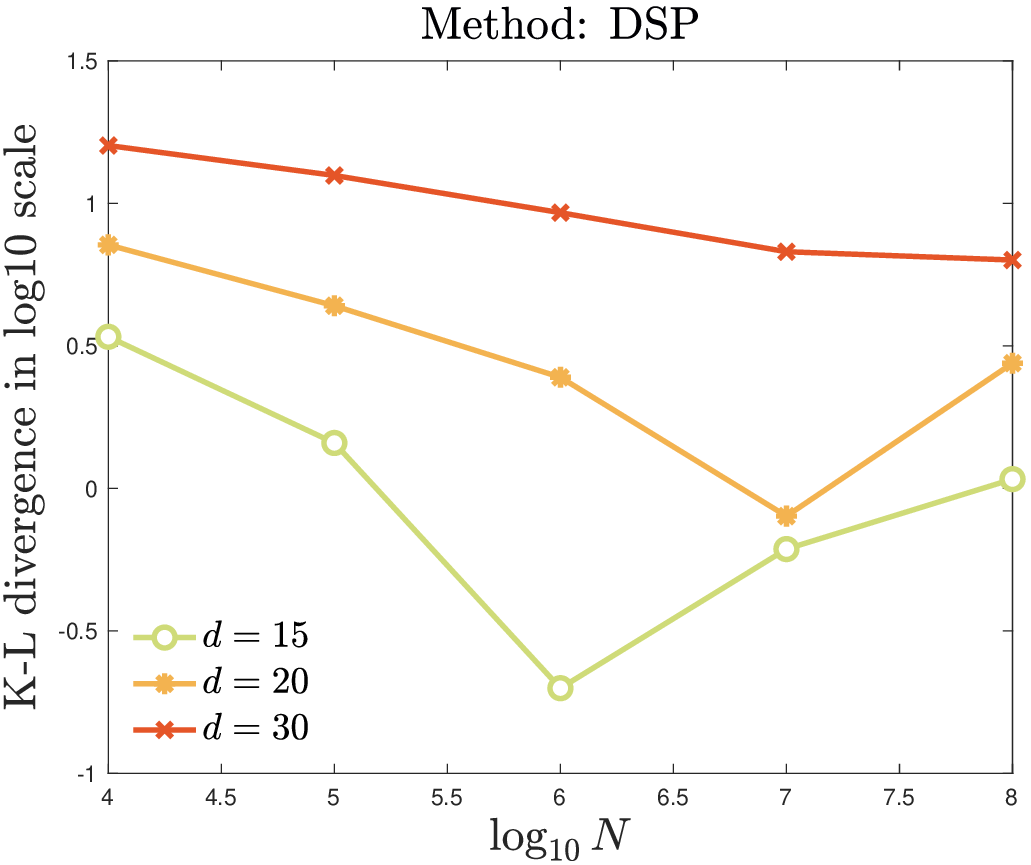}}
{\includegraphics[width=0.32\textwidth,height=0.24\textwidth]{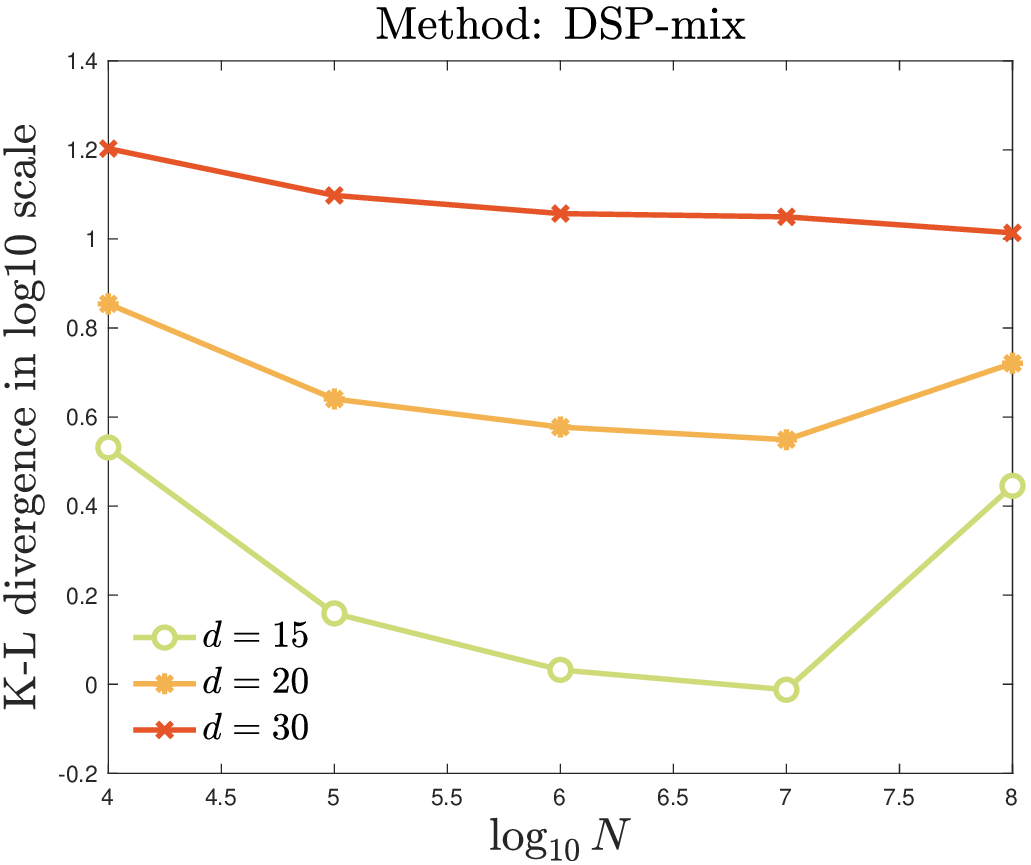}}
{\includegraphics[width=0.32\textwidth,height=0.24\textwidth]{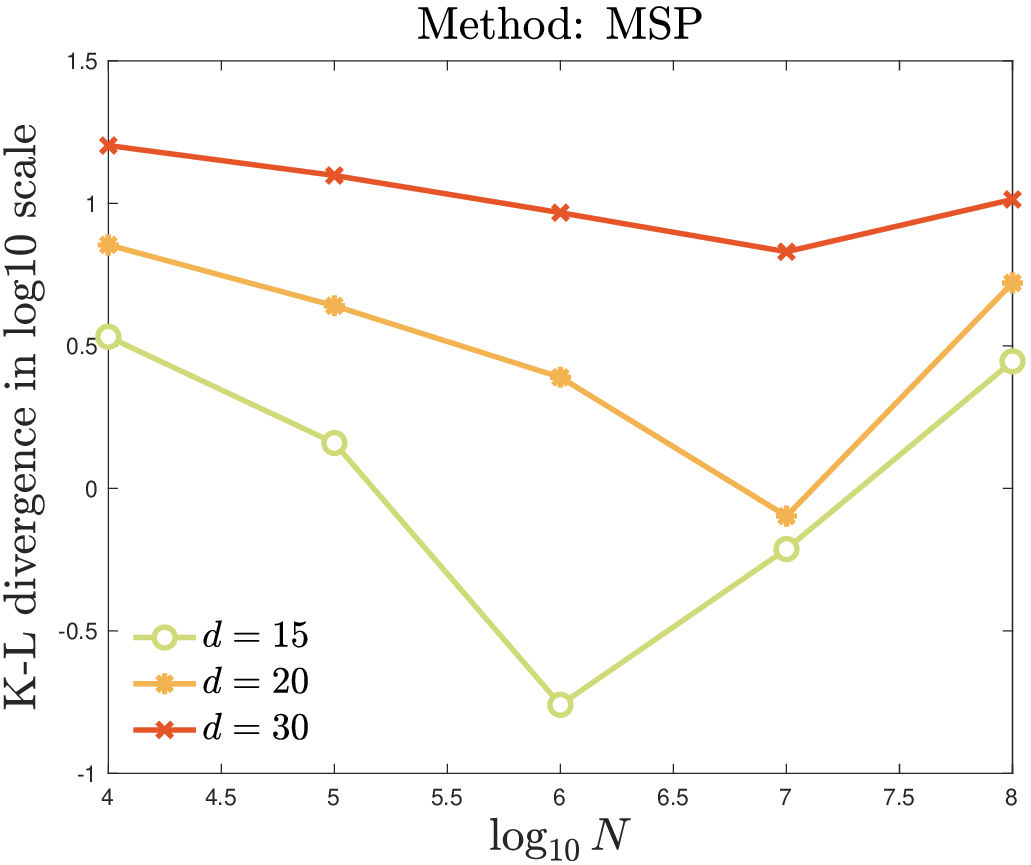}}}
\\
\centering
\subfigure[Hellinger distance (left: DSP, middle: DSP-mix, right: MSP). ]{
{\includegraphics[width=0.32\textwidth,height=0.24\textwidth]{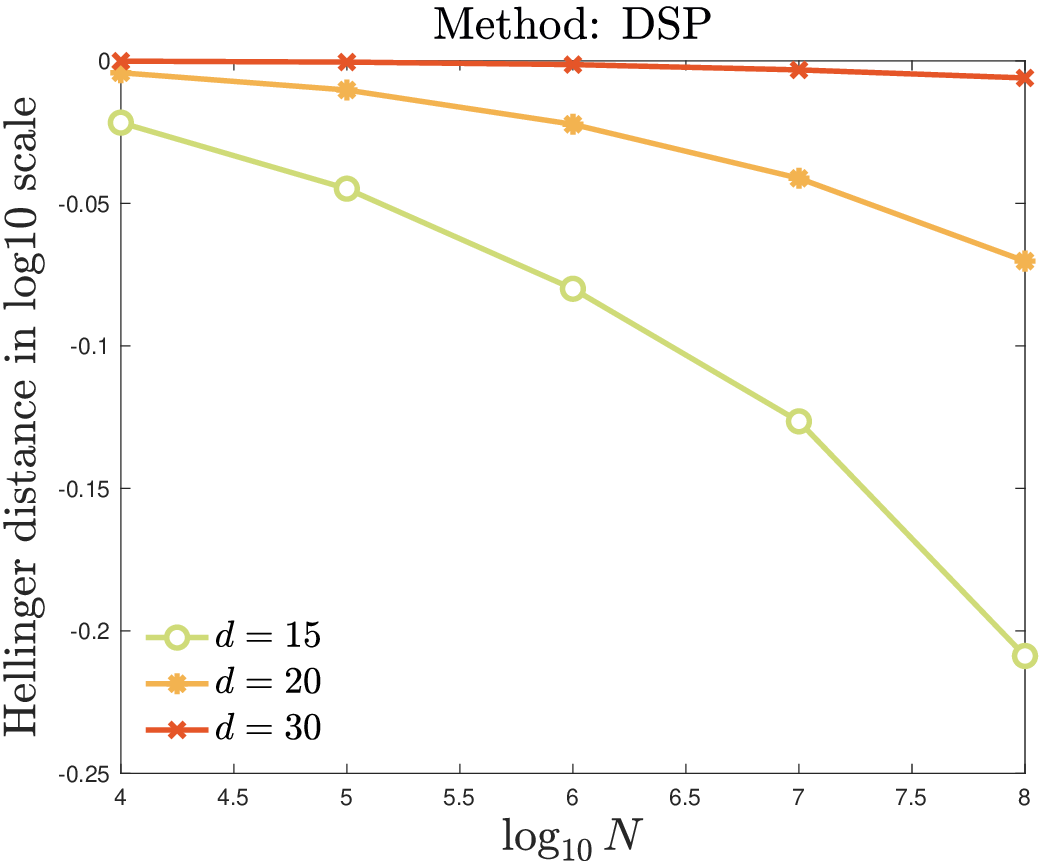}}
{\includegraphics[width=0.32\textwidth,height=0.24\textwidth]{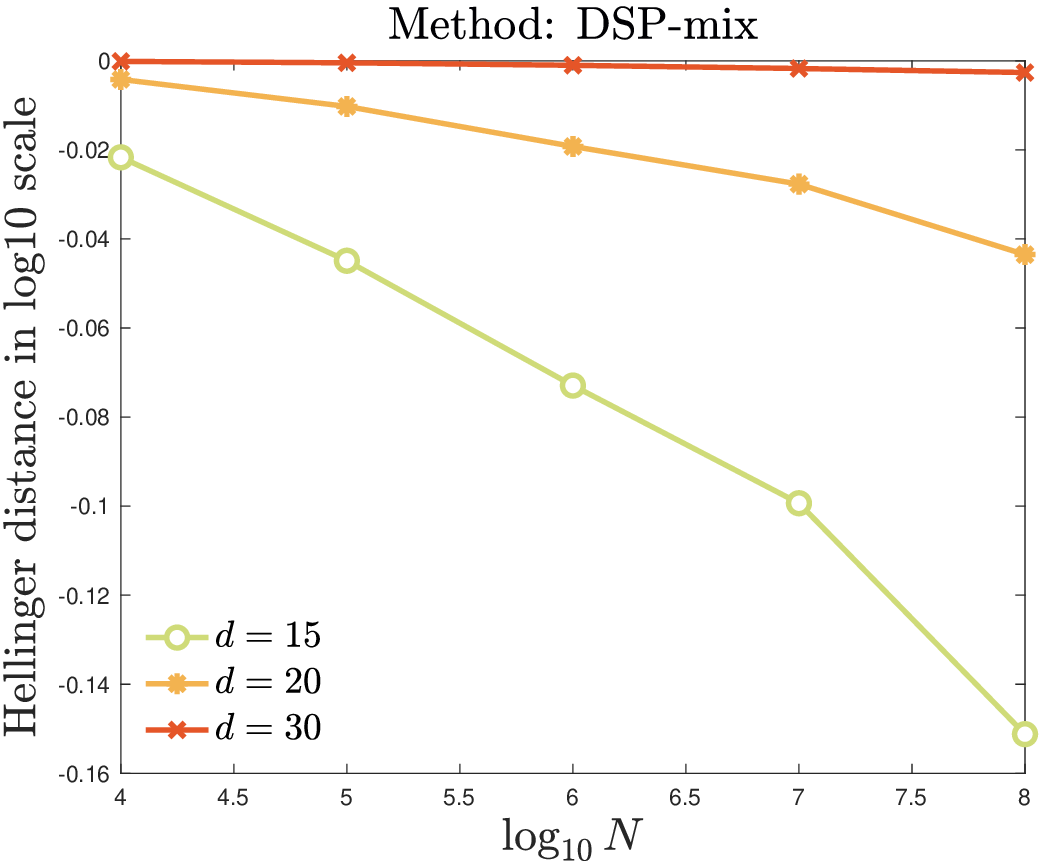}}
{\includegraphics[width=0.32\textwidth,height=0.24\textwidth]{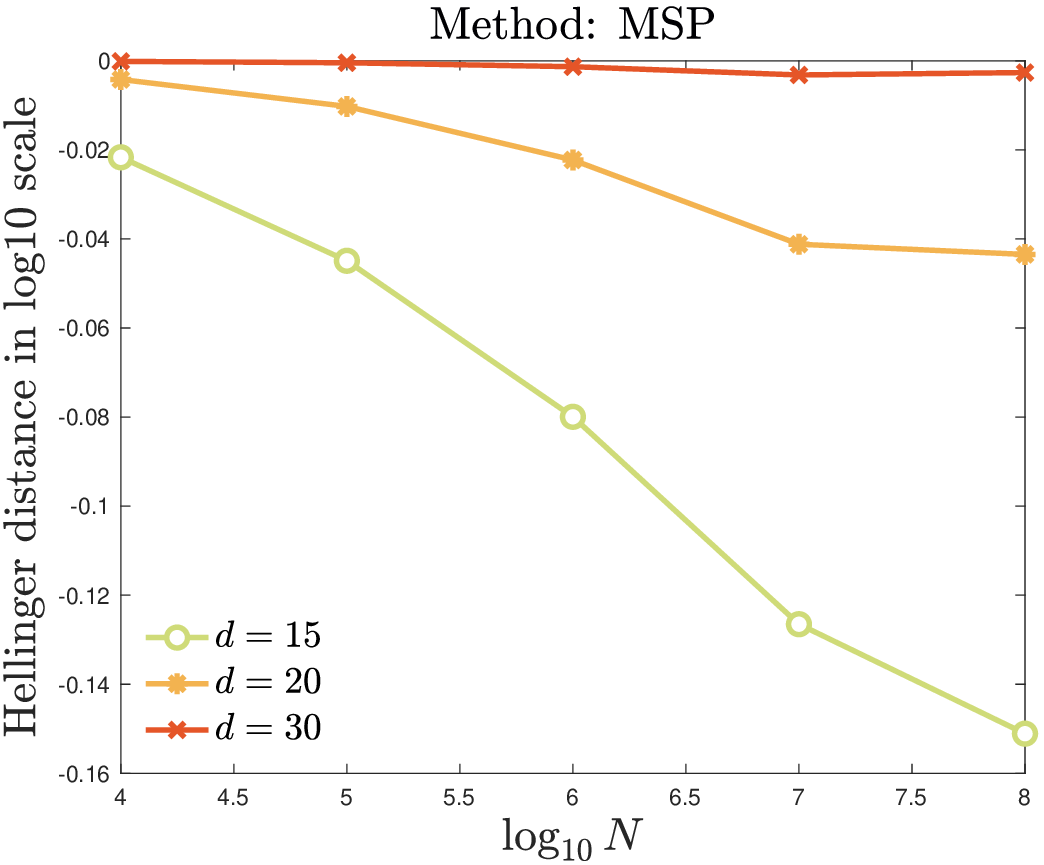}}}
\\
\centering
\subfigure[Partition level (left: DSP, middle: DSP-mix, right: MSP). ]{
{\includegraphics[width=0.32\textwidth,height=0.24\textwidth]{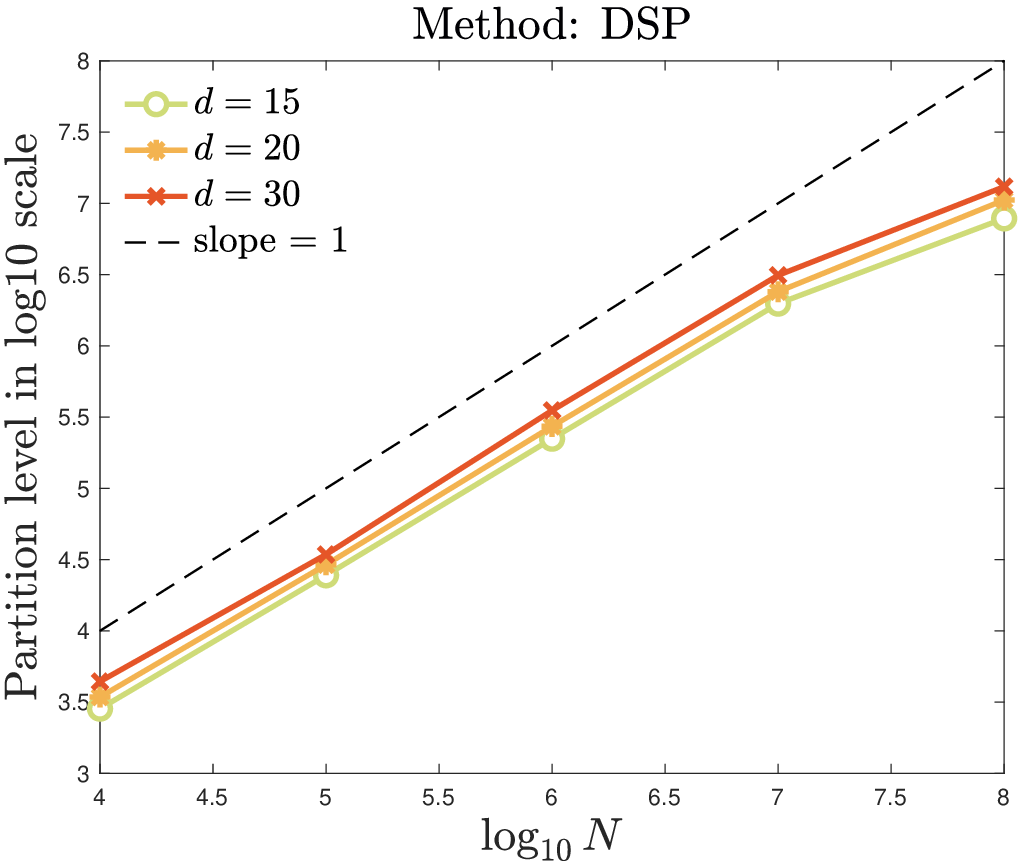}}
{\includegraphics[width=0.32\textwidth,height=0.24\textwidth]{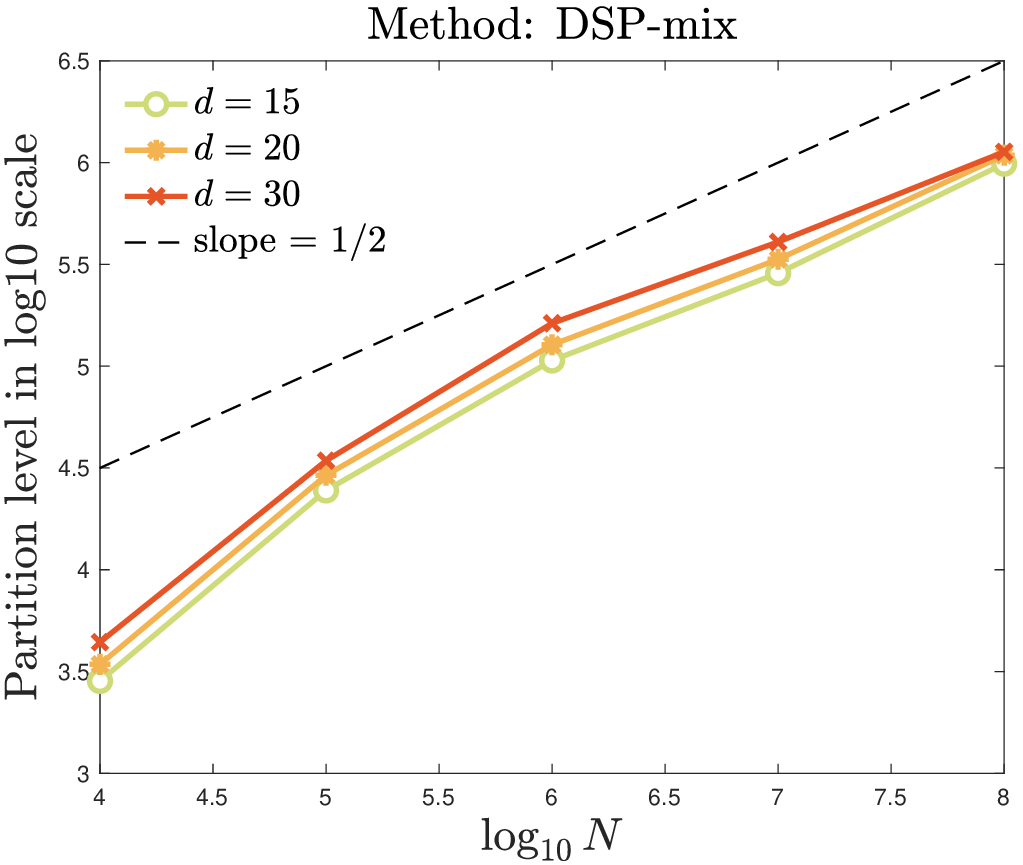}}
{\includegraphics[width=0.32\textwidth,height=0.24\textwidth]{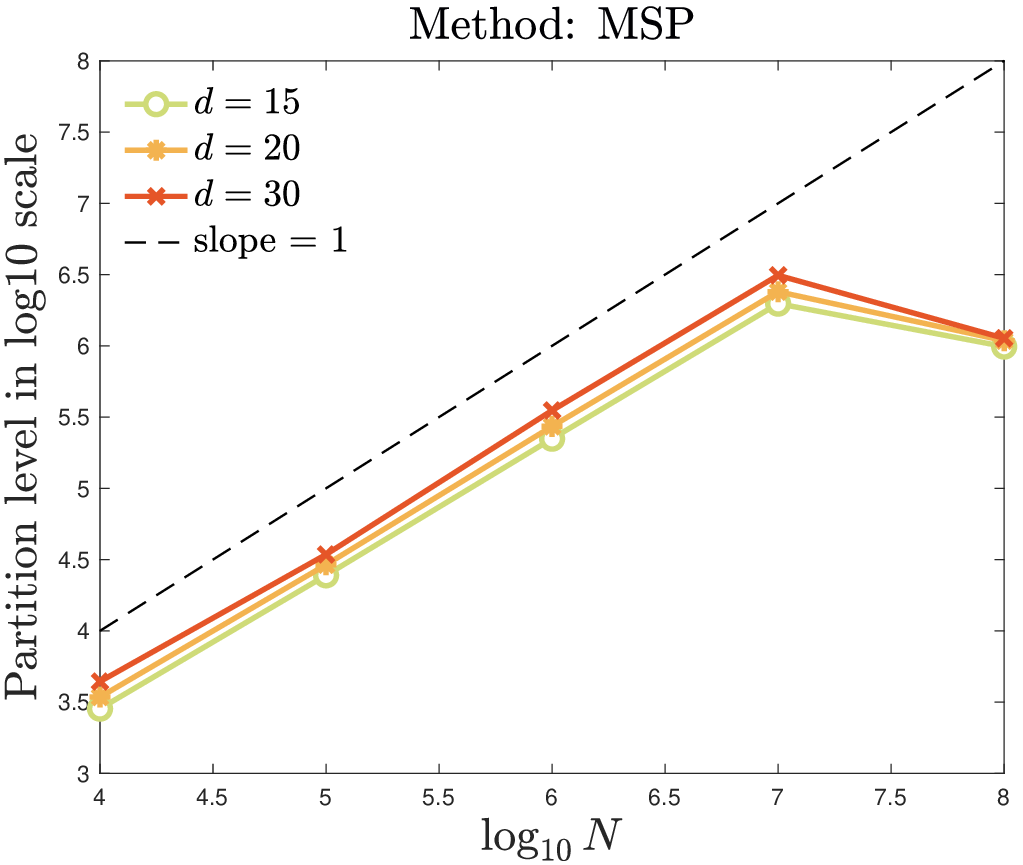}}}
\caption{\small $d$-D Cauchy mixtures: Convergence of DSP, DSP-mix and MSP with respect to $N$
for $d=15, 20, 30$.}  \label{hdcauchy_high_dimensional}
\end{figure}

\begin{figure}[H]
\centering
\subfigure[Partition level ($N = 10^4$) (left: DSP, middle: DSP-mix, right: MSP). ]{
{\includegraphics[width=0.32\textwidth,height=0.24\textwidth]{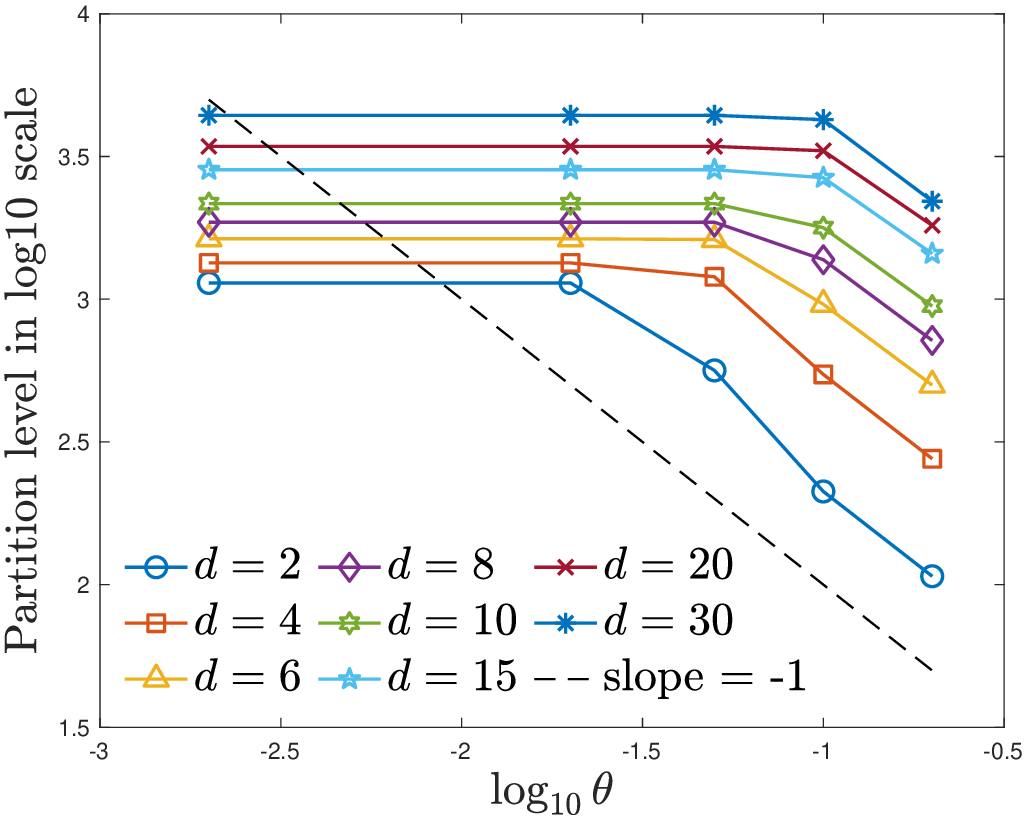}}
{\includegraphics[width=0.32\textwidth,height=0.24\textwidth]{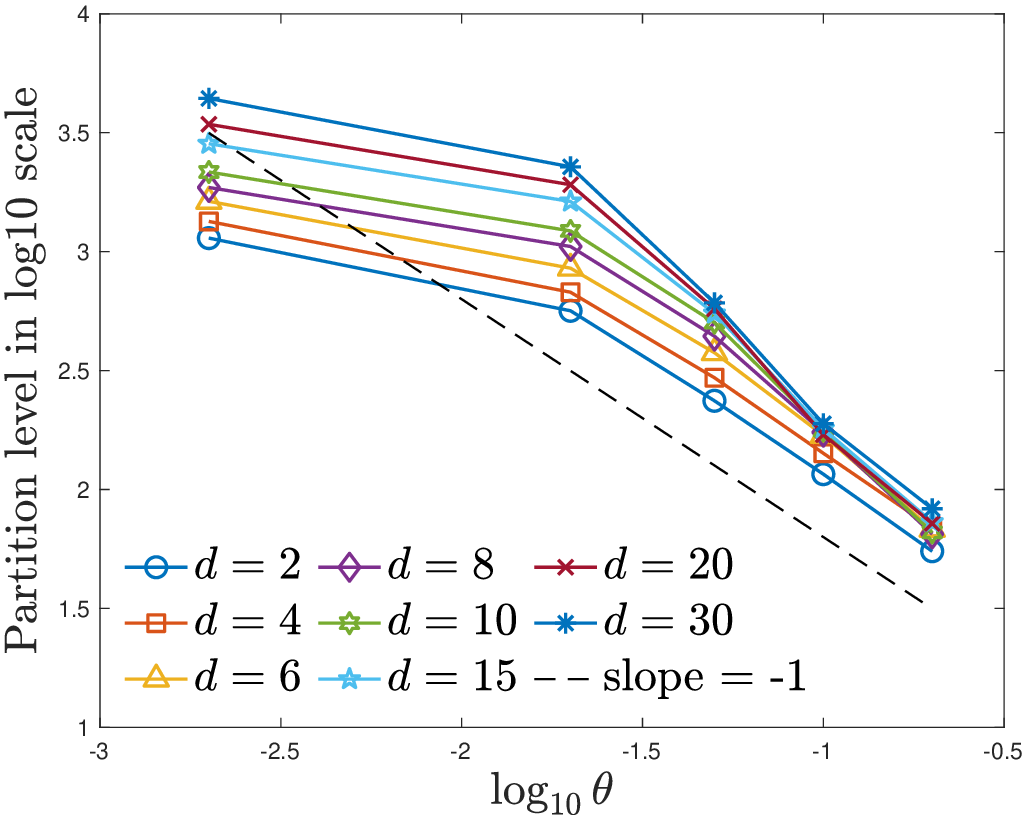}}
{\includegraphics[width=0.32\textwidth,height=0.24\textwidth]{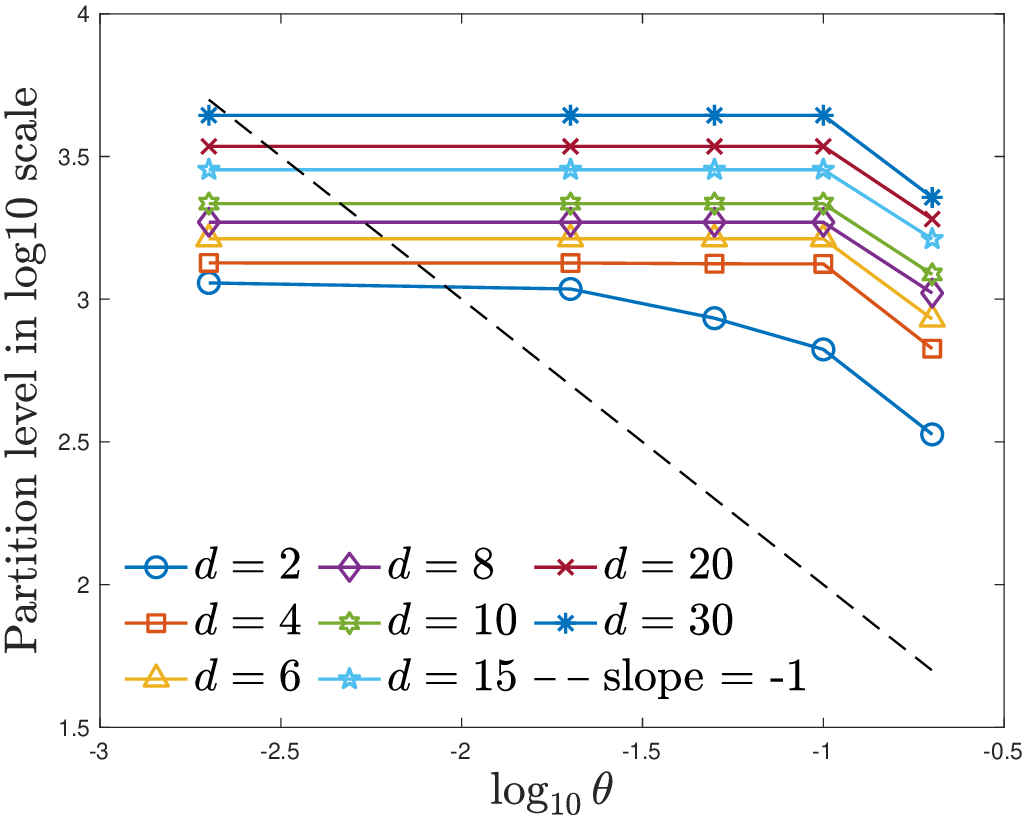}}}
\\
\centering
\subfigure[Partition level ($N = 10^5$) (left: DSP, middle: DSP-mix, right: MSP). ]{
{\includegraphics[width=0.32\textwidth,height=0.24\textwidth]{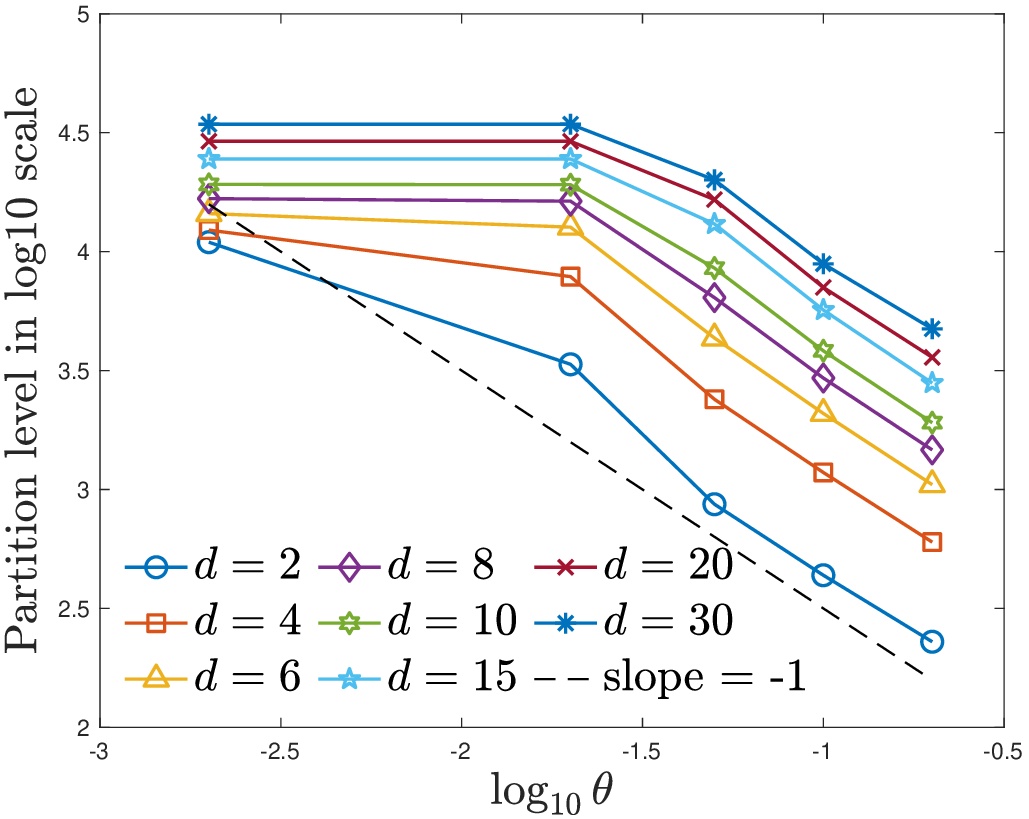}}
{\includegraphics[width=0.32\textwidth,height=0.24\textwidth]{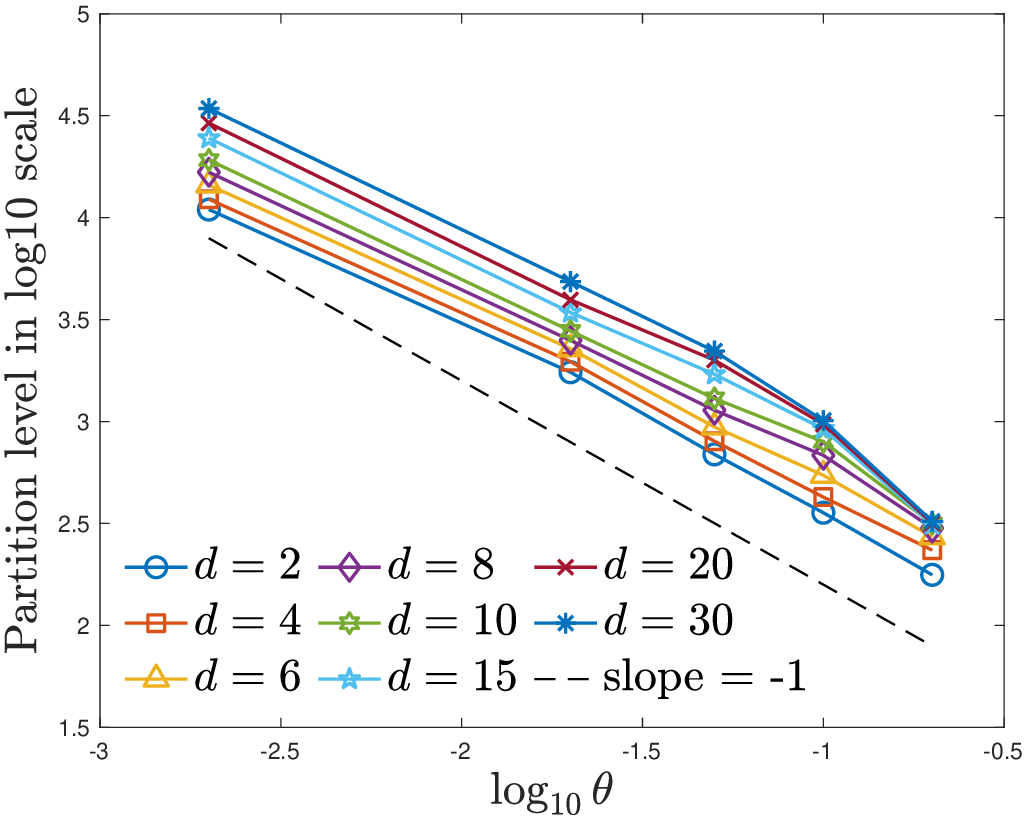}}
{\includegraphics[width=0.32\textwidth,height=0.24\textwidth]{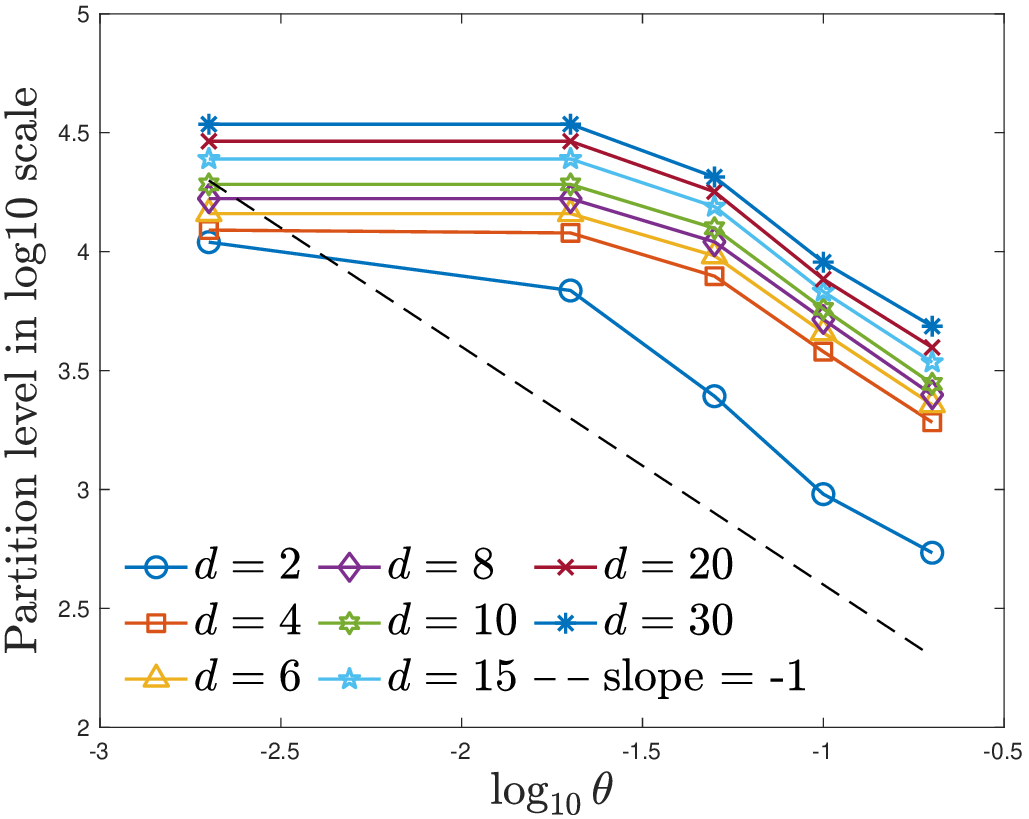}}}
\\
\centering
\subfigure[Partition level ($N = 10^6$) (left: DSP, middle: DSP-mix, right: MSP). ]{
{\includegraphics[width=0.32\textwidth,height=0.24\textwidth]{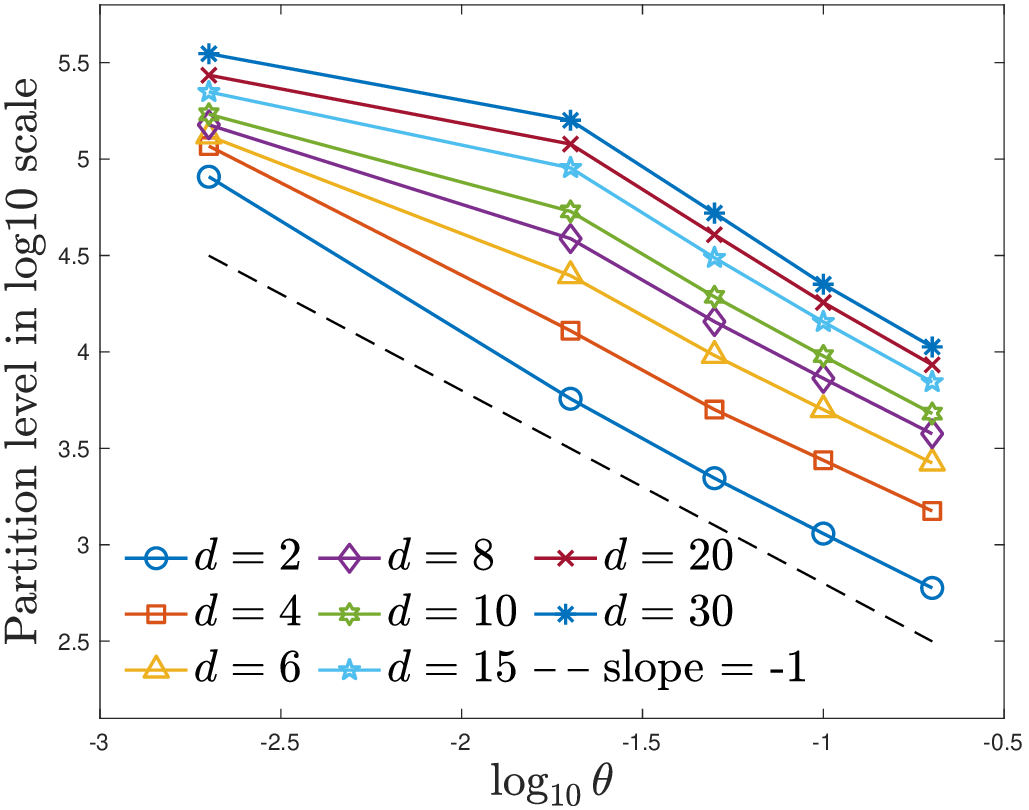}}
{\includegraphics[width=0.32\textwidth,height=0.24\textwidth]{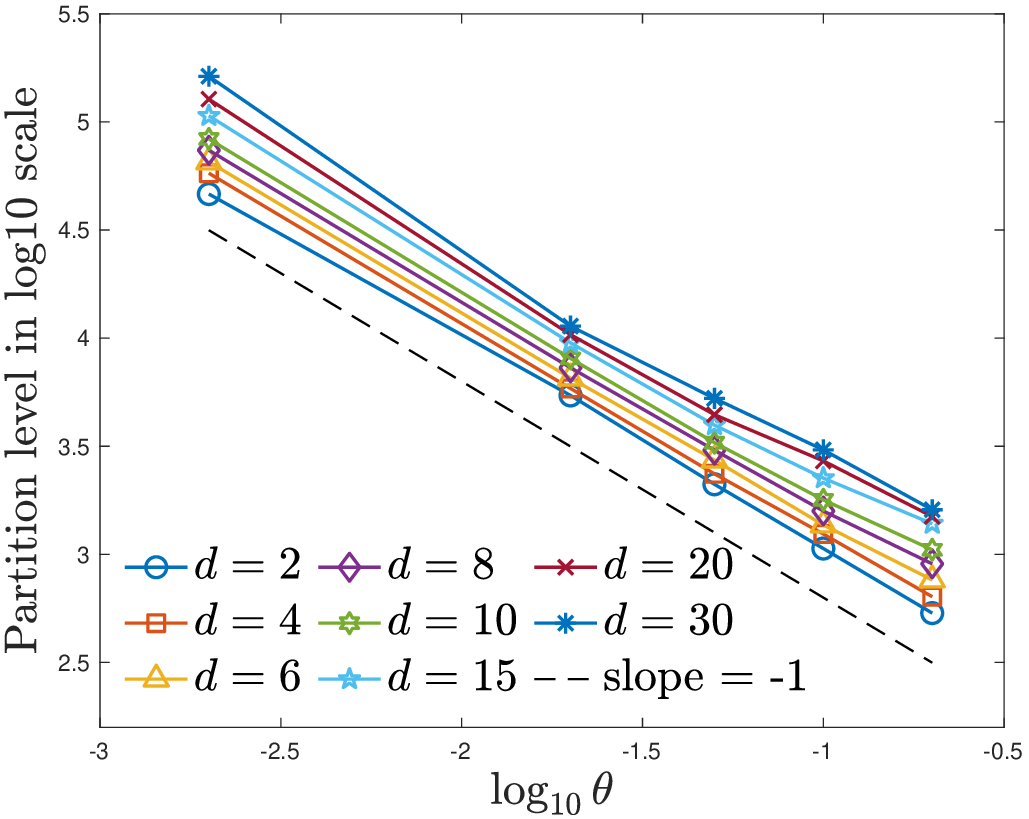}}
{\includegraphics[width=0.32\textwidth,height=0.24\textwidth]{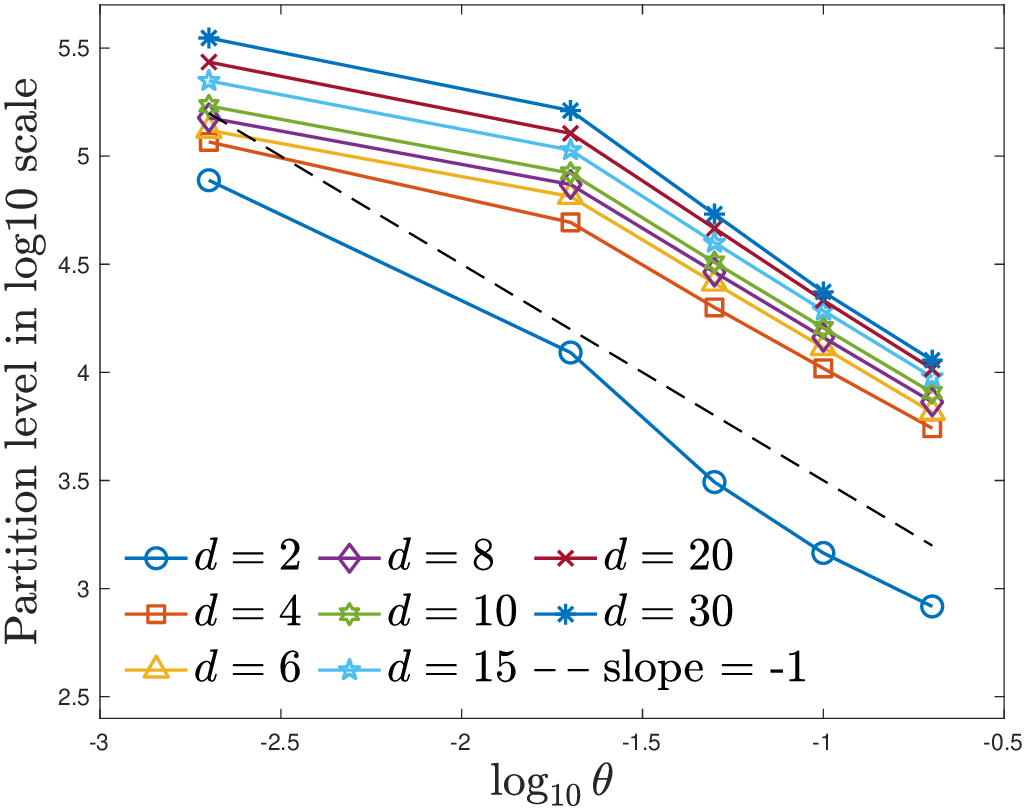}}}
\\
\centering
\subfigure[Partition level ($N = 10^7$) (left: DSP, middle: DSP-mix, right: MSP). ]{
{\includegraphics[width=0.32\textwidth,height=0.24\textwidth]{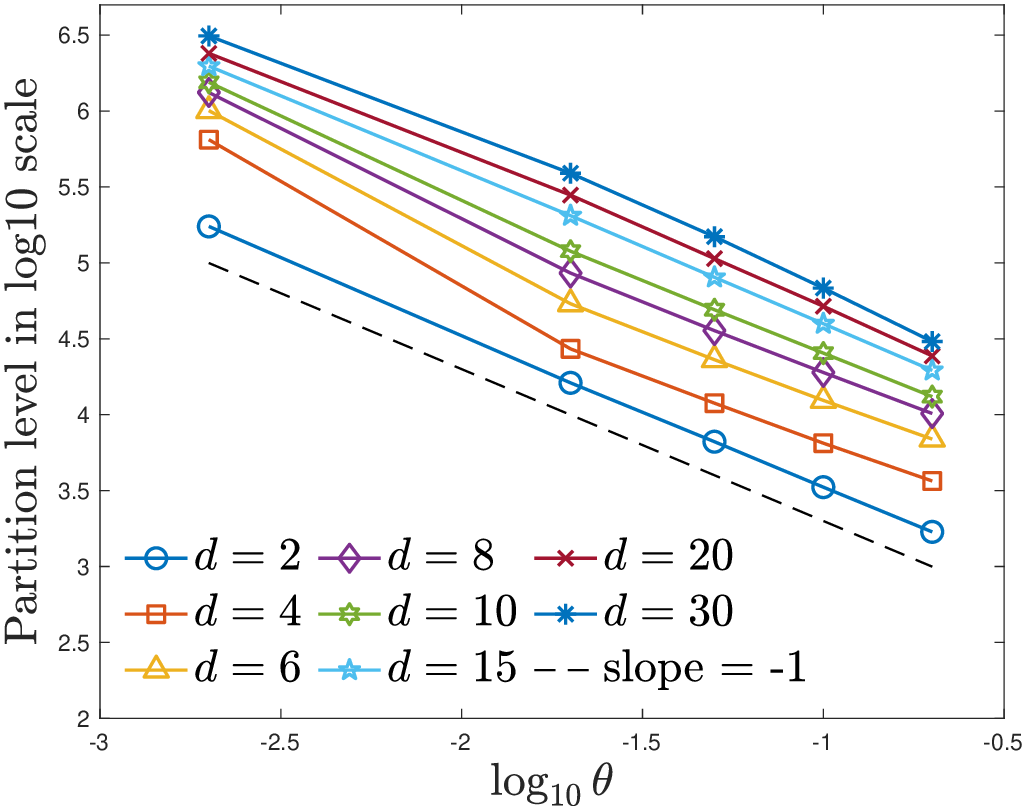}}
{\includegraphics[width=0.32\textwidth,height=0.24\textwidth]{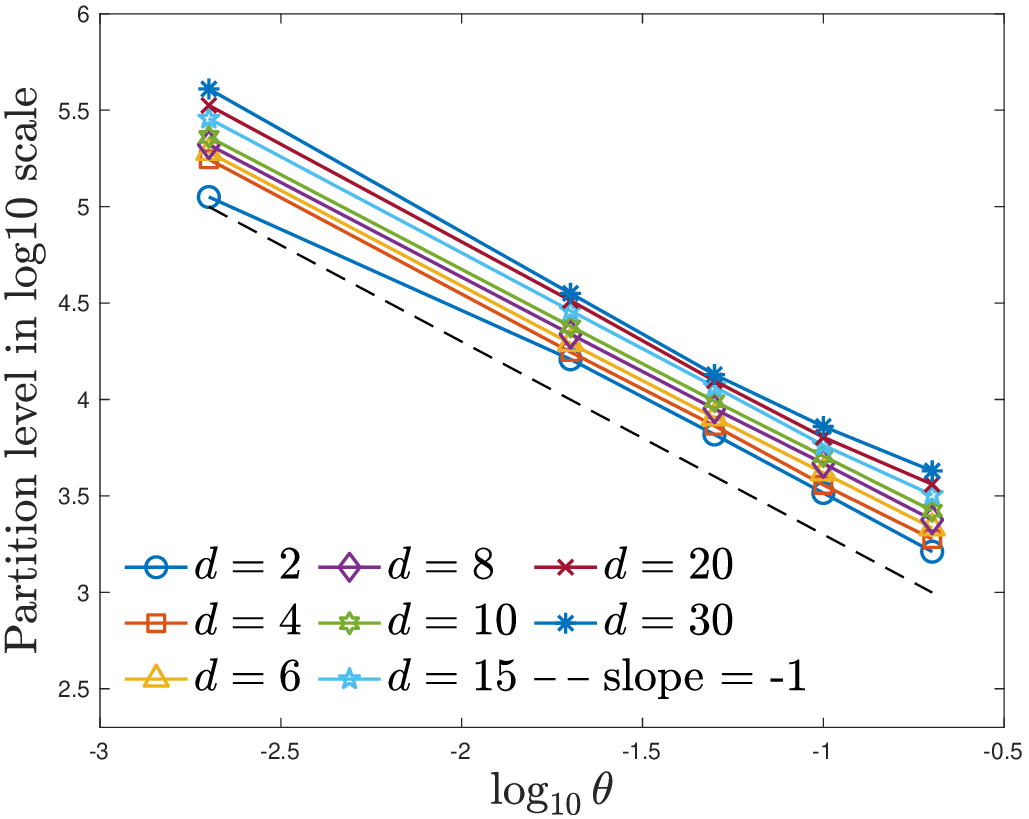}}
{\includegraphics[width=0.32\textwidth,height=0.24\textwidth]{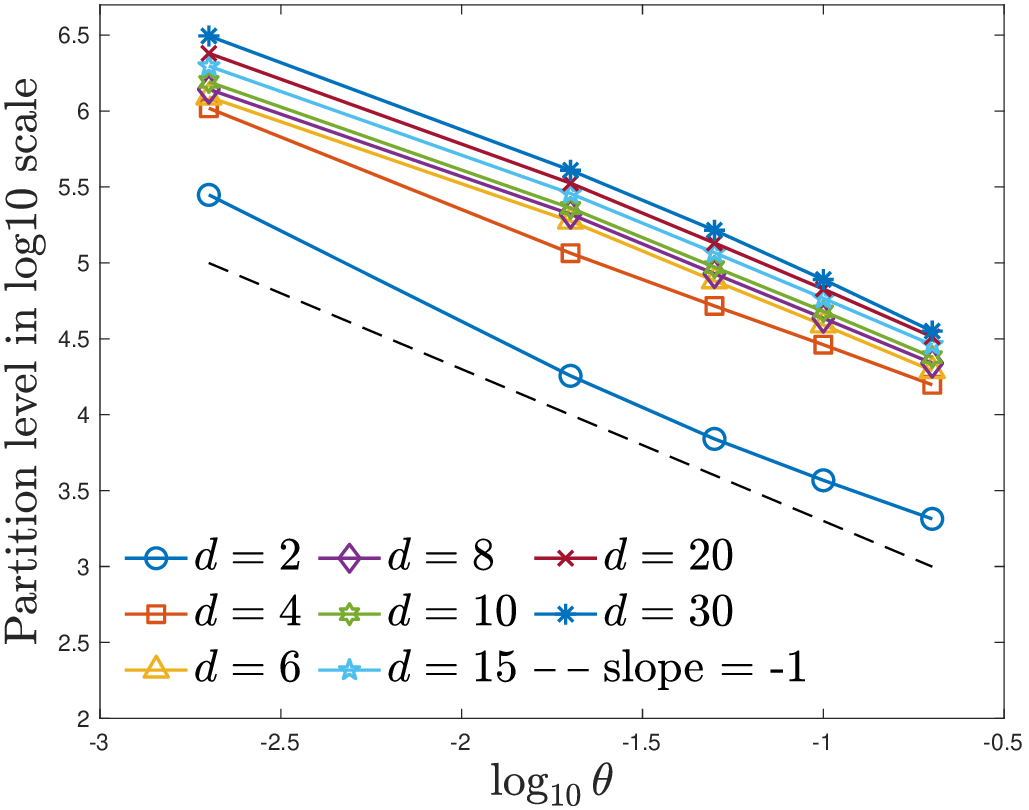}}}
\caption{\small $d$-D Cauchy mixtures: The partition level $L$ under different $\theta$, $N$ and $d$. \label{hdcauchy_partition}}
\end{figure}

For $d = 15, 20, 30$,  Figure \ref{hdcauchy_high_dimensional} presents the convergence of error metrics with respect to sample size $N = 10^4, 10^5, 10^6, 10^7, 10^8$. The raw data is collected in Tables \ref{hdCauchy_error_N1e7} and  \ref{hdGauss_high}. Figure \ref{hdcauchy_partition} plots the partition level $L$ under different $d$, $N$, with $\theta$ fixed to be $0.002$. The raw data is collected in Tables \ref{hdCauchy_K_time_N1e7} and \ref{hdCauchy_K_time_high}.
\begin{enumerate}

\item[(1)] Overall, the errors diminish as either sample size $N$ and the partition level $L$ increase. The accuracy of MSP is comparable to DSP, whereas DSP-mix is less accurate as dimension $d$ increases.

\item[(2)] Notably, for higher-dimensional cases with larger-sample, e.g., $d = 30, N = 10^7$, the KL divergence behaves smoothly across all $\theta$, confirming that the estimator can work well 
provided sufficient sample coverage of the heavy-tailed support.

\item[(3)] The trend $L \sim \theta^{-1}$ is still observed in Figure \ref{hdcauchy_partition}. This gives us a simple way to estimate the partition level by first testing on a small dataset and using extrapolation, which also helps control the computational complexity.

\end{enumerate}

\begin{table}[H]
	\centering
	\caption{$d$-D Cauchy mixtures: The KL divergence and Hellinger distance  under different $d$ and $\theta$. The sample size is fixed to be $N = 10^7$.}
	\setlength{\tabcolsep}{4pt}
	\begin{tabular}{c|c|cc|cc|cc|cc|cc}
		\toprule
		& $\theta$	& \multicolumn{2}{c}{$0.002$} & \multicolumn{2}{c}{$0.02$} & \multicolumn{2}{c}{$0.05$} &  \multicolumn{2}{c}{$0.1$} &  \multicolumn{2}{c}{$0.2$}\\ 
		\midrule
		 $d$	&	Method	& KL & $\hat{H}^2$ &  KL & $\hat{H}^2$  & KL & $\hat{H}^2$  & KL & $\hat{H}^2$  & KL & $\hat{H}^2$  \\
		 \midrule
		 \multirow{3}{*}{2}	&	DSP	&	0.4771 	&	0.1892 	&	0.3830 	&	0.1789 	&	0.3836 	&	0.1789 	&	0.3765 	&	0.1795 	&	0.3425 	&	0.1806 	\\
&	DSP-mix	&	0.4229 	&	0.1844 	&	0.3832 	&	0.1789 	&	0.3841 	&	0.1790 	&	0.3777 	&	0.1798 	&	0.3454 	&	0.1814 	\\
&	MSP	&	0.4414 	&	0.1962 	&	0.3825 	&	0.1790 	&	0.3832 	&	0.1788 	&	0.3755 	&	0.1792 	&	0.3407 	&	0.1801 	\\
						& (maxSD) &      	0.1507 	&	0.0002 	&	0.0325 	&	0.0001 	&	0.0190 	&	0.0001 	&	0.0121 	&	0.0001 	&	0.0182 	&	0.0001 		      \\							
		 \midrule
		 \multirow{3}{*}{4}	&	DSP	&	0.1406 	&	0.2198 	&	0.1003 	&	0.1922 	&	0.1190 	&	0.2019 	&	0.0520 	&	0.2128 	&	0.1000 	&	0.2255 	\\
&	DSP-mix	&	0.1276 	&	0.1888 	&	0.1334 	&	0.2015 	&	0.0730 	&	0.2164 	&	0.1408 	&	0.2350 	&	0.0086 	&	0.2573 	\\
&	MSP	&	0.1231 	&	0.2460 	&	0.0724 	&	0.1866 	&	0.1129 	&	0.1887 	&	0.1199 	&	0.1936 	&	0.1312 	&	0.2018 	\\
						& (maxSD) &        	0.2531 	&	0.0001 	&	0.0735 	&	0.0001 	&	0.0539 	&	0.0001 	&	0.0383 	&	0.0001 	&	0.0864 	&	0.0003 		     \\	
		 \midrule
		 \multirow{3}{*}{6}	&	DSP	&	-0.1851 	&	0.2715 	&	-0.0588 	&	0.2648 	&	-0.1155 	&	0.2925 	&	-0.0748 	&	0.3189 	&	-0.0257 	&	0.3481 	\\
&	DSP-mix	&	-0.0882 	&	0.2456 	&	0.0001 	&	0.3083 	&	0.0902 	&	0.3492 	&	0.1615 	&	0.3862 	&	0.2793 	&	0.4290 	\\
&	MSP	&	-0.2477 	&	0.2841 	&	-0.1375 	&	0.2455 	&	-0.1319 	&	0.2619 	&	-0.0682 	&	0.2813 	&	-0.0062 	&	0.3083 	\\
						& (maxSD) &       	0.1513 	&	0.0001 	&	0.1260 	&	0.0003 	&	0.1220 	&	0.0004 	&	0.1261 	&	0.0005 	&	0.0551 	&	0.0007 		    \\		
		 \midrule
		 \multirow{3}{*}{8}	&	DSP	&	-0.6951 	&	0.3909 	&	-0.2798 	&	0.3852 	&	-0.1054 	&	0.4298 	&	0.0831 	&	0.4677 	&	0.1232 	&	0.5088 	\\
&	DSP-mix	&	-0.3676 	&	0.3578 	&	0.1077 	&	0.4634 	&	0.2991 	&	0.5203 	&	0.4819 	&	0.5630 	&	0.8625 	&	0.6147 	\\
&	MSP	&	-0.7145 	&	0.3640 	&	-0.3676 	&	0.3578 	&	-0.1418 	&	0.3914 	&	-0.0605 	&	0.4254 	&	0.1077 	&	0.4634 	\\
						& (maxSD) &      	0.3285 	&	0.0888 	&	0.3285 	&	0.0005 	&	0.1064 	&	0.0005 	&	0.1113 	&	0.0006 	&	0.0973 	&	0.0007 		     \\	
		 \midrule
		 \multirow{3}{*}{10}	&	DSP	&	-0.9253 	&	0.4730 	&	-0.0914 	&	0.5246 	&	0.1218 	&	0.5767 	&	0.4481 	&	0.6182 	&	0.5655 	&	0.6593 	\\
&	DSP-mix	&	-0.1747 	&	0.4957 	&	0.5718 	&	0.6209 	&	0.8718 	&	0.6745 	&	1.2397 	&	0.7188 	&	1.5495 	&	0.7573 	\\
&	MSP	&	-0.9281 	&	0.4732 	&	-0.1747 	&	0.4958 	&	0.0123 	&	0.5399 	&	0.2138 	&	0.5793 	&	0.5718 	&	0.6209 	\\
						& (maxSD) &        	0.2714 	&	0.0002 	&	0.1065 	&	0.0003 	&	0.0482 	&	0.0005 	&	0.0852 	&	0.0005 	&	0.0401 	&	0.0003 		     \\	
		 \midrule
		 \multirow{3}{*}{15}	&	DSP	&	-0.6121 	&	0.7473 	&	1.0220 	&	0.8125 	&	1.7181 	&	0.8518 	&	2.2469 	&	0.8775 	&	2.8725 	&	0.9002 	\\
&	DSP-mix	&	0.9721 	&	0.7955 	&	2.6561 	&	0.8834 	&	3.4443 	&	0.9143 	&	4.0266 	&	0.9337 	&	4.5270 	&	0.9455 	\\
&	MSP	&	-0.6122 	&	0.7473 	&	0.9721 	&	0.7955 	&	1.6077 	&	0.8333 	&	2.1442 	&	0.8597 	&	2.6561 	&	0.8834 	\\
						& (maxSD) &      	0.1532 	&	0.0003 	&	0.0943 	&	0.0004 	&	0.0908 	&	0.0004 	&	0.0768 	&	0.0004 	&	0.0234 	&	0.0003 		       \\	
		 \midrule
		 \multirow{3}{*}{20}	&	DSP	&	0.7989 	&	0.9095 	&	3.5479 	&	0.9432 	&	4.6367 	&	0.9587 	&	5.3676 	&	0.9682 	&	6.2838 	&	0.9773 	\\
&	DSP-mix	&	3.5414 	&	0.9382 	&	6.1021 	&	0.9721 	&	7.1799 	&	0.9818 	&	7.9012 	&	0.9856 	&	8.4159 	&	0.9892 	\\
&	MSP	&	0.7989 	&	0.9095 	&	3.5414 	&	0.9382 	&	4.5788 	&	0.9533 	&	5.2970 	&	0.9629 	&	6.1021 	&	0.9721 	\\	
						& (maxSD) &      	0.1970 	&	0.0002 	&	0.0908 	&	0.0002 	&	0.0681 	&	0.0002 	&	0.0884 	&	0.0002 	&	0.0258 	&	0.0002 		      \\	
		 \midrule
		 \multirow{3}{*}{30}	&	DSP	&	6.7636 	&	0.9927 	&	11.2454 	&	0.9963 	&	12.8170 	&	0.9978 	&	14.0692 	&	0.9986 	&	15.1920 	&	0.9990 	\\
&	DSP-mix	&	11.2116 	&	0.9960 	&	15.1105 	&	0.9988 	&	16.4078 	&	0.9992 	&	17.1863 	&	0.9996 	&	18.0573 	&	0.9998 	\\
&	MSP	&	6.7636 	&	0.9927 	&	11.2116 	&	0.9960 	&	12.7554 	&	0.9975 	&	13.9290 	&	0.9984 	&	15.1105 	&	0.9988 	\\
						& (maxSD) &      	0.2761 	&	0.0001 	&	0.1189 	&	0.0001 	&	0.1874 	&	0.0000 	&	0.1519 	&	0.0000 	&	0.1898 	&	0.0000 		      \\																		
		\bottomrule
	\end{tabular}
	\label{hdCauchy_error_N1e7}
\end{table}

\begin{table}[H]
	\centering
	\caption{$d$-D Cauchy mixtures: The partition level $L$ and computational time (in seconds) under different $d$ and $\theta$. The sample size is fixed to be $N  = 10^7$.}
	\setlength{\tabcolsep}{4pt}
	\begin{tabular}{c|c|cc|cc|cc|cc|cc}
		\toprule
		& $\theta$	& \multicolumn{2}{c}{$0.002$} & \multicolumn{2}{c}{$0.02$} & \multicolumn{2}{c}{$0.05$} &  \multicolumn{2}{c}{$0.1$} &  \multicolumn{2}{c}{$0.2$}\\ 
		\midrule
		 $d$	&	Method	& $L$ & time & $L$ & time  & $L$ & time  & $L$ & time  & $L$ & time \\
		 \midrule
		 \multirow{3}{*}{2}	&	DSP	&	173841 	&	151.72 	&	16211 	&	87.81 	&	6639 	&	105.19 	&	3332 	&	126.31 	&	1691 	&	146.15 	\\
&	DSP-mix	&	112188 	&	8.84 	&	16154 	&	9.98 	&	6583 	&	16.92 	&	3274 	&	28.84 	&	1625 	&	52.87 	\\
&	MSP	&	280402 	&	8.83 	&	18032 	&	6.07 	&	6925 	&	5.63 	&	3698 	&	5.25 	&	2059 	&	5.12 	\\						
		 \midrule
		 \multirow{3}{*}{4}	&	DSP	&	648039 	&	687.78 	&	27136 	&	210.24 	&	11910 	&	192.20 	&	6492 	&	196.96 	&	3659 	&	205.85 	\\
&	DSP-mix	&	175907 	&	14.01 	&	17613 	&	17.35 	&	7307 	&	29.68 	&	3608 	&	51.81 	&	1890 	&	94.56 	\\
&	MSP	&	1043302 	&	17.84 	&	115907 	&	9.39 	&	52032 	&	9.07 	&	28907 	&	8.62 	&	15741 	&	8.29 	\\
		 \midrule
		 \multirow{3}{*}{6}	&	DSP	&	1006568 	&	678.35 	&	53796 	&	309.37 	&	23027 	&	257.94 	&	12423 	&	244.18 	&	6906 	&	245.08 	\\
&	DSP-mix	&	191134 	&	18.10 	&	19539 	&	24.00 	&	8038 	&	42.36 	&	4164 	&	73.33 	&	2158 	&	137.66 	\\
&	MSP	&	1236975 	&	22.90 	&	188354 	&	13.62 	&	76132 	&	12.40 	&	38998 	&	11.21 	&	19520 	&	10.63 	\\		
		 \midrule
		 \multirow{3}{*}{8}	&	DSP	&	1325701 	&	553.24 	&	85691 	&	375.92 	&	35788 	&	290.85 	&	18973 	&	274.74 	&	10181 	&	272.09 	\\
&	DSP-mix	&	209324 	&	23.47 	&	21821 	&	31.33 	&	8940 	&	54.64 	&	4658 	&	95.43 	&	2379 	&	177.23 	\\
&	MSP	&	1390910 	&	29.26 	&	209290 	&	17.38 	&	84843 	&	16.10 	&	43202 	&	14.63 	&	21821 	&	13.77 	\\
		 \midrule
		 \multirow{3}{*}{10}	&	DSP	&	1546736 	&	461.04 	&	119372 	&	408.73 	&	49117 	&	319.62 	&	25449 	&	301.88 	&	13155 	&	296.88 	\\
&	DSP-mix	&	229815 	&	28.18 	&	23945 	&	37.43 	&	9809 	&	66.84 	&	5086 	&	116.65 	&	2651 	&	210.02 	\\
&	MSP	&	1555407 	&	33.94 	&	229815 	&	21.78 	&	93768 	&	18.77 	&	47998 	&	17.08 	&	23945 	&	16.16 	\\	
		 \midrule
		 \multirow{3}{*}{15}	&	DSP	&	1977162 	&	340.83 	&	204560 	&	511.87 	&	80332 	&	421.72 	&	39741 	&	405.71 	&	19590 	&	395.28 	\\
&	DSP-mix	&	286280 	&	38.17 	&	28866 	&	54.95 	&	11644 	&	97.07 	&	5792 	&	164.47 	&	3199 	&	290.12 	\\
&	MSP	&	1977232 	&	51.67 	&	286280 	&	33.44 	&	116620 	&	28.26 	&	58439 	&	26.55 	&	28866 	&	24.54 	\\	
		 \midrule
		 \multirow{3}{*}{20}	&	DSP	&	2399337 	&	328.32 	&	279421 	&	636.23 	&	106726 	&	576.62 	&	51803 	&	560.24 	&	24388 	&	534.52 	\\
&	DSP-mix	&	334965 	&	53.23 	&	32580 	&	71.41 	&	12502 	&	126.16 	&	6370 	&	212.48 	&	3633 	&	363.47 	\\
&	MSP	&	2399363 	&	71.01 	&	334964 	&	44.12 	&	135152 	&	37.84 	&	67249 	&	36.07 	&	32580 	&	32.42 	\\	
		 \midrule
		 \multirow{3}{*}{30}	&	DSP	&	3125241 	&	386.49 	&	389512 	&	890.28 	&	148772 	&	890.74 	&	68230 	&	867.85 	&	30332 	&	839.55 	\\
&	DSP-mix	&	407694 	&	81.72 	&	35503 	&	104.93 	&	13466 	&	182.92 	&	7247 	&	298.40 	&	4264 	&	513.47 	\\
&	MSP	&	3125248 	&	109.90 	&	407670 	&	66.05 	&	164124 	&	59.87 	&	77742 	&	52.82 	&	35503 	&	48.01 	\\																		
		\bottomrule
	\end{tabular}
	\label{hdCauchy_K_time_N1e7}
\end{table}

\begin{table}[H]
	\centering
	\caption{$d$-D Cauchy mixtures: The KL divergence and Hellinger distance  under different $d$ and $N$. The parameter $\theta$ is fixed to be $0.002$.}
	\setlength{\tabcolsep}{4pt}
	\begin{tabular}{c|c|cc|cc|cc|cc|cc}
		\toprule
		& $N$	& \multicolumn{2}{c}{$1\times10^4$} & \multicolumn{2}{c}{$1\times10^5$} & \multicolumn{2}{c}{$1\times10^6$} &  \multicolumn{2}{c}{$1\times10^7$} &  \multicolumn{2}{c}{$1\times10^8$}\\ 
		\midrule
		 $d$	&	Method	& KL & $\hat{H}^2$ &  KL & $\hat{H}^2$  & KL & $\hat{H}^2$  & KL & $\hat{H}^2$  & KL & $\hat{H}^2$  \\
		 \midrule
		 \multirow{3}{*}{15}	&	DSP 		&	3.4022 	&	0.9513 	&	1.4427 	&	0.9019 	&	0.1989 	&	0.8318 	&	-0.6121 	&	0.7473 	&	1.0773 	&	0.6183 	\\
		 				&	DSP-mix	&	3.4022 	&	0.9513 	&	1.4427 	&	0.9018 	&	1.0771 	&	0.8454 	&	0.9721 	&	0.7955 	&	2.7906 	&	0.7060 	\\
						&	MSP		&	3.4022 	&	0.9514 	&	1.4427 	&	0.9018 	&	0.1739 	&	0.8319 	&	-0.6122 	&	0.7473 	&	2.7906 	&	0.7062  	\\
		 \midrule
		 \multirow{3}{*}{20}	&	DSP 		&	7.1493 	&	0.9904 	&	4.3705 	&	0.9767 	&	2.4521 	&	0.9500 	&	0.7989 	&	0.9095 	&	2.7486 	&	0.8506 	\\
		 				&	DSP-mix	&	7.1493 	&	0.9904 	&	4.3705 	&	0.9767 	&	3.7805 	&	0.9567 	&	3.5414 	&	0.9382 	&	5.2577 	&	0.9047 	\\
						&	MSP		&	7.1493 	&	0.9904 	&	4.3705 	&	0.9767 	&	2.4521 	&	0.9500 	&	0.7989 	&	0.9095 	&	5.2577 	&	0.9047 	\\
		 \midrule
		 \multirow{3}{*}{30}	&	DSP 		&	15.9541 	&	0.9997 	&	12.5253 	&	0.9990 	&	9.2726 	&	0.9970 	&	6.7636 	&	0.9927 	&	6.3239 	&	0.9863 	\\
		 				&	DSP-mix	&	15.9541 	&	0.9997 	&	12.5253 	&	0.9990 	&	11.4016 	&	0.9977 	&	11.2116 	&	0.9960 	&	10.3196 	&	0.9940 	\\
						&	MSP		&	15.9541 	&	0.9997 	&	12.5253 	&	0.9990 	&	9.2726 	&	0.9970 	&	6.7636 	&	0.9927 	&	10.3196 	&	0.9940 	\\																		
		\bottomrule
	\end{tabular}
	\label{hdCauchy_error_high}
\end{table}

\begin{table}[H]
	\centering
	\caption{$d$-D Cauchy mixtures: The partition level $L$ and computational time (in seconds) under different $d$ and $N$. The parameter $\theta$ is fixed to be $0.002$.}
	\setlength{\tabcolsep}{4pt}
	\begin{tabular}{c|c|cc|cc|cc|cc|cc}
		\toprule
		& $N$	& \multicolumn{2}{c}{$1\times10^4$} & \multicolumn{2}{c}{$1\times10^5$} & \multicolumn{2}{c}{$1\times10^6$} &  \multicolumn{2}{c}{$1\times10^7$} &  \multicolumn{2}{c}{$1\times10^8$}\\ 
		\midrule
		 $d$	&	Method	& $L$ & time & $L$ & time  & $L$ & time  & $L$ & time  & $L$ & time \\
		 \midrule
		 \multirow{3}{*}{15}	&	DSP 		&	2843 	&	0.03 	&	24527 	&	0.25 	&	222954 	&	4.33 	&	1977162 	&	340.83 	&	7840212 	&	19084.37  	\\
		 				&	DSP-mix	&	2843 	&	0.03 	&	24527 	&	0.26 	&	106704 	&	3.53 	&	286280 	&	38.17 	&	987999 	&	596.05 	\\
						&	MSP		&	2843 	&	0.02 	&	24527 	&	0.24 	&	222960 	&	4.46 	&	1977232 	&	51.67 	&	987999 	&	570.70  	\\	
		 \midrule
		 \multirow{3}{*}{20}	&	DSP 		&	3431 	&	0.04 	&	29100 	&	0.37 	&	272278 	&	6.39 	&	2399337 	&	328.32 	&	10550658 	&	20532.97 	\\
		 				&	DSP-mix	&	3431 	&	0.04 	&	29100 	&	0.38 	&	127445 	&	4.97 	&	334965 	&	53.23 	&	1086785 	&	716.39 	\\
						&	MSP		&	3431 	&	0.04 	&	29100 	&	0.35 	&	272278 	&	6.36 	&	2399363 	&	71.01 	&	1086785 	&	732.24 	\\	
		 \midrule
		 \multirow{3}{*}{30}	&	DSP 		&	4408 	&	0.07 	&	34327 	&	0.67 	&	351954 	&	9.92 	&	3125241 	&	386.49 	&	13079357 	&	23887.23 	\\
		 				&	DSP-mix	&	4408 	&	0.07 	&	34327 	&	0.74 	&	162494 	&	7.88 	&	407694 	&	81.72 	&	1129381 	&	867.59 	\\
						&	MSP		&	4408 	&	0.07 	&	34327 	&	0.65 	&	351954 	&	10.32 	&	3125248 	&	109.90 	&	1129381 	&	894.90 	\\																		
		\bottomrule
	\end{tabular}
	\label{hdCauchy_K_time_high}
\end{table}

\section{Conclusion and discussion}\label{conclus}

This work proposes the DSP-mix and MSP methods, both of which adopt the same framework of the DSP method in \cite{Li2016} to learn a adaptive piecewise constant function to approximate the underlying probability density. The novelty lies in the measures of uniformity for observations. DSP-mix and MSP adopt the mixture discrepancy and moments, respectively, instead of the star discrepancy in DSP. The mixture discrepancy and moments are easier to compute and theoretically maintain reflection invariance and rotation invariance, which are the shortcomings of the star discrepancy. Numerical experiments up to 30 dimension demonstrate that MSP can maintain the same accuracy compared with DSP, while gaining an increase in speed by a factor of two to twenty for large sample size. DSP-mix achieves satisfactory accuracy and boosts the efficiency in low-dimensional tests ($d \le 6$), but might lose accuracy in high-dimensional problems due to a reduction in partition level.

As a summary, the accuracy of DSP, DSP-mix and MSP can be improved by increasing the sample size $N$ under appropriate partition level $L$, although it still suffers from the overfitting problem when $L$ is too large. Fortunately, the overfitting can be alleviated as more samples are used. In practice, we can utilize the trend $L \sim \theta^{-1}$ to estimate the partition level by first testing on a small dataset and using extrapolation. This gives us a simple criterion to control the computational complexity.

There are still some issues worthy of further research, such as the quantitative error analysis of DSP-mix and MSP, the application to practical density estimation problems in higher dimensions, and more efficient uniformity measurement approaches.

\section*{Acknowledgement}
This research was supported by the National Natural Science Foundation of China (Nos.~12325112, 12288101, 12571413) and the High-performance Computing Platform of Peking University. The authors are thankful to the anonymous referee for his/her valuable suggestions.



\end{document}